\documentclass[preprint,12pt,authoryear]{elsarticle}

\usepackage{amssymb}
\usepackage{amsmath}

\usepackage{geometry}
\usepackage{setspace}
\usepackage{amsmath,amssymb,amsfonts,bm}
\usepackage{subcaption}
\usepackage{float}
\usepackage[section]{placeins}
\usepackage{booktabs}
\usepackage{multirow}
\usepackage{array}
\usepackage{tabularx}
\usepackage{longtable}
\usepackage{makecell}
\usepackage{threeparttable}
\usepackage{caption}
\newcolumntype{Y}{>{\raggedright\arraybackslash}X}
\newcolumntype{L}[1]{>{\raggedright\arraybackslash}p{#1}}
\newcolumntype{C}[1]{>{\centering\arraybackslash}p{#1}}
\newcommand{\meanstd}[2]{\ensuremath{#1{\scriptstyle\,\pm\,#2}}}
\journal{Journal of Industrial Information Integration}

\begin{document}

\begin{frontmatter}



\title{TPA-AD: A Two-Stage Pseudo Anomaly-Guided Method for Bearing Time-Series Anomaly Detection} 


\author[1]{Xiancheng Wang}
\ead{402196277@qq.com}

\author[2]{Zhibo Zhang}
\ead{zhangzhibo@cqsf.com}

\author[3]{Ran Li}
\ead{13573280825@163.com}

\author[1]{Rui Wang}
\ead{wangrui@hitwh.edu.cn}

\author[1]{Minghang Zhao}
\ead{zhaomh@hit.edu.cn}

\author[1]{Shisheng Zhong}
\ead{zhongss@hit.edu.cn}

\author[1]{Lin Wang\corref{cor1}}
\ead{wanglin\_007@hitwh.edu.cn}

\affiliation[1]{organization={School of Ocean Engineering, Harbin Institute of Technology},
            addressline={West Wenhua Road}, 
            city={Weihai},
            postcode={264209}, 
            state={Shandong},
            country={China}}

\affiliation[2]{organization={Technical Center, Bogie Development Department, CRRC Qingdao Sifang Locomotive and Rolling Stock Co., Ltd},
            addressline={Jinhong East Road}, 
            city={Qingdao},
            postcode={266111}, 
            state={Shandong},
            country={China}}

\affiliation[3]{organization={Qingdao University},
            addressline={No. 308 Ningxia Road},
            city={Qingdao},
            postcode={266071},
            state={Shandong},
            country={China}}
\cortext[cor1]{Corresponding author}


\begin{abstract}
This paper proposes a two-stage pseudo anomaly-guided anomaly detection method (\textbf{T}wo-stage \textbf{P}seudo \textbf{A}nomaly-guided \textbf{A}nomaly \textbf{D}etection, \textbf{TPA-AD}) for axle-box bearing time-series anomaly detection (time series anomaly detection, TSAD) under the setting where only normal samples are available for training. The method first generates pseudo-anomalous windows near the normal boundary using a reconstruction model and per-feature target-error control. It then learns anomaly-sensitive representations through contrastive learning between normal and pseudo-anomalous windows, and finally produces window-level and point-level anomaly scores using k-nearest neighbors (KNN). Compared with existing methods that rely on known fault categories, real anomaly priors, or random anomaly injection, TPA-AD improves the separability of the normal boundary by constructing pseudo-anomalies in boundary neighborhoods and can jointly handle continuous and discrete features in mixed-variable scenarios. The main experiments are conducted on bearing fault detection datasets and degradation-process datasets, with an additional exploratory extension on $13$ public TSAD datasets. The results show that the proposed method yields relatively stable anomaly responses, is sensitive to degradation evolution, and demonstrates a certain degree of broader applicability on public TSAD benchmarks and real high-speed-train-related bearing data.

\end{abstract}


\begin{highlights}
\item We propose a two-stage pseudo anomaly-guided framework for scenarios with normal-only training, combining reconstruction-driven pseudo-anomaly generation with contrastive representation learning to improve the separability of axle-box bearing anomalies without relying on real anomaly samples.
\item We design a per-feature target-error control and pseudo-anomaly filtering mechanism to generate boundary samples with controllable deviation magnitudes and a more continuous distribution in the continuous-feature subspace, thereby alleviating the problems of scattered injected anomalies and unstable boundaries in traditional anomaly injection.
\item We construct an experimental system covering fault detection, degradation detection, and an extension on public TSAD datasets, systematically evaluating the proposed method on multi-condition bearing data and long-horizon degradation sequences, while also analyzing its broader applicability to generic time-series anomaly detection tasks.
\end{highlights}


\end{frontmatter}




\section{Introduction}

Axle-box bearings of high-speed trains are critical rotating components in the bogie system, and their service condition directly affects operational safety, ride comfort, and maintenance cost. During long-term operation, axle-box bearings are influenced by high-speed rotation, wheel--rail excitation, load variation, track irregularity, and environmental noise. Under these conditions, abnormal states such as local damage, wear, and lubrication degradation may gradually accumulate and eventually evolve into severe faults. Vibration signals can sensitively reflect impact, modulation, and non-stationary dynamic characteristics inside the bearing. Therefore, state monitoring and anomaly detection based on vibration signals collected by axle-box accelerometers are of great practical importance. In recent years, rolling-bearing fault diagnosis has gradually shifted from handcrafted feature extraction to deep representation learning. Related surveys indicate that convolutional neural networks, attention mechanisms, graph neural networks, and Transformer-based models have become important tools for intelligent bearing diagnosis \cite{zhao2025comprehensive}. Under complex operating conditions, previous studies have improved model robustness from perspectives such as variable speed and sample imbalance \cite{dong2024rolling}, pseudo-label-based time--frequency supervised contrastive learning and unsupervised domain adaptation \cite{pang2024time}, and dynamic-model-assisted disentanglement \cite{xu2025dynamic}. In terms of model design, methods such as convolution with cross-fusion Transformer \cite{lin2023ccft}, GAF combined with CNN-ViT \cite{zhou2024novel}, and few-shot learning frameworks \cite{li2024deep} have been used to enhance vibration-signal feature representation. In multi-sensor and industrial scenarios, approaches such as multi-sensor frequency-domain fusion \cite{dai2025msff}, VMD combined with lightweight networks \cite{wang2024multi}, lightweight contrastive Transformer \cite{dong2024multi}, and multi-source residual convolutional fusion networks \cite{ye2025mrcfn} further address noise, limited samples, and multi-source information fusion. For high-speed-train bogie bearings, methods such as AGFCN consider diagnosis under complex conditions with strong noise and varying loads \cite{he2025agfcn}. In addition, zero-shot diagnosis \cite{li2025zero}, frequency-pattern graph modeling \cite{liu2025frequency}, and complex-domain completion with unsupervised time--frequency alignment under cross-domain missing-data settings \cite{wang2025intelligent} all suggest that real bearing-monitoring scenarios often involve insufficient labels, cross-condition shifts, scarce anomaly samples, and multi-source noise interference simultaneously.
\begin{figure}[!t]
    \centering
    \includegraphics[width=0.98\linewidth]{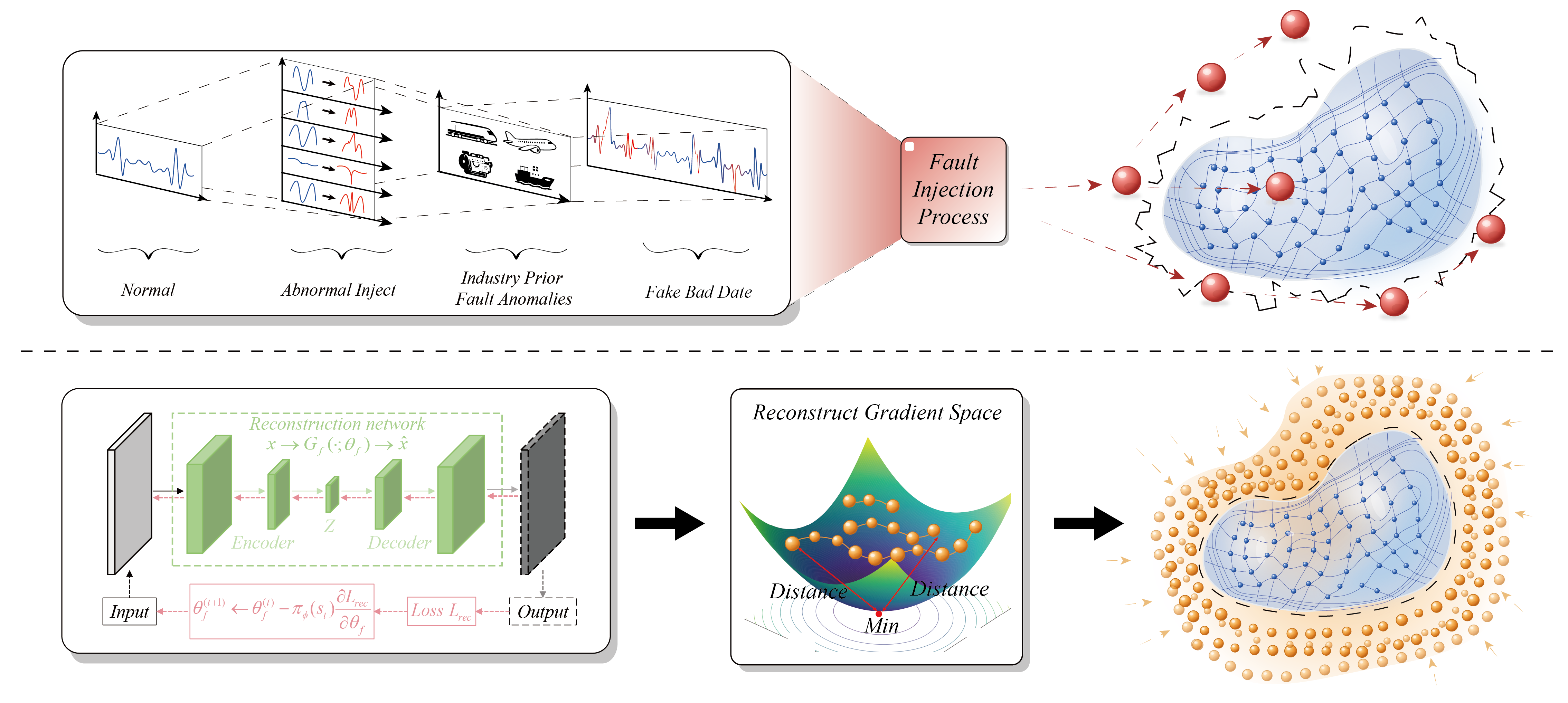}
    \caption{Schematic comparison between conventional fault injection and the pseudo-anomaly generation mechanism proposed in this paper. The former usually produces scattered negative samples and thus struggles to form a stable boundary outside the normal manifold. The latter generates pseudo-anomalous samples with multiple strength levels in the reconstruction-error space, thereby forming a more continuous anomalous boundary band around the normal region.}
    \label{fig:intro_pseudo_boundary}
\end{figure}
However, most existing bearing fault diagnosis studies target the identification of known fault categories and usually rely on fault samples, class labels, or source-domain fault knowledge. In real high-speed-train maintenance scenarios, severe fault samples are scarce, anomaly labels are difficult to obtain, and many anomalies do not belong to predefined fault categories. Therefore, modeling axle-box vibration monitoring as a time-series anomaly detection (TSAD) problem is more consistent with practical application needs. In recent years, deep TSAD methods have developed into a relatively complete research landscape, covering prediction-based, reconstruction-based, generative, and representation-learning paradigms \cite{zamanzadeh2024deep}. At the same time, the reliability of anomaly-detection benchmarks, evaluation metrics, and experimental protocols has also drawn increasing attention \cite{liu2024elephant,qiu2025tab}. In terms of specific methods, multi-pattern normality learning in the frequency domain \cite{chen2024learning}, joint self-supervised modeling in the time, frequency, and residual domains \cite{sun2024unraveling}, and time--frequency masked autoencoders \cite{fang2024temporal} all highlight the importance of frequency structure and multi-domain representation for anomaly detection. ImDiffusion combines imputation mechanisms with diffusion models for multivariate TSAD \cite{chen2023imdiffusion}, while METER addresses concept drift in online anomaly detection through dynamic concept adaptation \cite{zhu2023meter}. For multivariate time series, graph-alignment methods are used to characterize changes in channel dependencies \cite{wang2024interdependency}, and CATCH captures fine-grained frequency features and channel correlations through frequency-domain patching and channel-fusion modules \cite{wu2025catch}. In addition, recent methods such as DADA \cite{shentu2025towards}, hybrid prototype learning \cite{shen2025learn}, CARLA self-supervised contrastive representation learning \cite{darban2025carla}, KAN-AD \cite{zhou2024kan}, and RTDetector based on reconstruction trends \cite{liu2025rtdetector} have further advanced TSAD from the perspectives of generic anomaly detection, prototype constraints, contrastive learning, and suppression of reconstruction overfitting.

Although the above methods have achieved promising results in generic anomaly detection and mechanical fault diagnosis, two key limitations remain for axle-box bearing monitoring scenarios in which only normal training samples are available. First, many methods implicitly rely on real anomalies, anomaly-injection priors, or relatively stable fault templates. Second, even when artificial negative samples are introduced, they are often scattered and thus fail to form a continuous and stable discriminative boundary outside the normal manifold. Therefore, a key issue in improving the discriminative capability of TSAD for axle-box bearings is how to construct ``near-anomalous'' reference samples with genuine boundary significance without relying on real anomaly samples.

As illustrated in Fig.~\ref{fig:intro_pseudo_boundary}, conventional fault injection is usually highly task-specific, and the generated samples tend to resemble several known fault patterns. When unseen anomalies or weak anomalies at the early stage of degradation appear during testing, such models may fail to maintain stable discrimination. In contrast, this paper does not directly fabricate specific fault morphologies. Instead, it generates pseudo-anomalous windows with controllable deviation magnitudes based on the reconstruction behavior of normal samples within multiple feature-wise error intervals. The resulting pseudo-anomalous samples are more concentrated in the neighborhood just outside the normal manifold, thereby providing continuous and stable negative references for subsequent contrastive representation learning.

To more clearly position this work with respect to existing research on bearing fault diagnosis and time-series anomaly detection, Tables~\ref{tab:bearing_key_refs} and \ref{tab:tsad_key_refs} summarize two groups of representative studies that are most relevant to this paper.

\begin{table}[!t]
\centering
\scriptsize
\setlength{\tabcolsep}{3pt}
\renewcommand{\arraystretch}{0.94}
\caption{Key references in bearing fault diagnosis}
\label{tab:bearing_key_refs}
\begin{tabularx}{\textwidth}{p{2.2cm}p{2.6cm}p{4.0cm}Y}
\toprule
Reference & Scenario/Problem & Method & Relevance to this work \\
\midrule
Dong et al.\cite{dong2024rolling} & Variable speed and sample imbalance & Multi-scale dynamic supervised contrastive learning for enhancing discrimination among different state representations. & Supports the effectiveness of contrastive learning in bearing diagnosis under complex operating conditions. \\
Pang et al.\cite{pang2024time} & Unsupervised domain adaptation under variable speed & Combines pseudo labels with time--frequency supervised contrastive learning to alleviate cross-condition distribution shifts. & Related to our use of pseudo-anomaly/pseudo-label information for representation learning. \\
He et al.\cite{he2025agfcn} & Complex operating conditions of high-speed-train bogie bearings & Fault diagnosis for high-speed-train bogie bearings under strong noise and varying loads. & Closest to the application scenario of this paper and supports the background of axle-box bearing vibration monitoring. \\
Li et al.\cite{li2025zero} & Zero-shot bearing diagnosis & Uses fault-spectrum knowledge and self-driven contrastive learning under the absence of fault samples. & Supports the need to exploit weak priors or pseudo-information when real fault samples are scarce. \\
Wang et al.\cite{wang2025intelligent} & Cross-domain diagnosis with missing data & Addresses cross-domain and missing-data issues through complex-domain completion and unsupervised time--frequency alignment. & Supports the need to handle distribution shifts, missingness, and unsupervised alignment in real industrial data. \\
\bottomrule
\end{tabularx}
\end{table}

\vspace{-0.9em}
\begin{table}[!t]
\centering
\scriptsize
\setlength{\tabcolsep}{3pt}
\renewcommand{\arraystretch}{0.94}
\caption{Key references in time-series anomaly detection}
\label{tab:tsad_key_refs}
\begin{tabularx}{\textwidth}{p{2.2cm}p{2.6cm}p{4.0cm}Y}
\toprule
Reference & Scenario/Problem & Method & Relevance to this work \\
\midrule
Zamanzadeh Darban et al.\cite{zamanzadeh2024deep} & Survey of deep TSAD & Systematically summarizes prediction-based, reconstruction-based, generative, and representation-learning methods. & Provides the overall TSAD research context and problem definition. \\
Liu and Paparrizos\cite{liu2024elephant} & TSAD benchmarks and evaluation & Analyzes how datasets, metrics, and evaluation protocols affect result reliability. & Supports the need for point-level scoring and careful metric design in this work. \\
Chen et al.\cite{chen2024learning} & Multi-pattern normality learning & Learns multiple normal patterns in the frequency domain to improve anomaly detection efficiency in complex systems. & Relevant to multiple normal operating modes and frequency structure in axle-box vibration signals. \\
Wu et al.\cite{wu2025catch} & Multivariate frequency-domain and channel relations & Uses frequency-domain patching and channel-fusion modules to capture frequency features and channel correlations. & Supports the importance of frequency-domain and channel information in multi-channel vibration signals. \\
Darban et al.\cite{darban2025carla} & Self-supervised contrastive TSAD & Obtains discriminative time-series representations through anomaly injection and contrastive learning. & Related to our idea of constructing boundaries in the embedding space using pseudo-anomalous negative samples. \\
\bottomrule
\end{tabularx}
\end{table}

In summary, this paper establishes a TSAD framework for high-speed-train axle-box bearing monitoring centered on ``pseudo-anomaly construction--contrastive representation learning--point-level anomaly scoring.'' The main contributions of this paper are as follows:
\begin{enumerate}
    \item We propose a two-stage pseudo anomaly-guided framework for scenarios with normal-only training, combining reconstruction-driven pseudo-anomaly generation with contrastive representation learning to improve the separability of axle-box bearing anomalies without relying on real anomaly samples.
    \item We design a per-feature target-error control and pseudo-anomaly filtering mechanism to generate boundary samples with controllable deviation magnitudes and a more continuous distribution in the continuous-feature subspace, thereby alleviating the problems of scattered injected anomalies and unstable boundaries in traditional anomaly injection.
    \item We construct an experimental system covering fault detection, degradation detection, and an extension on public TSAD datasets, systematically evaluating the proposed method on multi-condition bearing data and long-horizon degradation sequences, while also analyzing its broader applicability to generic time-series anomaly detection tasks.
\end{enumerate}

\section{Method}
\label{sec:method}

\begin{figure}[t]
    \centering
    \includegraphics[width=\linewidth]{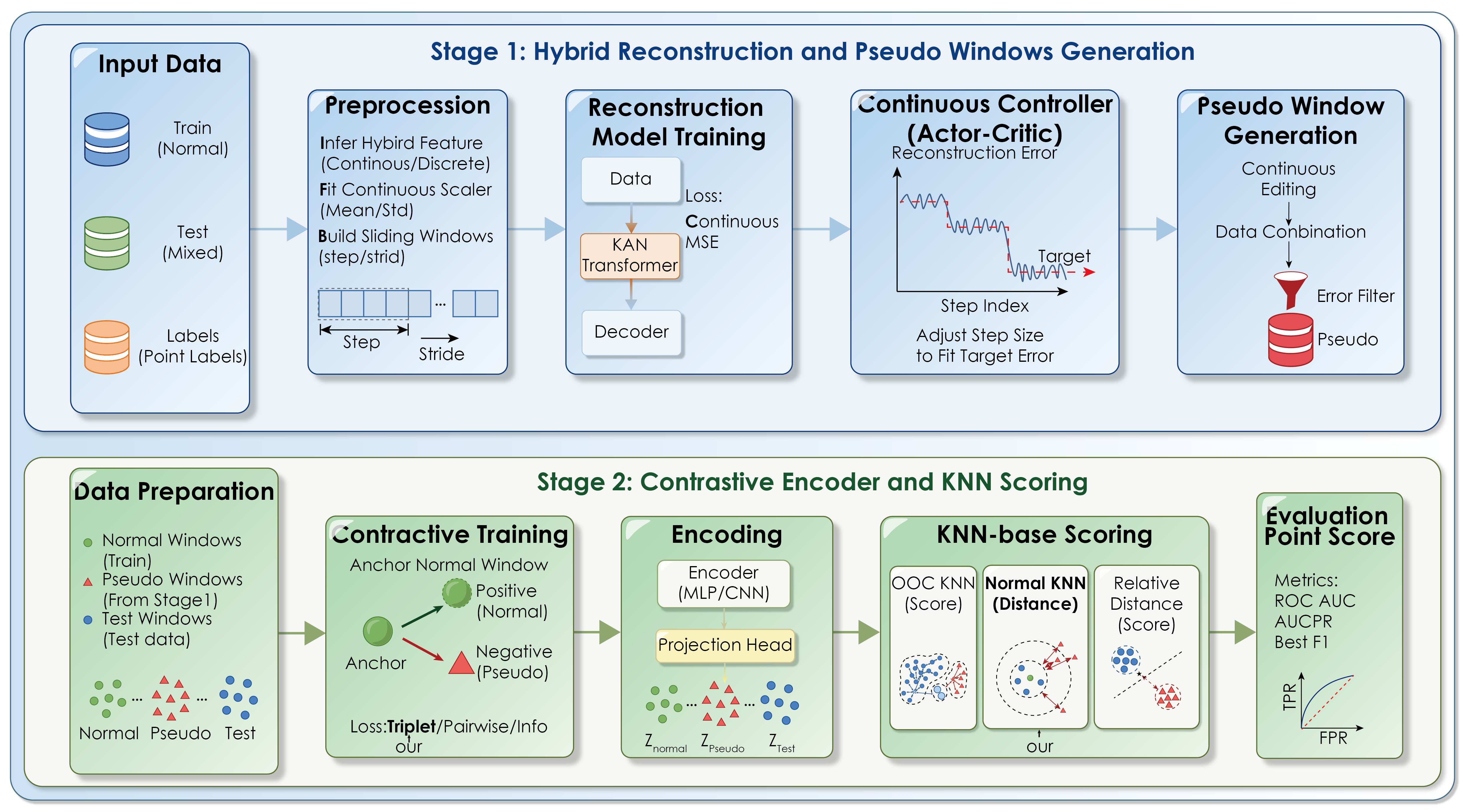}
    \caption{The proposed two-stage framework. The upper part shows reconstruction-driven pseudo-anomalous window generation, and the lower part shows normal--pseudo-anomalous contrastive representation learning and KNN scoring. Stage 1 generates controlled pseudo-anomalous samples from normal windows, while Stage 2 pulls normal neighborhoods closer, pushes pseudo-anomalous samples farther away in the embedding space, and outputs both window-level and point-level anomaly scores.}
    \label{fig:overall_framework}
\end{figure}

This paper studies time-series anomaly detection under the setting where only normal samples are used for training. Given normal training sequences, the goal is to assign higher scores to anomalous time points in the test sequence without using test labels. To this end, we propose a two-stage pseudo anomaly-guided representation learning framework, as shown in Fig.~\ref{fig:overall_framework}. In the first stage, a reconstruction model is learned in the continuous subspace, and pseudo-anomalous windows near the normal boundary are generated through per-feature error control. In the second stage, contrastive representation learning is performed using normal windows and pseudo-anomalous windows, and anomaly scoring is carried out in the embedding space based on KNN distances. By constructing ``near-anomalous'' reference samples, the framework enhances the separability of the normal boundary without relying on any real anomaly samples.

\subsection{Problem formulation}
\label{subsec:problem_formulation}

Let the training sequence and test sequence be denoted by \(\mathbf{X}^{\mathrm{tr}}=\{\mathbf{x}^{\mathrm{tr}}_t\}_{t=1}^{T_{\mathrm{tr}}}\) and \(\mathbf{X}^{\mathrm{te}}=\{\mathbf{x}^{\mathrm{te}}_t\}_{t=1}^{T_{\mathrm{te}}}\), respectively, where \(\mathbf{x}_t\in\mathbb{R}^{D}\). The training set contains only normal operating states, while the test set may contain anomalous segments. The test labels \(\mathbf{y}^{\mathrm{te}}\in\{0,1\}^{T_{\mathrm{te}}}\) are used only for final evaluation and do not participate in model training or pseudo-anomaly generation.

Given a window length \(W\) and a sliding stride \(s\), the sequence is segmented into a set of windows
\begin{equation}
    \mathbf{X}_{i}=
    [\mathbf{x}_{t_i},\mathbf{x}_{t_i+1},\ldots,\mathbf{x}_{t_i+W-1}]
    \in\mathbb{R}^{W\times D},\quad
    t_i=1+(i-1)s.
\end{equation}
During evaluation, the window label is determined by the point labels within its covered interval, namely \(y_i=\max_{t\in[t_i,t_i+W-1]} y_t\).
The goal of this paper is to learn a point-level anomaly scoring function \(S(t)\) such that anomalous segments receive higher scores.

\subsection{Preprocessing for mixed features}
\label{subsec:hybrid_preprocess}

Industrial sensing data often contain both continuous features and low-cardinality discrete features. We therefore first identify the feature type according to the cardinality observed in the training set: if the number of unique values in the \(j\)-th dimension of the training set does not exceed a threshold \(K_{\max}\), that dimension is treated as discrete; otherwise, it is treated as continuous. Let the index sets of continuous and discrete features be \(\mathcal{C}\) and \(\mathcal{D}\), respectively, satisfying \(\mathcal{C}\cup\mathcal{D}=\{1,\ldots,D\}\) and \(\mathcal{C}\cap\mathcal{D}=\emptyset\).

Stage 1 performs reconstruction and pseudo-anomaly generation only in the continuous subspace. For \(j\in\mathcal{C}\), feature-wise scaling is applied using training-set statistics, i.e., \(\tilde{x}_{t,j}=\frac{x_{t,j}-a_j}{b_j}\), where \(a_j\) and \(b_j\) are estimated from the training set. The framework supports three scaling schemes: z-score, min--max, and robust min--max. Discrete features are not involved in the continuous editing procedure of Stage 1; instead, they are modeled and scored later through an independent discrete KNN branch.

\subsection{Stage 1: Reconstruction-driven pseudo-anomalous window generation}
\label{subsec:stage1}

\begin{figure}[t]
    \centering
    \includegraphics[width=\linewidth]{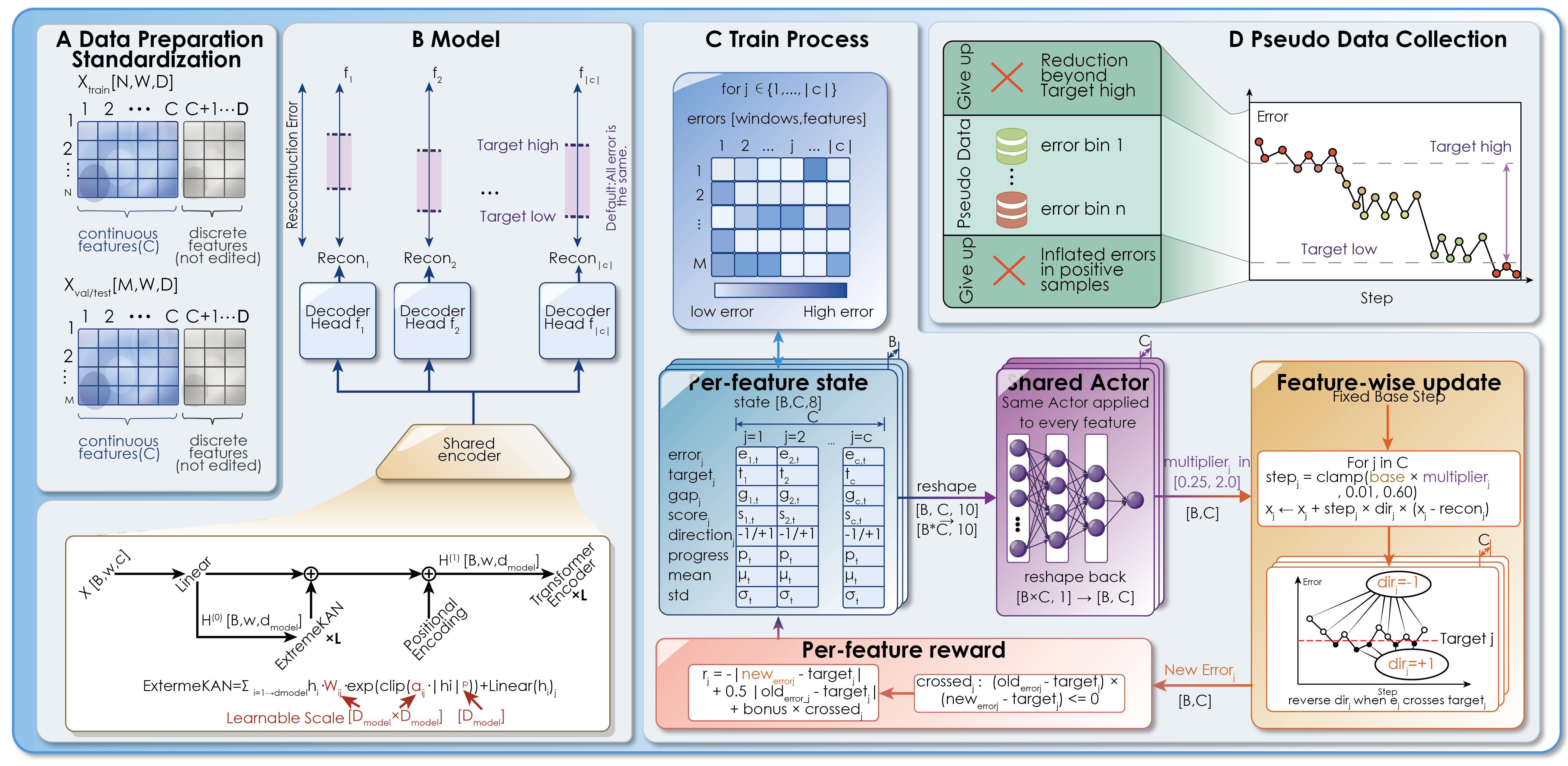}
    \caption{Stage 1 pipeline. The left part shows data normalization in the continuous subspace and the reconstruction model; the middle part shows a shared controller based on per-feature states; and the right part shows the pseudo-anomalous window collection process based on target-error constraints and error-bin balancing.}
    \label{fig:stage1}
\end{figure}

As shown in Fig.~\ref{fig:stage1}, Stage 1 consists of four steps: continuous-feature normalization, reconstruction-model training, per-feature error control, and pseudo-anomalous window collection. The goal of this stage is not to directly perform anomaly discrimination, but rather to generate a set of boundary samples with controllable deviation magnitudes around the normal manifold, thereby providing effective negative references for the representation learning in Stage 2.

\subsubsection{Continuous-subspace reconstruction model}
\label{subsubsec:continuous_reconstruction}

Let the representation of window \(\mathbf{X}_i\) in the continuous subspace be denoted by \(\tilde{\mathbf{X}}^{\mathcal{C}}_{i}\in\mathbb{R}^{W\times |\mathcal{C}|}\). For each input window, an initial hidden representation is first obtained through a linear layer, i.e., \(\mathbf{H}^{(0)}=\mathrm{Linear}\big(\tilde{\mathbf{X}}^{\mathcal{C}}_{i}\big)\).
Then, an ExtremeKAN module is introduced to enhance amplitude-related nonlinear responses, and its output is injected into the Transformer encoder as a residual term:
\begin{equation}
    \mathbf{H}^{(1)}
    =
    \mathrm{TransformerEncoder}
    \left(
    \mathrm{PE}\big(
    \mathbf{H}^{(0)}+\mathrm{EKAN}(\mathbf{H}^{(0)})
    \big)
    \right).
\end{equation}
Here, \(\mathrm{PE}(\cdot)\) denotes positional encoding and \(\mathrm{EKAN}(\cdot)\) denotes the ExtremeKAN nonlinear enhancement module.

To preserve the independent reconstruction capability of each sensor dimension, a dedicated decoding head is assigned to every continuous feature. Let \(\mathbf{h}_{i,t}\) denote the hidden state of \(\mathbf{H}^{(1)}\) at time step \(t\). The reconstruction result for the \(j\)-th continuous feature is then \(\hat{x}_{i,t,j}=D_j(\mathbf{h}_{i,t})\), where \(j\in\mathcal{C}\) and \(D_j(\cdot)\) is the feature-specific decoding mapping. Accordingly, the window-wise, feature-wise reconstruction error is defined as
\begin{equation}
    e_{i,j}
    =
    \frac{1}{W}\sum_{t=1}^{W}
    \left(
    \tilde{x}_{i,t,j}-\hat{x}_{i,t,j}
    \right)^2,\qquad j\in\mathcal{C}.
\end{equation}
The reconstruction model is trained using the mean squared error over continuous features:
\begin{equation}
    \mathcal{L}_{\mathrm{rec}}
    =
    \frac{1}{|\mathcal{C}|}
    \sum_{j\in\mathcal{C}}
    \frac{1}{N}
    \sum_{i=1}^{N}
    e_{i,j}.
\end{equation}

\subsubsection{Target-error modeling and per-feature control}
\label{subsubsec:per_feature_control}

To generate pseudo-anomalous windows that deviate from the normal manifold without becoming excessively distorted, this paper does not directly impose arbitrary perturbations on the raw signal. Instead, a target error interval is specified for each continuous feature in the reconstruction-error space. Specifically, the reconstruction-error distribution on the training set is used to construct target intervals \(\tau_j^{\mathrm{low}}=Q_{q_l}\left(\{e_{i,j}^{\mathrm{train}}\}_i\right)\) and \(\tau_j^{\mathrm{high}}=Q_{q_u}\left(\{e_{i,j}^{\mathrm{train}}\}_i\right)\), where \(Q_q(\cdot)\) denotes the \(q\)-th quantile. A target error is then sampled for each window and each edited feature as \(\tau_{i,j}=\tau_j^{\mathrm{low}}+u_{i,j}\left(\tau_j^{\mathrm{high}}-\tau_j^{\mathrm{low}}\right)\), where \(u_{i,j}\in[0,1]\) can be obtained by uniform sampling, Beta sampling, or grid sampling. This design is mainly motivated by an empirical phenomenon observed in bearing data: the upper tail of the reconstruction-error distribution for normal training windows is often closer to the boundary-strength range required for subsequent pseudo-anomaly generation. Therefore, the higher-quantile interval of reconstruction errors on normal training samples in Stage~1 can serve as an empirical reference for the target error strength of pseudo-anomalies. It should be emphasized that this reference is derived solely from the normal training data, does not rely on anomaly labels, and does not constitute the final anomaly decision threshold.

As shown in the middle part of Fig.~\ref{fig:stage1}, we further adopt a shared per-feature actor--critic controller to adaptively adjust the editing step size for each continuous feature. Let \(\mathcal{C}_e\subseteq\mathcal{C}\) denote the subset of features being edited in the current window. At the \(r\)-th iteration, the state vector corresponding to the \(j\)-th feature of window \(i\) is defined as
\begin{equation}
    \mathbf{s}^{(r)}_{i,j}
    =
    \big[
    e^{(r)}_{i,j},
    \tau_{i,j},
    e^{(r)}_{i,j}-\tau_{i,j},
    \delta^{(r)}_{i,j},
    d^{(r)}_{i,j},
    r/R,
    \bar{e}^{(r)}_{i},
    \mathrm{std}(\mathbf{e}^{(r)}_{i})
    \big],
\end{equation}
where \(\delta^{(r)}_{i,j}\) is the mean absolute reconstruction residual, \(d^{(r)}_{i,j}\in\{-1,+1\}\) is the current editing direction, \(R\) is the maximum number of iterations, and \(\bar{e}^{(r)}_{i}\) and \(\mathrm{std}(\mathbf{e}^{(r)}_{i})\) are the mean and standard deviation of the per-feature errors within the current window, respectively. All features share the same actor parameters so that the control policy remains consistent and transferable across different sensor dimensions.

Given the state \(\mathbf{s}^{(r)}_{i,j}\), the actor outputs a step-size multiplier \(m^{(r)}_{i,j}=m_{\min}+\left(m_{\max}-m_{\min}\right)\pi_{\phi}\big(\mathbf{s}^{(r)}_{i,j}\big)\), which is then used to obtain the actual update step \(\eta^{(r)}_{i,j}=\mathrm{clip}\left(\eta_{0}m^{(r)}_{i,j},\eta_{\min},\eta_{\max}\right)\), where \(\eta_0\) is the base step size. The editing direction is determined by the relative position between the current error and the target error:
\begin{equation}
    d^{(r)}_{i,j}
    =
    \begin{cases}
        +1, & e^{(r)}_{i,j}\le \tau_{i,j},\\
        -1, & e^{(r)}_{i,j}>\tau_{i,j}.
    \end{cases}
\end{equation}
The update applied to the \(j\)-th continuous feature in the original window is therefore
\begin{equation}
    \tilde{x}^{(r+1)}_{i,t,j}
    =
    \tilde{x}^{(r)}_{i,t,j}
    +
    \eta^{(r)}_{i,j}d^{(r)}_{i,j}
    \left(
    \tilde{x}^{(r)}_{i,t,j}-\hat{x}^{(r)}_{i,t,j}
    \right).
\end{equation}
If the error crosses the target value before and after the update, a crossing indicator is defined as
\begin{equation}
    c^{(r)}_{i,j}
    =
    \mathbb{I}
    \left[
    \left(e^{(r)}_{i,j}-\tau_{i,j}\right)
    \left(e^{(r+1)}_{i,j}-\tau_{i,j}\right)
    \le 0
    \right].
\end{equation}
The corresponding per-feature reward is defined as
\begin{equation}
    R^{(r)}_{i,j}
    =
    -\left|e^{(r+1)}_{i,j}-\tau_{i,j}\right|
    +\lambda_{\mathrm{old}}
    \left|e^{(r)}_{i,j}-\tau_{i,j}\right|
    +\lambda_{\mathrm{cross}}c^{(r)}_{i,j}.
\end{equation}
This reward encourages the updated error to approach the target interval while preserving an incentive for effective crossing behavior. The controller is trained using experience replay and a target network.

\subsubsection{Pseudo-anomalous window filtering}
\label{subsubsec:pseudo_filtering}

After completing \(R\) iterations, the deviation between the final error and the target error is computed for each candidate pseudo-anomalous window as \(\Delta_{i,j}=\left|e_{i,j}^{(R)}-\tau_{i,j}\right|\).
If the \(j\)-th feature satisfies
\begin{equation}
    \Delta_{i,j}
    \le
    \max\left(
    \epsilon_{\mathrm{abs}},
    \epsilon_{\mathrm{rel}}|\tau_{i,j}|
    \right),
\end{equation}
then that feature is regarded as having hit the target. The hit rate of window \(i\) is further defined as
\begin{equation}
    h_i
    =
    \frac{1}{|\mathcal{C}_e|}
    \sum_{j\in\mathcal{C}_e}
    \mathbb{I}
    \left[
    \Delta_{i,j}
    \le
    \max\left(
    \epsilon_{\mathrm{abs}},
    \epsilon_{\mathrm{rel}}|\tau_{i,j}|
    \right)
    \right].
\end{equation}
Only candidate windows whose hit rate exceeds a threshold are retained as valid pseudo-anomalous samples. To prevent samples from concentrating in a single error-strength region, we further adopt an error-bin balancing strategy to collect pseudo-anomalous windows evenly across different target-error intervals, as shown on the right side of Fig.~\ref{fig:stage1}. The final pseudo-anomalous window set is denoted by \(\mathcal{P}=\{\mathbf{P}_k\}_{k=1}^{N_p}\), where \(\mathbf{P}_k\in\mathbb{R}^{W\times |\mathcal{C}|}\).

\subsection{Stage 2: Normal--pseudo-anomalous contrastive representation learning}
\label{subsec:stage2}

\begin{figure}[t]
    \centering
    \includegraphics[width=\linewidth]{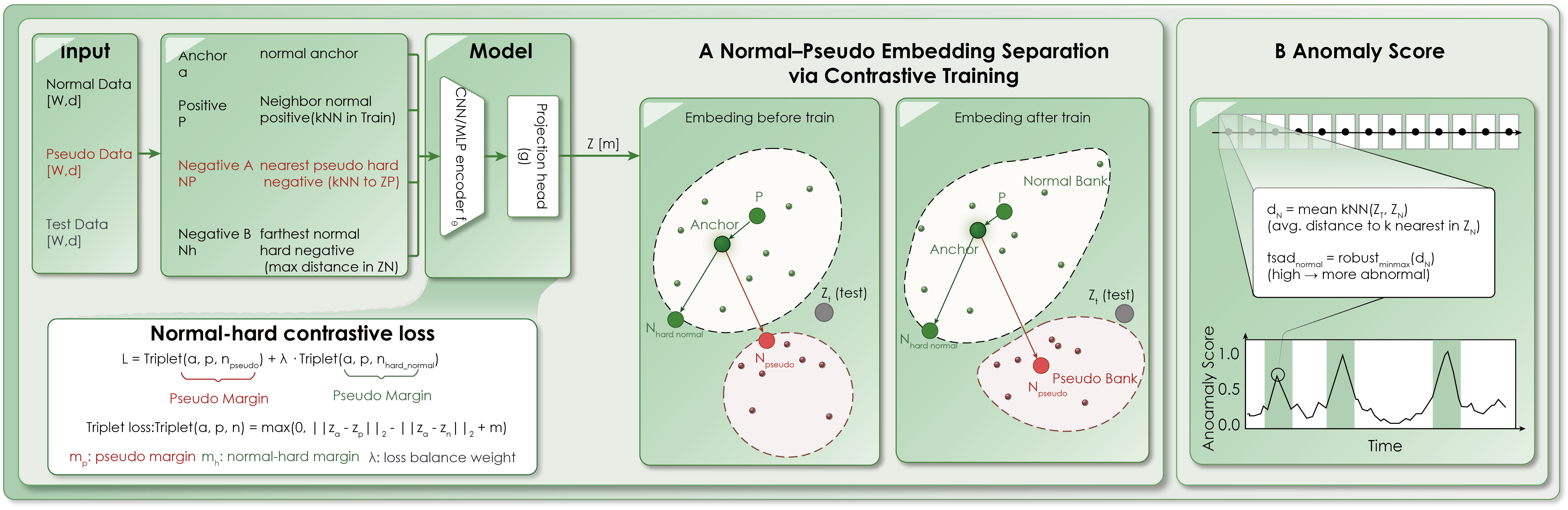}
    \caption{Stage 2 pipeline. The left side shows normal--pseudo-anomalous contrastive training, where normal neighbors are pulled closer while pseudo-anomalous windows and hard normal samples are pushed farther away. The right side shows KNN-based anomaly scoring at test time, where each test window obtains an anomaly score according to its distance to the normal sample bank.}
    \label{fig:stage2}
\end{figure}

As shown in Fig.~\ref{fig:stage2}, Stage 2 uses normal windows and pseudo-anomalous windows generated in Stage 1 to learn an anomaly-sensitive representation space. The left side of the figure corresponds to training-sample construction and the contrastive objective, while the right side corresponds to KNN-based anomaly scoring during testing.

\subsubsection{Training-sample construction and contrastive objective}
\label{subsubsec:contrastive_learning}

Let the normal window bank be \(\mathcal{N}\) and the pseudo-anomalous window bank be \(\mathcal{P}\). For a window \(\mathbf{X}\) in the continuous subspace, its embedding is defined as \(\mathbf{z}=g_{\psi}(f_{\theta}(\mathbf{X}))\in\mathbb{R}^{m}\), where \(f_{\theta}\) is the window encoder and \(g_{\psi}\) is the projection head. In the experiments, a CNN is used as the default encoder.

In each training iteration, an anchor window \(\mathbf{X}^{a}\) is sampled from \(\mathcal{N}\). Its corresponding positive sample \(\mathbf{X}^{+}\) is selected from its normal neighbors, i.e., a nearest-neighbor list is first constructed for normal samples in the raw window space, and then one neighboring window is randomly chosen as the positive sample. The pseudo-anomalous negative sample \(\mathbf{X}^{p-}\) is selected by hard negative mining from the pseudo-anomalous candidate pool as the sample closest to the current anchor in the embedding space. Meanwhile, the farthest normal sample from the current anchor is selected from the normal candidate pool as a normal-hard negative sample \(\mathbf{X}^{h-}\).

Given a triplet \((\mathbf{X}^{a},\mathbf{X}^{+},\mathbf{X}^{-})\), the triplet loss is defined as
\begin{equation}
    \mathcal{L}_{\mathrm{tri}}
    (\mathbf{X}^{a},\mathbf{X}^{+},\mathbf{X}^{-};m)
    =
    \max
    \left(
    0,
    \|\mathbf{z}^{a}-\mathbf{z}^{+}\|_2
    -
    \|\mathbf{z}^{a}-\mathbf{z}^{-}\|_2
    +
    m
    \right).
\end{equation}
The overall contrastive objective consists of a pseudo-anomalous negative term and a normal-hard negative term:
\begin{equation}
    \mathcal{L}_{\mathrm{stage2}}
    =
    \mathcal{L}_{\mathrm{tri}}
    (\mathbf{X}^{a},\mathbf{X}^{+},\mathbf{X}^{p-};m_p)
    +
    \lambda
    \mathcal{L}_{\mathrm{tri}}
    (\mathbf{X}^{a},\mathbf{X}^{+},\mathbf{X}^{h-};m_h),
\end{equation}
where \(m_p\) and \(m_h\) denote the pseudo-anomaly margin and the normal-hard margin, respectively, and \(\lambda\) is a weighting coefficient. This loss keeps normal neighborhoods compact while pushing pseudo-anomalous boundary samples and difficult normal samples away from the anchor, thereby producing a clearer distance structure.

\subsubsection{KNN anomaly scoring and point-level aggregation}
\label{subsubsec:knn_scoring}

After training, normal windows, pseudo-anomalous windows, and test windows are encoded to obtain the embedding sets
\(\mathcal{Z}_{N}\), \(\mathcal{Z}_{P}\), and \(\mathcal{Z}_{T}\), respectively.
For a test-window embedding \(\mathbf{z}^{t}_i\), its average KNN distance to the normal sample bank is defined as
\begin{equation}
    d_{N}(\mathbf{z}^{t}_i)
    =
    \frac{1}{k}
    \sum_{\mathbf{z}\in\mathcal{N}_{k}(\mathbf{z}^{t}_i;\mathcal{Z}_{N})}
    \|\mathbf{z}^{t}_i-\mathbf{z}\|_2,
\end{equation}
and its average KNN distance to the pseudo-anomalous sample bank is defined as
\begin{equation}
    d_{P}(\mathbf{z}^{t}_i)
    =
    \frac{1}{k}
    \sum_{\mathbf{z}\in\mathcal{N}_{k}(\mathbf{z}^{t}_i;\mathcal{Z}_{P})}
    \|\mathbf{z}^{t}_i-\mathbf{z}\|_2.
\end{equation}

By default, this paper adopts a window-level anomaly score based on the distance to the normal sample bank. Let \(d_{N}(\mathbf{z}^{t}_i)\) denote the average KNN distance from the test-window embedding \(\mathbf{z}^{t}_i\) to the normal sample bank. After robust normalization with quantile clipping, the final window-level anomaly score is obtained as
\(S_i=\mathrm{robust\_minmax}\big(d_{N}(\mathbf{z}^{t}_i)\big)\).
Hereafter, \(S_i\) is referred to as the \emph{normal score}. It should be noted that \(\mathrm{robust\_minmax}(\cdot)\) here operates on the anomaly-distance scores output by Stage~2, which is different from the robust scaling used earlier for input-feature preprocessing. As a complement, a relative-distance score can also be defined as \(S^{\mathrm{rel}}_i=\frac{d_{N}(\mathbf{z}^{t}_i)}{d_{N}(\mathbf{z}^{t}_i)+d_{P}(\mathbf{z}^{t}_i)+\epsilon}\), which is referred to below as the \emph{relative score}. In addition, a distance-weighted KNN classifier can be trained on \(\mathcal{Z}_{N}\cup\mathcal{Z}_{P}\) to obtain an auxiliary \emph{OOC score}.

Since the model outputs window-level scores while the evaluation uses point-level labels, the point-level anomaly score is obtained by averaging the scores of all windows covering that time point:
\begin{equation}
    S(t)
    =
    \frac{1}{|\Omega(t)|}
    \sum_{i\in\Omega(t)} S_i,
    \qquad
    \Omega(t)=\{i\mid t_i\le t\le t_i+W-1\}.
\end{equation}

\subsection{Discrete-feature processing and score fusion}
\label{subsec:discrete_knn}

For fully discrete data, the continuous reconstruction and pseudo-anomaly generation of Stage 1 are no longer applicable. In this case, each discrete dimension is first one-hot encoded according to the training-set categories, and each window is flattened into a vector. The anomaly score is then defined by comparing the test window against the bank of normal training windows using KNN:
\begin{equation}
    S_{\mathrm{disc}}(\mathbf{X}^{t}_i)
    =
    \frac{1}{k}
    \sum_{\mathbf{X}\in\mathcal{N}_{k}(\mathbf{X}^{t}_i;\mathcal{N}_{\mathrm{disc}})}
    d(\mathbf{X}^{t}_i,\mathbf{X}),
\end{equation}
where \(d(\cdot,\cdot)\) is Hamming distance by default, but can also be replaced with Euclidean distance.

For mixed-feature data, anomaly scores are computed separately by the continuous branch and the discrete branch. Let the continuous score produced by Stage~2 be \(S_{\mathrm{cont}}\), and let the discrete KNN score be \(S_{\mathrm{disc}}\). After normalizing the two scores to \([0,1]\), denote them by \(\bar{S}_{\mathrm{cont}}=\mathrm{Norm}_{01}(S_{\mathrm{cont}})\) and \(\bar{S}_{\mathrm{disc}}=\mathrm{Norm}_{01}(S_{\mathrm{disc}})\), respectively. Score-level fusion is then performed using \(S_{\mathrm{fuse}}=\max\left(\bar{S}_{\mathrm{cont}},\bar{S}_{\mathrm{disc}}\right)\).
This strategy avoids the combinatorial explosion that would arise from generating pseudo-anomalous samples in the discrete state space, while preserving the ability to jointly sense continuous anomalies and discrete-state anomalies.

\subsection{Evaluation metrics}
\label{subsec:metrics}

The experiments use AUROC, AUPR, and best F1 to evaluate the ranking ability of anomaly scores and the performance of threshold-based detection, and further use VUS-ROC and VUS-PR to evaluate detection robustness at the anomaly-interval level. These metrics can be computed on either window-level or point-level anomaly scores depending on the experimental setting, as detailed in the experimental section. Together, they reflect model performance from three aspects: ranking capability, interval tolerance, and threshold-based detection performance.

\section{Experiments and Results Analysis}
\label{sec:experiments}

This section evaluates the effectiveness of the proposed method from four aspects: bearing fault detection, degradation-process detection, ablation analysis, and hyperparameter sensitivity. The central question is whether the proposed two-stage pseudo anomaly-guided representation learning framework can produce stable anomaly scores on different rotating machinery datasets and maintain sensitivity to anomaly evolution in complex degradation scenarios when trained using only normal samples. Results on public TSAD datasets are presented later in the discussion section as a supplementary analysis of the method's broader applicability. Unless otherwise specified, all experiments use normal samples as the training set, while the test set consists of normal segments and fault/anomalous segments. Labels are used only for metric computation and visualization during testing.

\subsection{Experimental datasets and evaluation settings}
\label{subsec:exp_setup}

\subsubsection{Dataset overview}
\label{subsubsec:dataset_overview}

To systematically evaluate the applicability of the proposed method under different equipment conditions, fault patterns, and temporal scales, we construct an experimental suite centered on bearing fault detection datasets and degradation-process datasets. The fault detection part includes the CWRU (Case Western Reserve University) bearing dataset~\cite{cwru_data}, the HTBF dataset related to high-speed-train bogie faults, the PHM2009 gearbox challenge dataset~\cite{phm2009_data}, and the REALBOX axle-box bearing field-measurement dataset. The degradation-detection part includes the XJTU-SY full-life bearing degradation dataset~\cite{xjtu_sy_data} and the IMS run-to-failure bearing dataset~\cite{ims_data}. In addition, the discussion section later provides an extended analysis on simulated anomaly data and $13$ public time-series datasets from the TSB-AD benchmark to examine the broader applicability of the method when transferred from bearing vibration scenarios to more general TSAD tasks.

The four fault-detection datasets correspond to different validation goals. CWRU was collected on a standard test rig and contains clearly defined fault types with a relatively high signal-to-noise ratio, making it suitable for validating the model's basic discriminative ability on typical bearing faults. In this work, we use the normal baseline and the DE channel from the $12\,\mathrm{kHz}$ drive-end and fan-end fault files, covering ball, inner-race, and outer-race fault states; the $48\,\mathrm{kHz}$ drive-end data are not used. HTBF comes from a mechanical system related to high-speed-train bogies and contains gearbox faults, axle-box faults, and compound fault combinations under multiple operating conditions. We use five synchronized vibration channels to assess model stability in multi-channel and multi-fault-combination scenarios. PHM2009 is based on a generic industrial gearbox. We use the two acceleration channels at the input and output ends and exclude the tachometer channel; spur 1 and helical 1 are treated as normal classes, while spur 2--8 and helical 2--6 are treated as fault classes. REALBOX consists of measured axial acceleration vibration records from an axle-box bearing of a high-speed train, sampled at $10\,\mathrm{kHz}$, with an inner-race failure as the fault type. Because measured fault samples are limited, this dataset is mainly used to verify the applicability of the method in small-sample real engineering scenarios.

Unlike fixed fault-category detection, degradation-process detection involves anomalous states that usually do not appear abruptly, but instead evolve gradually over a long period. The XJTU-SY dataset contains complete run-to-failure sequences for $15$ bearings under $3$ speed/load conditions, with both horizontal and vertical vibration channels recorded at each acquisition. The IMS dataset contains run-to-failure bearing sequences over an even longer time scale, making it suitable for examining the model's response to weak degradation, long transitional states, and late-stage failure. We do not interpret degradation-detection results as remaining useful life prediction; instead, anomaly scores are viewed as the degree to which test windows deviate from the early-life normal training distribution.

\subsubsection{Construction of fault-detection datasets}
\label{subsubsec:fault_dataset_construction}

To conform to the one-class anomaly-detection setting, all fault-detection experiments train the model using only normal-state samples, while fault samples are used only for evaluation at test time. Rather than directly using the complete raw sequences from the public datasets, we extract fixed-length segments from the raw files according to a unified protocol to construct a normal training segment and a normal/fault test sequence. Specifically, a normal training segment $X_{\mathrm{tr}}^{\mathrm{N}}$ is first extracted from the normal-state data. This segment is used for Stage~1 reconstruction-model training, Stage~2 representation-encoder training, and the construction of the KNN normal reference bank. The remaining normal test segment is then concatenated directly with fault segments along the time dimension to form a long test sequence. For multi-channel data, the channel dimension remains unchanged during concatenation, and only the time dimension is extended.

Let the retained normal test segment be $X_{\mathrm{te}}^{\mathrm{N}}$, and let the $m$-th fault segment be $X_{\mathrm{te}}^{\mathrm{F},m}$. The test sequence for fault detection is then constructed as
\begin{equation}
    X_{\mathrm{te}} =
    [X_{\mathrm{te}}^{\mathrm{N}};
    X_{\mathrm{te}}^{\mathrm{F},1};
    X_{\mathrm{te}}^{\mathrm{F},2};
    \cdots;
    X_{\mathrm{te}}^{\mathrm{F},M}],
\end{equation}
where $[\cdot;\cdot]$ denotes concatenation along the time dimension. The corresponding point-level labels are defined as
\begin{equation}
    y_t =
    \begin{cases}
        0, & x_t \in X_{\mathrm{te}}^{\mathrm{N}},\\
        1, & x_t \in X_{\mathrm{te}}^{\mathrm{F},m}.
    \end{cases}
\end{equation}
The normal test segment participates in the final computation of AUROC, AUPR, best F1, Precision, and Recall as the negative class, but it is not used for parameter updates in Stage~1 or Stage~2 and is not added to the KNN normal reference bank. The construction of test sequences for each fault-detection dataset is summarized in Table~\ref{tab:fault_test_construction}.

\begin{table}[H]
\centering
\caption{Construction of test sequences in the fault-detection experiments}
\label{tab:fault_test_construction}
\footnotesize
\begingroup
\setlength{\tabcolsep}{5pt}
\renewcommand{\arraystretch}{1.12}
\begin{threeparttable}
\begin{tabularx}{\textwidth}{@{}L{1.9cm}C{2.1cm}C{2.1cm}Y@{}}
\toprule
Dataset & \makecell[c]{Normal segments\\($10^{4}$ points)} & \makecell[c]{Fault/anomalous segments\\($10^{4}$ points)} & Description of test-sequence construction \\
\midrule
CWRU
& $4\times 5.0$
& $105\times 2.0$
& The first $20.0\times 10^{4}$ sampling points form the normal test segment, followed by $105$ fault segments covering ball, inner-race, and outer-race faults. \\

HTBF
& $3\times 15.0$
& $84\times 1.0$
& The front part consists of normal test segments under three operating conditions, followed by $84$ fault combinations; the five vibration channels remain time-aligned. \\

PHM2009
& $2\times 10.0$
& $40.0$
& helical~1 and spur~1 form the normal test segment; the anomalous part is obtained by approximately balanced concatenation of $12$ fault classes from helical~2--6 and spur~2--8. \\

REALBOX
& $7.0+12.0+1.0$
& $40.0$
& The first $20.0\times 10^{4}$ sampling points are formed by three normal fragments, followed by an inner-race fault fragment; only samples in the speed interval $295$--$305$ are retained during construction. \\
\bottomrule
\end{tabularx}
\begin{tablenotes}[flushleft]
\footnotesize
\item Note: The values for normal and fault/anomalous segments in the table are measured in units of $10^{4}$ sampling points. For example, $4\times 5.0$ indicates four fragments, each with length $5.0\times 10^{4}$. The test fragments are concatenated sequentially along the time dimension only for presentation and for plotting continuous anomaly-score curves. During window construction, however, each original fragment is segmented separately, and no windows are created across concatenation boundaries between different fragments. Window labels are determined by the point-level labels of the underlying fragments. CWRU uses the DE single channel, while the other multi-channel datasets keep channel synchronization during concatenation.
\end{tablenotes}
\end{threeparttable}
\endgroup
\end{table}

The above test sequences are concatenated sequentially along the time dimension only for description and result visualization. No additional padding, smoothing segments, or transition segments are inserted. This design facilitates unified presentation of the temporal relationship between the normal segment and the subsequent fault segments, as well as continuous plotting of anomaly-score curves.

\subsubsection{Construction of degradation-detection datasets}
\label{subsubsec:degradation_dataset_construction}

For the XJTU-SY and IMS degradation datasets, the test sequences preserve the original temporal continuity of the full-life process and do not artificially concatenate normal files with fault files. The model uses only early-life snapshots as the normal training segment, and the remaining snapshots are treated as the test segment. The degradation onset is automatically determined using snapshot-level vibration-amplitude statistics. Specifically, statistics such as RMS, p95, and maximum amplitude are computed for each snapshot, compared by ratio against statistics from the early normal segment, and an abnormal increase is required to persist for at least $3$ snapshots (sustain$=3$) to reduce the influence of isolated shocks on onset detection. Sampling points after the degradation onset are labeled as anomalous, while test samples before the onset that do not belong to the training segment are labeled as normal.

The XJTU-SY samples used in this paper include Bearing1-1, Bearing1-3, Bearing2-2, Bearing2-5, Bearing3-3, and Bearing3-5, covering three operating conditions: $35$ Hz/$12$ kN, $37.5$ Hz/$11$ kN, and $40$ Hz/$10$ kN. Except for Bearing3-5, which uses the first $23$ snapshots as the training segment due to an earlier detected degradation onset, all other XJTU-SY samples use the first $30$ snapshots for training. For IMS, Bearing4 from the 1st-test is used, with the first $200$ snapshots for training and the remaining snapshots for testing. The training lengths and degradation-onset settings are listed in Table~\ref{tab:degradation_onset_settings}.

\begin{table}[H]
\centering
\small
\caption{Training-segment and degradation-onset settings for degradation detection}
\label{tab:degradation_onset_settings}
\begingroup
\setlength{\tabcolsep}{6pt}
\renewcommand{\arraystretch}{1.08}
\begin{threeparttable}
\begin{tabular}{llccc}
\toprule
Dataset & Sample & Number of training snapshots & Total snapshots & Degradation onset \\
\midrule
\multirow{6}{*}{XJTU-SY} & Bearing1-1 & 30 & 123 & 77 \\
 & Bearing1-3 & 30 & 158 & 89 \\
 & Bearing2-2 & 30 & 161 & 49 \\
 & Bearing2-5 & 30 & 339 & 182 \\
 & Bearing3-3 & 30 & 371 & 343 \\
 & Bearing3-5 & 23 & 114 & 24 \\
\midrule
IMS & Bearing4 & 200 & 2156 & 1436 \\
\bottomrule
\end{tabular}
\begin{tablenotes}[flushleft]
\footnotesize
\item Note: Sampling points after the degradation onset are labeled as anomalous, and the window label is determined by whether the window contains anomalous points. For Bearing3-5, the training segment is restricted to the period before the degradation onset because the onset is detected early.
\end{tablenotes}
\end{threeparttable}
\endgroup
\end{table}

\subsubsection{Window segmentation, label generation, and normalization}
\label{subsubsec:window_label_scaling}

All sequences are segmented using sliding windows with length $L=512$ and stride $H=256$. The $j$-th window is defined as
\begin{equation}
    W_j = X_{s_j:s_j+L-1,:}, \quad s_j=jH .
\end{equation}
Sampling points at the tail that do not form a complete window are discarded. Window-level labels are generated using a conservative rule: if any time point within the window belongs to an anomalous interval, that window is labeled as anomalous, i.e.,
\begin{equation}
    Y_j =
    \mathbb{I}\left(
    \sum_{t=s_j}^{s_j+L-1} y_t > 0
    \right).
\end{equation}
Therefore, a window that crosses a normal/fault concatenation boundary or a degradation-onset boundary is labeled anomalous as long as it contains any anomalous point. All main experimental metrics are computed on window-level anomaly scores and window-level labels.

To avoid leakage of test information, normalization statistics for continuous features are estimated only from the normal training segment and then fixed for the test sequence and for baseline methods. Multi-channel data are normalized channel-wise. Stage~2 applies the same input transformation using the training-normal statistics saved from Stage~1. Baseline methods likewise estimate normalization parameters from the normal training windows and compute anomaly scores on the same test windows. For the window-energy heatmaps shown in the figures, we compute the RMS energy for each window and feature channel as
\begin{equation}
    r_{i,c}=\sqrt{\frac{1}{L}\sum_{t=1}^{L}x_{i+t,c}^{2}},
\end{equation}
and then use the $5$th and $99$th percentiles of the RMS energy of the same channel in the training windows as lower and upper bounds to robustly scale the test-window values $r_{i,c}$ and clip them to $[0,1]$. This quantity is denoted as scaled RMS energy in the figures and is used only to visualize input-energy changes across windows and channels; it does not participate in Stage~2 KNN anomaly scoring and is not included in the final evaluation metrics.

\subsubsection{Training settings and pseudo-anomalous samples}
\label{subsubsec:training_pseudo_settings}

The proposed method consists of two training stages. Stage~1 trains a reconstruction model on normal training windows to generate pseudo-anomalous windows with controllable deviation strength. Stage~2 uses the normal training windows together with the pseudo-anomalous windows generated in Stage~1 to train the representation encoder and compute anomaly scores in the embedding space using a normal reference bank. Unless otherwise specified, the Stage~1 reconstruction model is trained for $12$ epochs, including $4$ epochs of pretraining for the continuous branch, and the Stage~2 representation encoder is trained for $12$ epochs. In the main experiments, the \emph{normal score} with $k=5$ (i.e., the anomaly score based on the KNN distance to the normal sample bank) is used as the window-level anomaly score. The KNN normal reference bank consists only of embeddings of normal training windows and does not include normal test windows or fault windows.

Pseudo-anomalous windows are generated only from normal training windows; no fault test samples or test labels are used. The number of pseudo-anomalous windows for each dataset is jointly determined by Stage~1 generation and error-interval filtering. For CWRU, HTBF, and REALBOX, pseudo-anomalous samples are balanced across five error bins, with $2400$ windows retained per bin, yielding a total of $12000$ pseudo-anomalous windows. For PHM2009, the acceptable sample numbers after error-bin filtering are $866$, $890$, $970$, $992$, and $1006$ in the five bins, respectively, so the final number of retained pseudo-anomalous windows is $4724$. For the XJTU-SY and IMS degradation experiments, $6500$ pseudo-anomalous windows are retained in each case. Table~\ref{tab:dataset_settings_revised} summarizes the main data scales and training configurations for all datasets.

\begin{table}[H]
\centering
\scriptsize
\caption{Experimental scale and main configurations of each dataset}
\label{tab:dataset_settings_revised}
\begingroup
\setlength{\tabcolsep}{3pt}
\renewcommand{\arraystretch}{1.06}
\begin{threeparttable}
\resizebox{\linewidth}{!}{
\begin{tabular}{@{}llcccccc@{}}
\toprule
Dataset & Sample/subset & Dimension index & \makecell[c]{Training\\sequence length} & \makecell[c]{Test\\sequence length} & \makecell[c]{Number of\\training windows} & \makecell[c]{Number of\\test windows} & \makecell[c]{Number of\\pseudo-anomalous windows} \\
\midrule
CWRU & -- & [1] & $400000$ & $2300000$ & $1561$ & $8983$ & $12000$ \\
HTBF & -- & \makecell[c]{[1,2,3,\\4,5]} & $900000$ & $1290000$ & $3514$ & $5038$ & $12000$ \\
PHM2009 & -- & [1,2] & $400000$ & $600000$ & $1561$ & $2342$ & $4724$ \\
REALBOX & -- & [1] & $400000$ & $600000$ & $1561$ & $2342$ & $12000$ \\
XJTU-SY & Bearing1-1 & [1,2] & $983040$ & $3047424$ & $3839$ & $11903$ & $6500$ \\
IMS & Bearing4 & [1,2] & $4096000$ & $40058880$ & $15999$ & $156479$ & $6500$ \\
\bottomrule
\end{tabular}}
\begin{tablenotes}[flushleft]
\footnotesize
\item Note: The dimension index indicates the channel numbers used for modeling; for example, [1,2] means that Channels 1 and 2 are used. Unless otherwise specified, all experiments use sliding windows with length $L=512$ and stride $H=256$.
\end{tablenotes}
\end{threeparttable}
\endgroup
\end{table}

\subsubsection{Evaluation metrics and visualization}
\label{subsubsec:metrics_visualization_revised}

The quantitative evaluation metrics include AUROC, AUPR, best F1, Precision, and Recall. Among them, AUROC and AUPR measure the ranking ability of anomaly scores on normal and anomalous windows and do not depend on a fixed threshold. Best F1 is obtained by sweeping thresholds over the anomaly scores of the test windows, and Precision and Recall are reported at the threshold that maximizes F1. Therefore, best F1 and its corresponding Precision/Recall reflect the upper-bound thresholding performance under the given test labels, rather than the performance of a fixed threshold determined in advance for unsupervised deployment. For the TSAD extension analysis in the later discussion, VUS-ROC and VUS-PR~\cite{boniol2025vus} are further reported to characterize detection quality in the neighborhood of anomalous intervals.

In addition to quantitative metrics, we conduct qualitative analysis using anomaly-score curves and representation-space visualization. Anomaly-score curves are used to observe the temporal response of the model to fault segments, degradation stages, or anomalous intervals. t-SNE or PCA visualization is used to analyze the relative distributions of normal training windows, normal test windows, pseudo-anomalous windows, and real anomalous windows in the representation space. The fault-detection result figures follow a unified layout: Part A presents the scaled RMS energy heatmap together with the anomaly-score curves of the proposed method and baseline methods, where light background shading indicates anomalous intervals and dashed lines denote the posterior best-F1 threshold obtained by scanning on the test-window labels, used only to facilitate interpretation of the anomaly-score curves; Part B presents radar charts for AUROC, AUPR, Precision, Recall, and best F1; and Part C shows representation-space visualization. It should be emphasized again that normal test windows participate in final metric evaluation but are not used for model training and are not added to the KNN normal reference bank.

To ensure consistency across methods, the bearing fault-detection and degradation-detection experiments use fixed data splits, fixed window settings, and a fixed random seed of $42$. Deterministic baseline methods are reported directly under this fixed protocol. For deep models involving random initialization or random sampling, the random seed is fixed to ensure reproducibility. Repeated runs are used only to check implementation stability and are not treated as independent repeated trials. All baseline methods use the same training windows, test windows, and window-level labels as the proposed method and are evaluated under the same protocol using AUROC, AUPR, best F1, Precision, and Recall.

\subsubsection{Computational cost}
To complement the implementation analysis, we further report the inference cost of each method for completing one full scoring pass over all test windows on CWRU, as shown in Table~\ref{tab:cwru_inference_cost}. This experiment scores all $8983$ test windows once, and training/fitting time as well as reference-bank construction cost are excluded for all methods. For the proposed method, only the online inference of Stage~2 is counted, i.e., test-window encoding and KNN distance computation against the normal sample bank; Stage~1 pseudo-anomaly generation is excluded. On CWRU, the proposed method requires only $0.113\,\mathrm{s}$ on average to complete one full scoring pass over all test windows, corresponding to $0.0126\,\mathrm{ms/window}$, with peak GPU memory of about $169.3\,\mathrm{MB}$. Compared with other methods, our method is clearly faster than heavier deep sequence baselines such as Adjacent Transformer, Transformer AE, and TranAD, but slower than KNN Distance, LOF Novelty, and Deep SVDD, which do not require reconstruction decoding or use lighter scoring schemes. This indicates that even after introducing encoder forward passes and KNN retrieval against the normal bank, the online inference cost of the proposed method remains within an acceptable range.

\begin{table}[H]
    \centering
    \caption{Comparison of inference cost for one complete test-window scoring pass on CWRU. The statistics are based on scoring all $8983$ test windows once; for the proposed method, only Stage~2 inference is counted and Stage~1 pseudo-anomaly generation is excluded; training/fitting and reference-bank construction costs are not included for any method. (3080ti)}
    \label{tab:cwru_inference_cost}
    \small
    \renewcommand{\arraystretch}{1.08}
    \setlength{\tabcolsep}{4pt}
    \resizebox{\linewidth}{!}{
    \begin{tabular}{lrrrrrr}
        \toprule
        Method & Model params & Reference bank size & Test windows & Inference time (s) & ms/window & Peak GPU mem (MB) \\
        \midrule
        Deep SVDD & 213,568 & 0 & 8,983 & 0.044 & 0.0049 & 13.5 \\
        KNN Distance & 0 & 1,561 & 8,983 & 0.062 & 0.0069 & 0.0 \\
        LOF Novelty & 0 & 1,561 & 8,983 & 0.066 & 0.0073 & 0.0 \\
        \textbf{TPA-AD (Stage~2 only)} & \textbf{123,840} & \textbf{1,561} & \textbf{8,983} & \textbf{0.113} & \textbf{0.0126} & \textbf{169.3} \\
        Isolation Forest & 0 & 1,561 & 8,983 & 0.226 & 0.0251 & 0.0 \\
        One-class SVM & 0 & 1,561 & 8,983 & 0.360 & 0.0401 & 0.0 \\
        TranAD & 401 & 0 & 8,983 & 0.493 & 0.0549 & 302.3 \\
        Transformer AE & 67,265 & 0 & 8,983 & 0.548 & 0.0610 & 141.9 \\
        Adjacent Transformer & 67,265 & 0 & 8,983 & 0.961 & 0.1070 & 1261.4 \\
        \bottomrule
\end{tabular}}
\end{table}

\FloatBarrier

\subsection{Bearing fault detection experiments}
\label{subsec:bearing_fault_detection}

This subsection evaluates the proposed method on four datasets, namely CWRU, HTBF, PHM2009, and REALBOX, to assess its discriminative ability on fault samples. The analysis focuses on two issues: first, whether anomaly scores can form a stable and clear boundary between normal and fault segments; and second, whether the representation space learned in Stage~2 can effectively push real fault samples away from the normal region. The data presentation and detection results are discussed below on a dataset-by-dataset basis.

\subsubsection{CWRU dataset}
\label{subsubsec:cw_result}
\begin{figure}[H]
    \centering
    \includegraphics[width=0.95\linewidth]{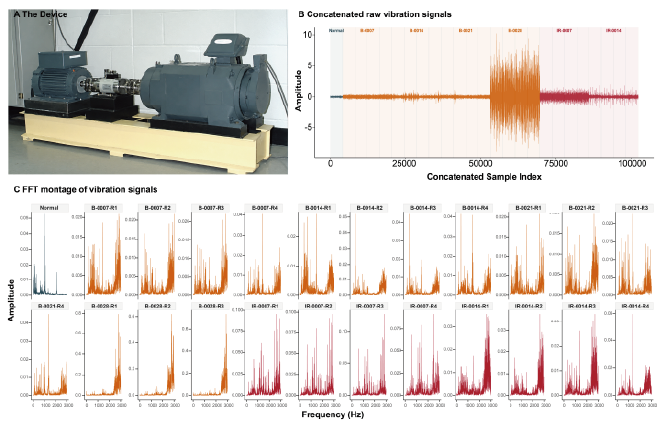}
    \caption{Illustration of the CWRU dataset. A shows the experimental device, B shows the long sequence obtained by concatenating samples from different classes, and C shows the corresponding FFT spectra. The figure highlights the time-domain and frequency-domain differences between normal samples and different fault types.}
    \label{fig:cw_data}
\end{figure}
The CWRU dataset was released by the Case Western Reserve University Bearing Data Center~\cite{cwru_data}. Its test rig consists of a $2$ hp motor, a torque transducer/encoder, and a dynamometer. Faults of different sizes were introduced by electro-discharge machining on the drive-end and fan-end bearings, and vibration signals were collected under loads of $0$--$3$ hp. The original repository provides both $12\,\mathrm{kHz}$ and $48\,\mathrm{kHz}$ sampled data and covers multiple fault locations such as ball, inner-race, and outer-race faults. In this work, we use the normal baseline and the DE channel from the $12\,\mathrm{kHz}$ drive-end and fan-end fault files, covering ball, inner-race, and outer-race fault states, while the $48\,\mathrm{kHz}$ drive-end data are not used. A training set is constructed from normal segments, and the remaining normal segments are concatenated with fault segments to form the test sequence. Figure~\ref{fig:cw_data} shows the experimental setup, the concatenated long-horizon sequence, and its FFT spectrum. As can be seen, normal and fault samples differ clearly in both time-domain amplitude and frequency-domain structure, making CWRU a benchmark dataset for validating basic fault separability.

\begin{figure}[H]
    \centering
    \includegraphics[width=0.98\linewidth]{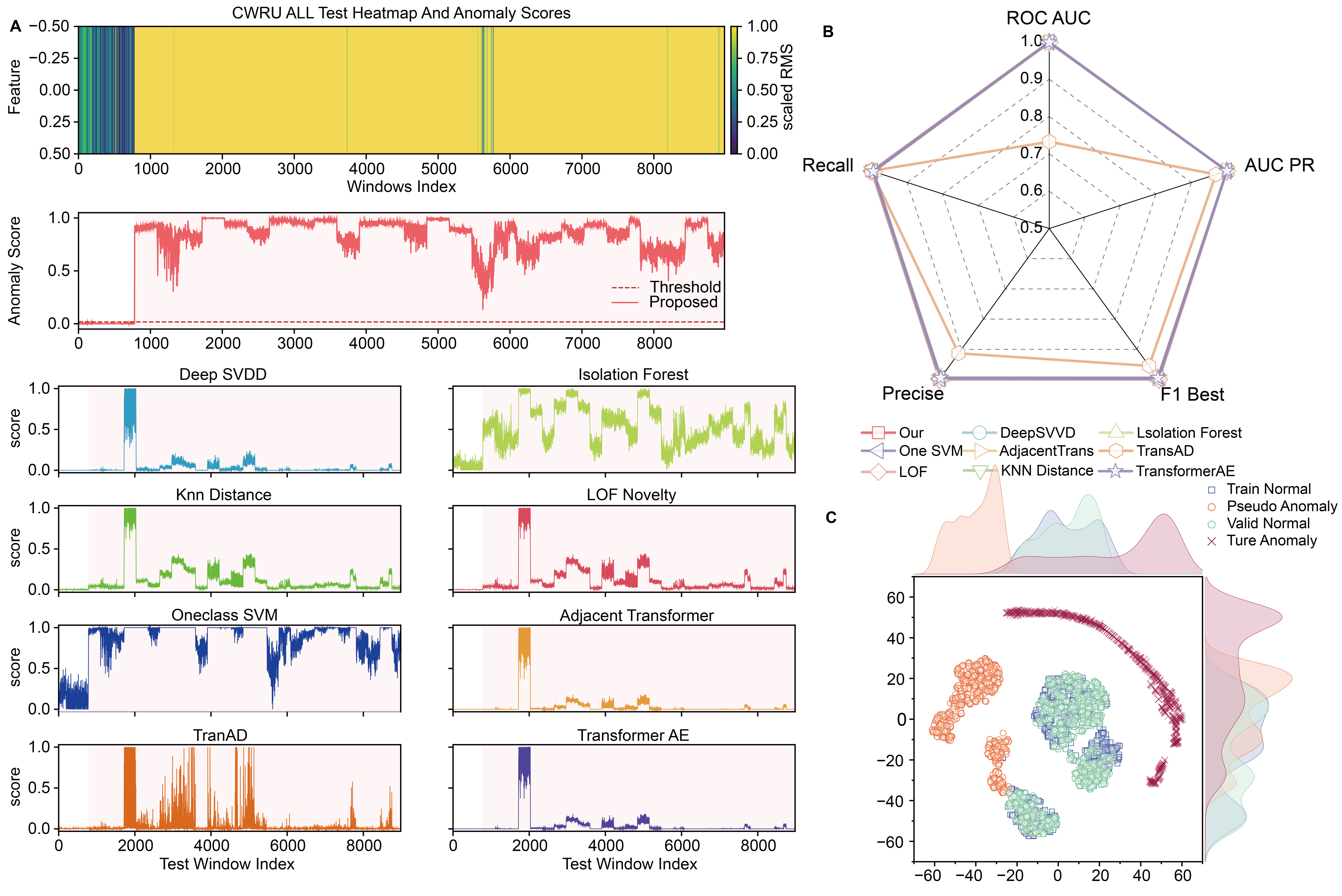}
    \caption{Fault-detection results on the CWRU dataset. A shows the window-energy heatmap together with the anomaly-score curves of the proposed method and baseline methods; B compares AUROC, AUPR, Precision, Recall, and best F1 across methods; and C visualizes the low-dimensional distributions of normal training windows, normal test windows, pseudo-anomalous windows, and real anomalous windows.}
    \label{fig:cw_result}
\end{figure}

Figure~\ref{fig:cw_result} shows that the proposed method maintains anomaly scores close to zero during the normal stage of the CWRU test sequence, rapidly crosses the threshold when the fault segment begins, and then remains in a high-score region for most of the remaining sequence, with only brief local drops around a few mixed windows. Part A further shows that the jump in the proposed anomaly score is broadly aligned with the entrance into the high-energy region in the heatmap, thereby producing a clear stage boundary among the normal segment, transition segment, and fault segment. By contrast, Deep SVDD, KNN Distance, LOF Novelty, Adjacent Transformer, and Transformer AE all respond strongly around the first fault injection, but their scores later fall back close to zero for many fault windows, behaving more like instantaneous amplification of local transients than stable characterization of persistent abnormal states. One-class SVM maintains an elevated background score over a long interval, which weakens threshold interpretability, while TranAD mainly produces isolated spikes and lacks continuity. Therefore, in this relatively ideal test-rig scenario, the main advantage of the proposed method is not simply generating an earlier spike at a single point, but rather integrating local anomaly evidence into a sustained high-score plateau, thereby balancing low false alarms with a clear stage boundary. The radar chart in Part B is almost saturated along the outer ring, indicating that on strongly separable data such as CWRU, most methods can already achieve very high AUROC, AUPR, and Recall. Hence, the key issue here is not whether a method can detect anomalies at all, but whether it can maintain more stable Precision and best F1 while preserving high Recall. Although the outward expansion of the proposed method on the radar chart is only slightly better than that of several baselines, this slight advantage is consistent with the smoother high-score plateau observed in Part A. The t-SNE visualization in Part C provides more interpretable evidence: normal training windows and normal test windows overlap heavily in the center, pseudo-anomalous windows are mainly distributed along the outer edge of the normal cluster, and real anomalous windows form an independent arc-shaped region far from the center. This structure suggests that the pseudo-anomalous windows generated in Stage~1 do not need to coincide point by point with the real fault cluster; rather, they act more like a learnable expanded boundary around the normal region, thereby providing an effective directional constraint for the discriminative representation learning in Stage~2. It should also be noted that CWRU is a relatively ideal laboratory dataset, so this result mainly serves to validate the effectiveness of the method on typical bearing-fault scenarios.

\subsubsection{HTBF dataset}
\label{subsubsec:htbf_result}

\begin{figure}[H]
    \centering
    \includegraphics[width=0.95\linewidth]{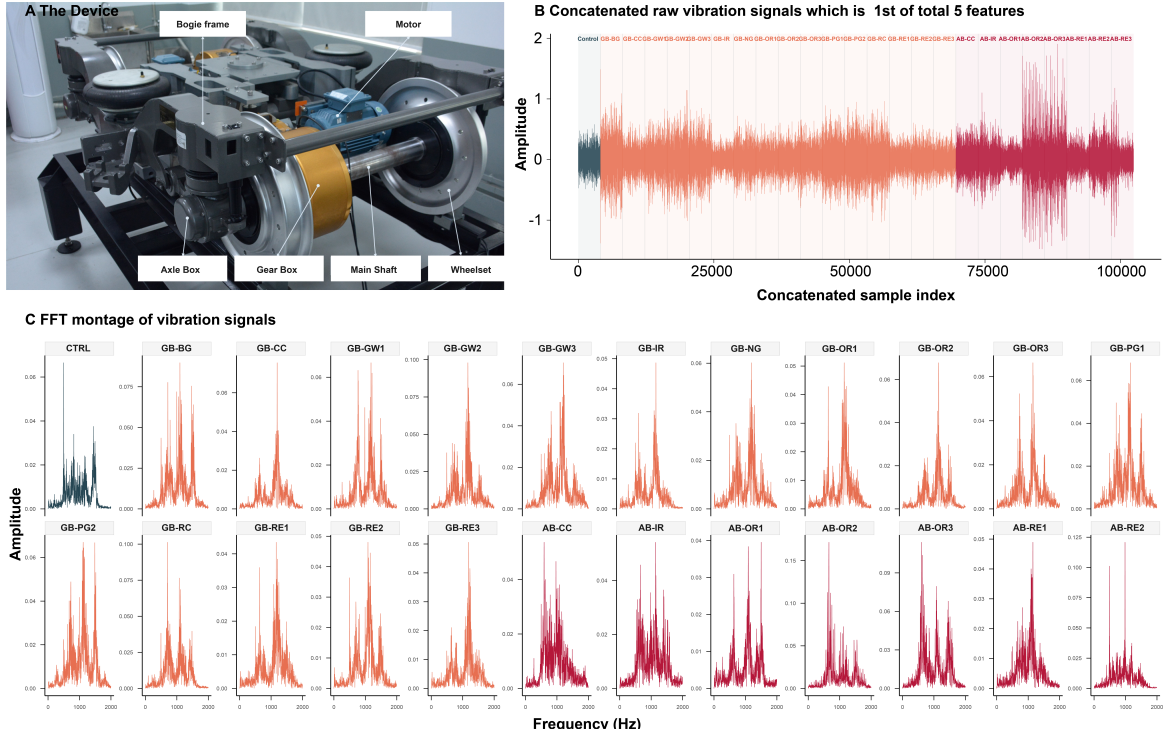}
    \caption{Illustration of the HTBF dataset. A shows the experimental setup, B shows the long sequence obtained by concatenating samples from different classes, and C shows the corresponding FFT spectra. This dataset contains high-speed-train-bogie-related vibration signals under multiple operating conditions and is used to evaluate the model's detection ability in multi-channel, multi-fault-category scenarios.}
    \label{fig:htbf_data}
\end{figure}
The HTBF dataset is used to evaluate performance on a mechanical system related to high-speed-train bogies. It is built around the bogie drivetrain and covers key components such as the gearbox, axle box, main shaft, and wheelset. In this work, five synchronized vibration channels are used to construct long sequences. Compared with the single test-rig bearing data in CWRU, HTBF contains both control states and multiple gearbox-/axle-box-related fault categories, with stronger operating-condition disturbance and more pronounced channel coupling. Figure~\ref{fig:htbf_data} shows the experimental setup, the long sequence obtained by concatenating multiple categories, and the corresponding FFT spectra. As the figure indicates, HTBF exhibits not only differences among fault categories but also more significant changes in operating conditions and channels, making it a more challenging detection scenario than single-rig bearing data.

\begin{figure}[H]
    \centering
    \includegraphics[width=0.98\linewidth]{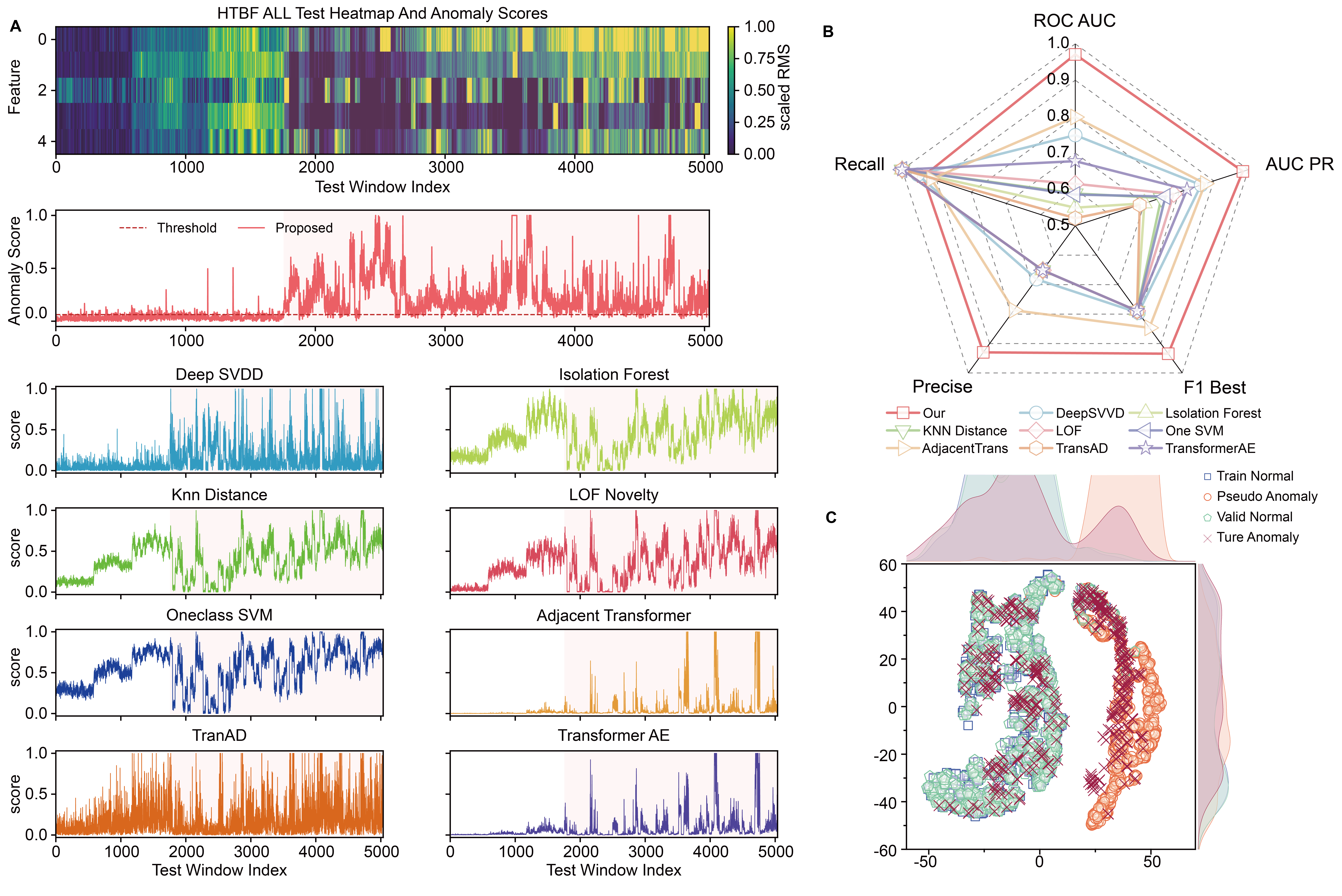}
    \caption{Fault-detection results on the HTBF dataset. A shows the window-energy heatmap and the anomaly-score curves of different methods, B shows the multi-metric radar chart, and C shows the representation-space distributions of normal training windows, normal test windows, pseudo-anomalous windows, and real anomalous windows.}
    \label{fig:htbf_result}
\end{figure}

Figure~\ref{fig:htbf_result} shows that the boundary between normal and anomalous samples is weaker on HTBF than on CWRU, yet the proposed method still forms relatively continuous score responses over multiple anomalous intervals. Part A shows that the proposed method maintains a low background score in most of the early normal interval. After about $1800$ windows and during subsequent condition/fault switching intervals, its score rises along with local energy enhancement and remains at a medium-to-high level over clearly anomalous segments. In contrast, Deep SVDD and TranAD exhibit strong fluctuations over long intervals, already producing many isolated high peaks during normal segments. KNN Distance, LOF Novelty, Isolation Forest, and One-class SVM can sense overall distribution shifts, but they are also more likely to map operating-condition switching and channel-amplitude differences to elevated background scores, thereby limiting Precision. Adjacent Transformer and Transformer AE respond more sparsely and often peak only at a few sharp impacts, providing insufficient coverage for weak anomalies spread across wider temporal ranges. In other words, in a multi-channel and strongly perturbed scenario such as HTBF, the main advantage of the proposed method lies in its balance of ``low background + continuous response'': it neither prematurely raises the whole sequence to a high-score level nor relies solely on a few isolated spikes to indicate anomalies. The radar chart in Part B further shows that this advantage mainly comes from simultaneous improvements in Precision, best F1, AUROC, and AUPR, rather than from simply achieving higher Recall. In fact, multiple baselines already approach the outer ring on Recall, indicating that they are not short of the ability to ``flag anomalies.'' The more important difference is whether they also raise many normal windows to high scores. The proposed method keeps Recall high while placing Precision and best F1 farther outward, indicating that its gains mainly come from better false-alarm control. The visualization in Part C is consistent with this observation: normal training windows and normal test windows form a large left-side manifold; pseudo-anomalous windows are distributed more along the right-side outer edge; and real anomalous windows partly lie adjacent to this outer edge, while another part still mixes into the left-side normal manifold. This structure suggests that HTBF indeed contains a subset of pronounced anomalies that can be effectively captured by the pseudo-anomalous boundary, but it also contains weak anomalies or transitional samples that remain close to the normal state. In such a case, the proposed method improves the separability of the major anomalous samples, but cannot yet completely separate all anomalous samples.

\subsubsection{PHM2009 dataset}
\label{subsubsec:phm2009_result}
\begin{figure}[H]
    \centering
    \includegraphics[width=0.95\linewidth]{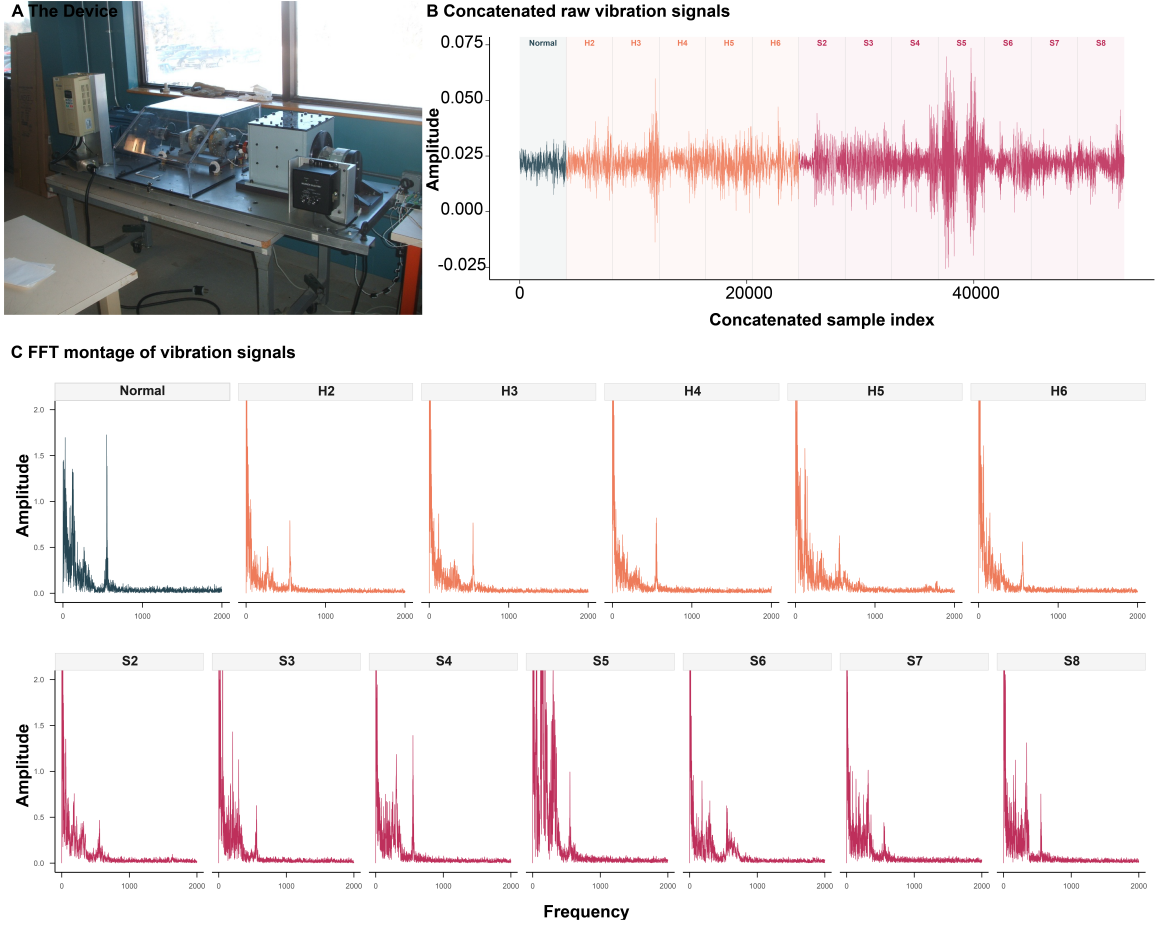}
    \caption{Illustration of the PHM2009 dataset. A shows the experimental setup, B shows the long sequence obtained by concatenating samples from different classes, and C shows the corresponding FFT spectra. The figure illustrates the input characteristics of complex rotating-machinery fault data from a gearbox system.}
    \label{fig:phm2009_data}
\end{figure}
The PHM2009 dataset comes from the 2009 Gearbox Challenge released by the PHM Society~\cite{phm2009_data}. It is based on a generic industrial gearbox and synchronously records two vibration channels on the retaining plates of the input and output shafts as well as a tachometer pulse, covering shaft speeds of $30$--$50$ Hz and both high- and low-load conditions. The faulty objects are no longer limited to a single bearing, but rather to a more complex rotating transmission system. In this paper, the two vibration sequences are recast into a one-class anomaly-detection task to provide supplementary cross-equipment validation. Figure~\ref{fig:phm2009_data} shows the experimental setup, the long sequence obtained by concatenating samples from different states, and the corresponding FFT spectra. Compared with CWRU, some anomalous samples in PHM2009 are more similar to normal samples in both the time and frequency domains, making it useful for testing method robustness under weak-difference fault scenarios.

\begin{figure}[H]
    \centering
    \includegraphics[width=0.98\linewidth]{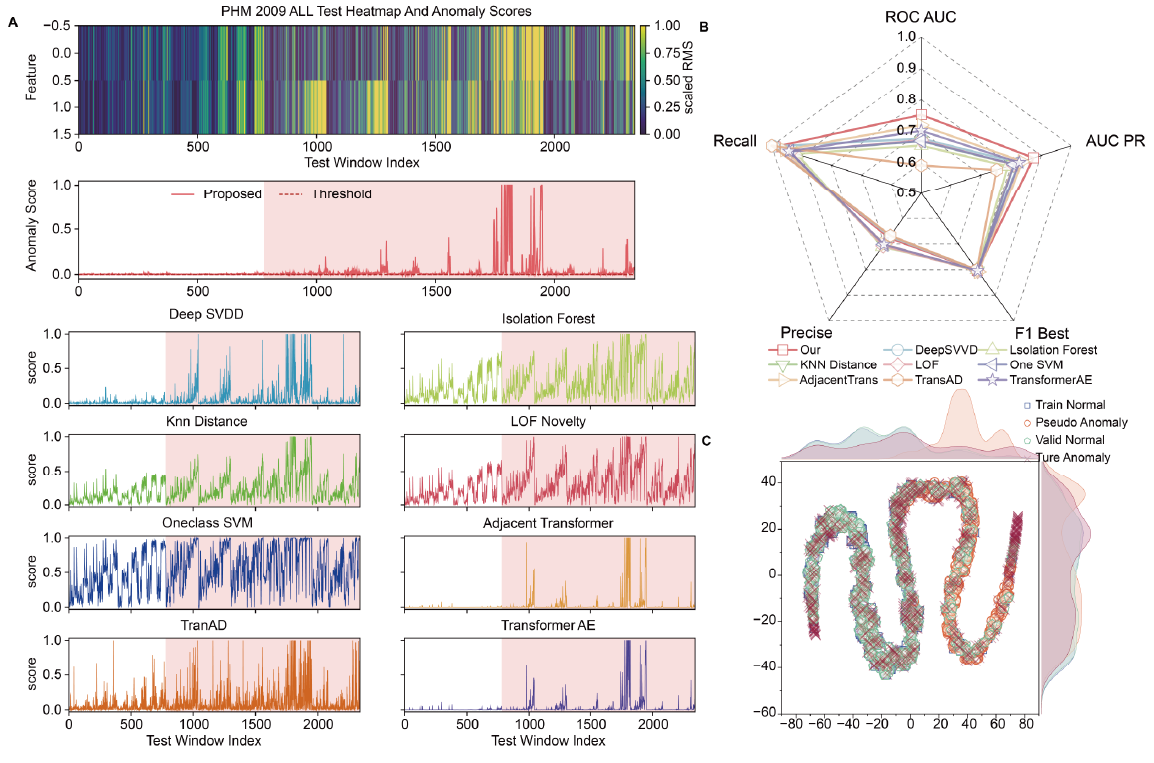}
    \caption{Fault-detection results on the PHM2009 dataset. A shows the window-energy heatmap and the anomaly-score curves of different methods, B shows the multi-metric radar chart, and C shows the representation-space distributions of normal training windows, normal test windows, pseudo-anomalous windows, and real anomalous windows. In this dataset, normal windows and some anomalous windows are highly similar.}
    \label{fig:phm2009_result}
\end{figure}

Figure~\ref{fig:phm2009_result} indicates that the metric differences among methods are relatively small on PHM2009, and the anomaly-score curves do not exhibit a fully separated pattern. Part A shows that the proposed method maintains a low background score over most normal windows in the early portion and produces concentrated, relatively strong peak responses mainly in the high-energy region around $1700$--$1950$, so its high scores are more focused on locally complex anomalies corresponding to the bright regions of the heatmap. By contrast, Deep SVDD, KNN Distance, Isolation Forest, and LOF Novelty are more likely to maintain moderately high scores over a wider temporal range, suggesting that they can sense overall distribution perturbations but find it difficult to further separate truly stronger anomalous segments from ordinary fluctuations. One-class SVM oscillates strongly across almost the entire test segment, making thresholding less meaningful. Adjacent Transformer, Transformer AE, and TranAD tend to respond only to a few sharp events with discrete peaks and therefore under-cover sustained but not strong anomalies. Hence, the advantage of the proposed method on this dataset is no longer ``complete separation,'' but rather its ability to prioritize relatively more suspicious compound-fault windows into a high-score region without substantially raising the normal background. The radar chart in Part B confirms this interpretation: the proposed method maintains a slight outward expansion on AUROC and AUPR, and its Recall is close to that of the best baseline, but its gains in Precision and best F1 are not large. This suggests that the difficulty of this dataset is not complete miss detection, but the fact that many anomalous windows differ from normal windows only weakly, making it hard for any method to produce a particularly sharp threshold boundary. The visualization in Part C further reveals the source of this difficulty: normal training windows, normal test windows, and real anomalous windows together form several intertwined ring-like manifolds, while pseudo-anomalous windows are more concentrated on an outer branch toward the right. In other words, pseudo-anomalous windows provide the model with one dominant anomaly direction, but real complex faults do not deviate along only that single direction. As a result, a considerable portion of real anomalous windows remain close to the normal manifold. This finding is consistent with the characteristics of PHM2009, where some fault windows in compound transmission systems exhibit only local or relatively weak frequency-domain differences. Overall, the proposed method provides a useful boundary reference for PHM2009, but pseudo-anomalous windows still cannot completely substitute for real complex-fault windows.
\begin{figure}[H]
    \centering
    \includegraphics[width=0.95\linewidth]{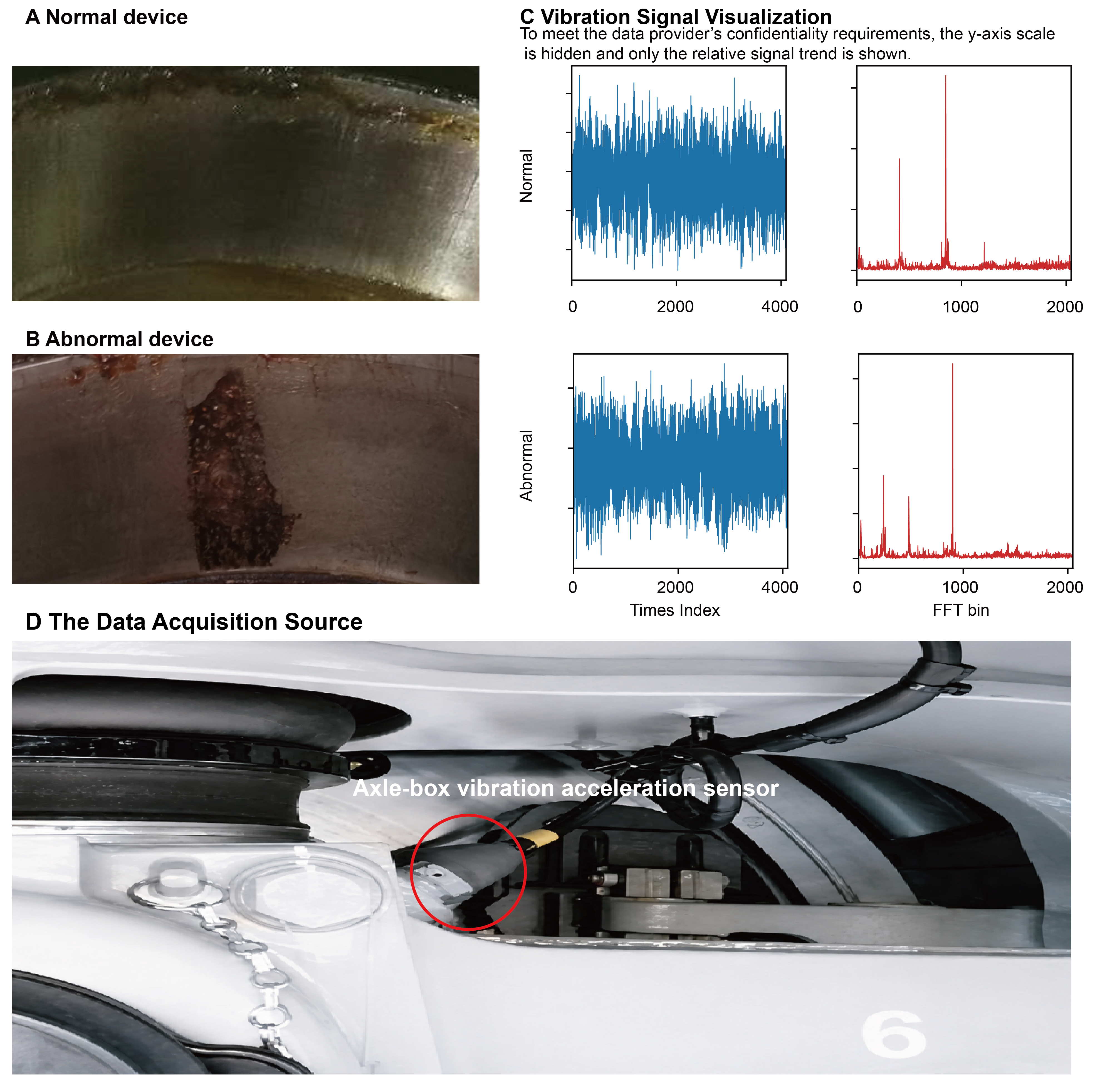}
    \caption{Illustration of the REALBOX field-measurement data. A shows the real data-acquisition scenario, B shows the long sequence formed by concatenating normal and fault samples, and C shows the corresponding FFT spectra. The data come from axial acceleration vibration records of an axle-box bearing of a high-speed train, sampled at $10\,\mathrm{kHz}$, with an inner-race failure as the fault type; part of the vertical-axis information is hidden at the request of the data provider.}
    \label{fig:realbox_data}
\end{figure}
\subsubsection{REALBOX field-measurement data}
\label{subsubsec:realbox_result}

The REALBOX dataset consists of axial vibration records from an axle-box bearing of a high-speed train, sampled at $10\,\mathrm{kHz}$ by an accelerometer, with an inner-race failure as the fault type. Compared with public laboratory datasets, its main characteristic is the extreme scarcity of measured fault samples. In real high-speed-train maintenance, severe fault samples are inherently rare, and related anomalies are often discovered or handled early by other monitoring means, so publicly available vibration fault records for analysis are very limited. The REALBOX data used in this paper come from one such retained fault-vibration record under a condition where temperature monitoring failed. Figure~\ref{fig:realbox_data} shows the data-collection scenario, the concatenated normal/fault time series, and the corresponding FFT spectra. Due to confidentiality requirements from the data provider, some vertical-axis information in the figure is hidden. Despite its limited sample size, this dataset is closer to real maintenance conditions and is therefore suitable for examining the applicability of the method in scarce real engineering scenarios.

\begin{figure}[H]
    \centering
    \includegraphics[width=0.98\linewidth]{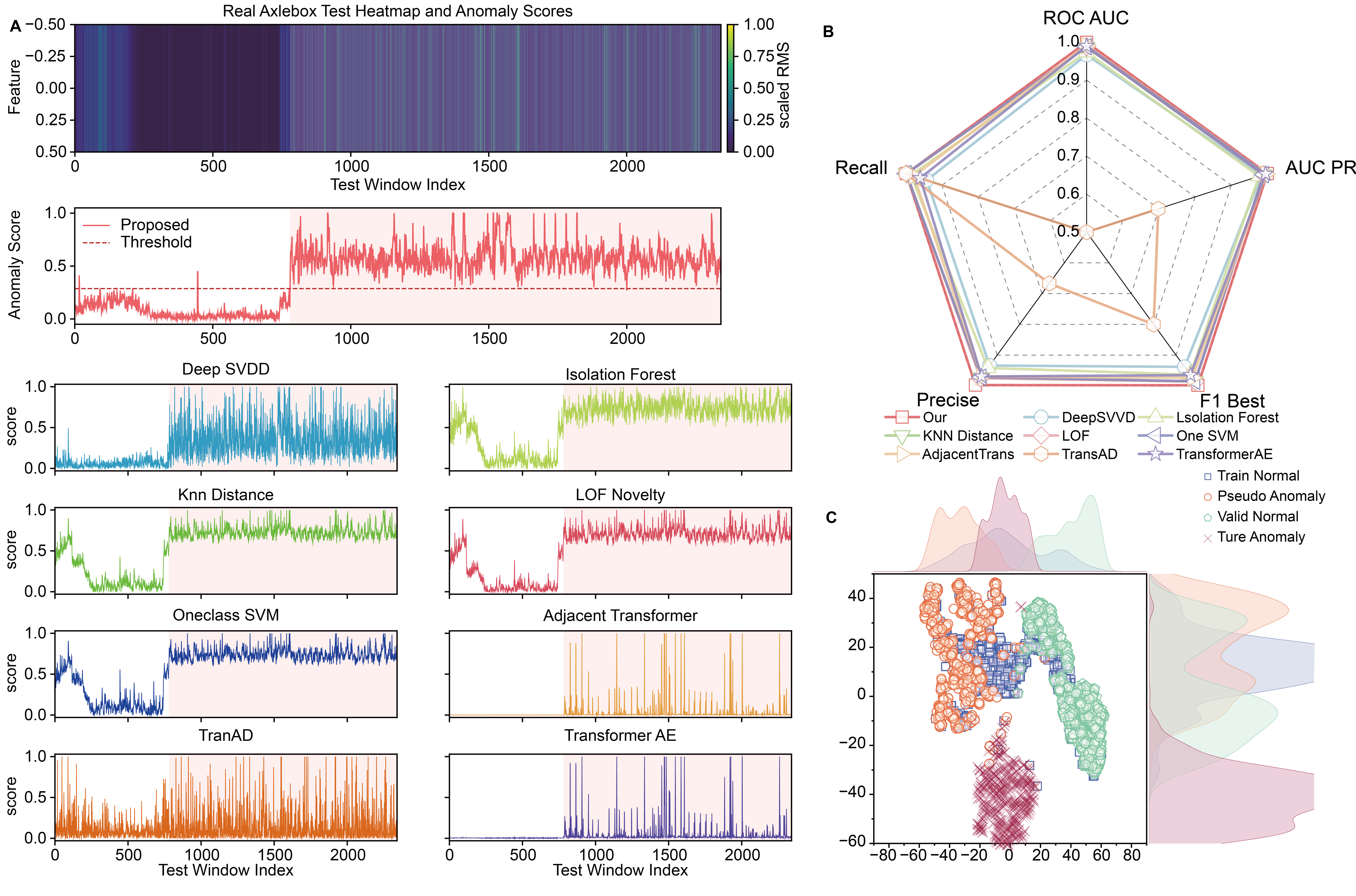}
    \caption{Fault-detection results on the REALBOX field-measurement data. A shows the window-energy heatmap and the anomaly-score curves of different methods, B shows the multi-metric radar chart, and C shows the representation-space distributions of normal training windows, normal test windows, pseudo-anomalous windows, and real anomalous windows. The dataset comes from scarce vibration records of a high-speed-train axle-box bearing with an inner-race failure.}
    \label{fig:realbox_result}
\end{figure}

Figure~\ref{fig:realbox_result} shows that the proposed method forms a relatively clear normal/anomalous boundary on REALBOX. In Part A, the proposed method maintains low scores in the early normal interval, then exhibits a clear step-like rise after about $800$ windows, and subsequently remains on a relatively stable medium-to-high plateau during the fault segment, with even higher peaks only when local impacts intensify. This ``boundary first, plateau later'' pattern makes the threshold easy to interpret. By contrast, although Isolation Forest, KNN Distance, LOF Novelty, and One-class SVM can also sense the distribution shift in the later segment, they are more prone to issues such as elevated background scores in the normal segment, near saturation in the later segment, or compressed contrast between earlier and later parts. Deep SVDD fluctuates more strongly inside the fault segment. Adjacent Transformer and Transformer AE behave more like sparse pulses, while TranAD produces spikes scattered throughout the whole sequence, making it difficult to form a stable anomaly plateau. Therefore, in this scarce field-measurement scenario, the main advantage of the proposed method is that it neither excessively raises the normal segment nor represents the fault segment merely as a few isolated peaks; instead, it produces a continuous response that is more favorable for thresholding and engineering interpretation. The radar chart in Part B shows that, except for TranAD, many methods are already close to the outer ring, so the chart is somewhat saturated here as well. Nevertheless, the proposed method still maintains the outermost contour on AUROC, AUPR, best F1, and Precision, while Recall does not decrease despite the tighter boundary. This indicates that its gains come mainly from clearer separation rather than from simply producing more anomaly alarms. The visualization in Part C provides an even more practically meaningful explanation: normal training windows lie in the middle, normal test windows form a neighboring but still continuous normal manifold on the right, pseudo-anomalous windows mainly occupy the left side and outer regions, and real anomalous windows are pushed as a whole into an independent cluster below. In other words, the pseudo-anomalous windows do not simply duplicate the real fault cluster; rather, together with the real anomalies, they surround the normal manifold from different directions, thereby encouraging a wider safety margin in the representation space. Since REALBOX contains only one publicly usable vibration fault record, this experiment should be regarded as engineering validation rather than a large-sample statistical generalization result.

Taken together, the results on the four fault-detection datasets show that the proposed method performs most stably on CWRU and REALBOX, while still maintaining favorable metrics on more complex datasets such as HTBF and PHM2009, although some anomalous samples still overlap with normal ones. If one looks only at the score curves in Part A of the figures, conventional one-class classification or distance-based methods are more easily affected by operating-condition switching and overall amplitude drift and thus tend to raise the background level, while reconstruction-/prediction-based temporal models more often appear as isolated spikes or locally delayed responses. In contrast, the proposed method more stably achieves a response pattern of ``low background in normal segments and continuous elevation in anomalous segments.'' If Parts B and C are considered together, radar charts on strongly separable datasets such as CWRU and REALBOX are often close to saturation, and the real distinction among methods lies in whether the representation space forms a clear normal core and anomalous outer edge. On complex datasets such as HTBF and PHM2009, the outward expansions in Precision/best F1 or AUROC/AUPR on the radar charts correspond more directly to whether anomalous samples truly leave the normal manifold in Part C. This is consistent with the design goal of our method: pseudo-anomalous samples are not direct substitutes for real fault samples, but controllable boundary samples for one-class detection, used to enhance the discriminability of the normal representation space.

\subsection{Degradation-process detection experiments}
\label{subsec:degradation_detection}

In addition to fixed fault-category detection, we further validate the ability of the proposed method to respond to life-cycle evolution on the XJTU-SY and IMS degradation datasets. The key difference between degradation data and fault-classification data is that anomalies usually do not emerge abruptly, but instead develop gradually over a relatively long period. Therefore, an ideal anomaly score should not only rise during the late failure stage, but should also exhibit continuous trend changes during at least part of the early degradation stage.

\subsubsection{XJTU-SY data presentation and experimental split}
\label{subsubsec:xj_data_setup}

\begin{figure}[H]
    \centering
    \includegraphics[width=0.95\linewidth]{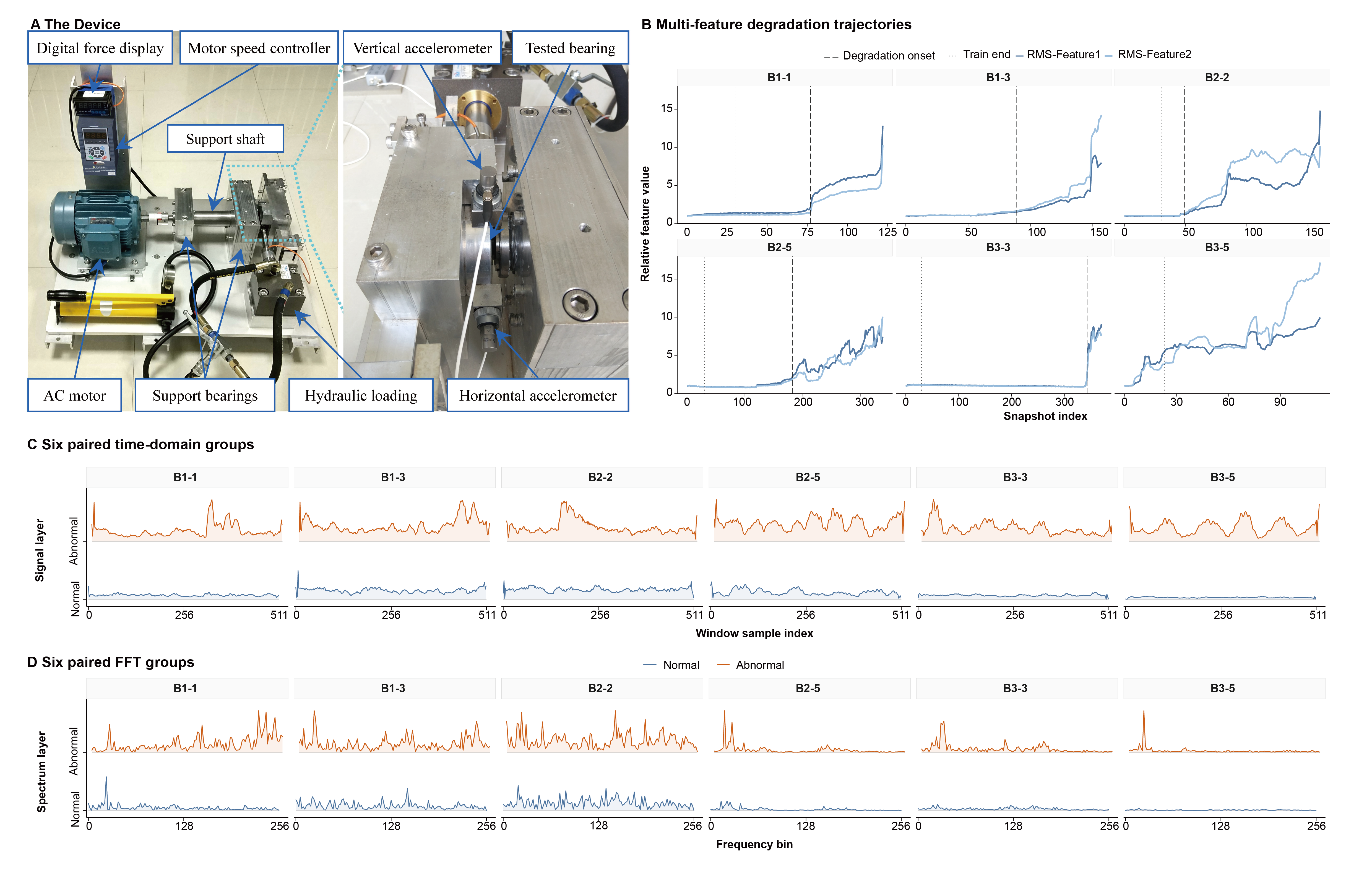}
    \caption{Illustration of the XJTU-SY degradation data. A shows the experimental setup, B shows the full-life degradation trends, and C and D show representative raw sequences and their FFT spectra. The figure is intended to illustrate the temporal evolution characteristics in the degradation experiments.}
    \label{fig:xj_data}
\end{figure}

The XJTU-SY dataset was jointly released by Xi'an Jiaotong University and Changxing Sumyoung~\cite{xjtu_sy_data}. It contains complete run-to-failure sequences for $15$ bearings under $3$ speed/load conditions. At each acquisition, both horizontal and vertical acceleration channels are recorded, with a sampling frequency of $25.6\,\mathrm{kHz}$ and a signal length of $32768$ saved every minute. Part B of Fig.~\ref{fig:xj_data} shows that different bearings exhibit markedly different degradation trajectories: some remain stable for a long time and then rise sharply, while others show sustained gradual growth. Parts C and D further show that the changes in time-domain impacts and frequency-domain energy concentration in the late degradation stage are also inconsistent across samples. Therefore, this dataset is more suitable for validating the adaptability of anomaly scores to different degradation trends rather than evaluating only a single failure pattern.

We select Bearing1-1, Bearing1-3, Bearing2-2, Bearing2-5, Bearing3-3, and Bearing3-5 for validation, covering different degradation trajectories under the three operating conditions of $35$ Hz/$12$ kN, $37.5$ Hz/$11$ kN, and $40$ Hz/$10$ kN. For each sample, the model is trained on early-life normal snapshots, and the remaining sequence is used for testing. Bearing3-5 uses the first $23$ snapshots as training data because an earlier degradation onset is detected, whereas all other samples use the first $30$ snapshots. The window length and stride are $512$ and $256$, respectively. The target error interval for Stage~1 is set to $[0.0008,0.006]$, and pseudo-anomalous windows are sampled from this interval. The purpose of this setting is not to directly predict remaining useful life, but to test whether the anomaly score can reflect deviations in health state relative to the early normal state.

\subsubsection{Multi-bearing degradation detection results on XJTU-SY}
\label{subsubsec:xj_results}

\begin{figure}[H]
    \centering
    \includegraphics[width=0.98\linewidth]{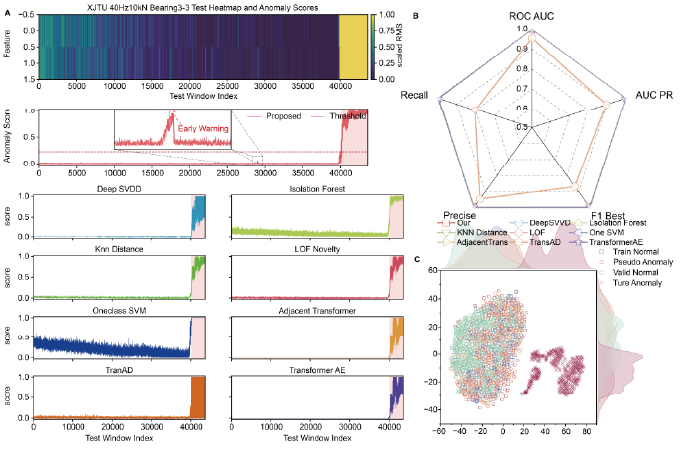}
    \caption{Degradation-detection results for XJTU-SY Bearing1-1. The proposed method maintains low scores during the normal stage and produces a steadily increasing anomaly response in the late degradation stage; a slight score rise appears in some pre-failure windows.}
    \label{fig:xj_11_result}
\end{figure}

\begin{figure}[H]
    \centering
    \includegraphics[width=0.98\linewidth]{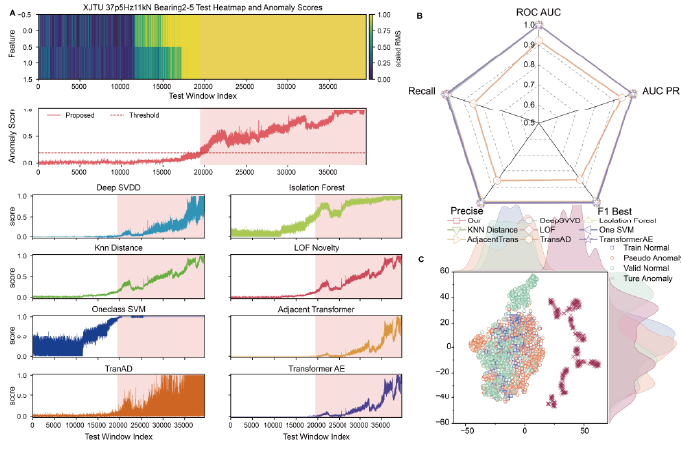}
    \caption{Degradation-detection results for XJTU-SY Bearing1-3. The anomaly-score curve and representation-space visualization jointly show the model's response to the degradation stages.}
    \label{fig:xj_13_result}
\end{figure}

\begin{figure}[H]
    \centering
    \includegraphics[width=0.98\linewidth]{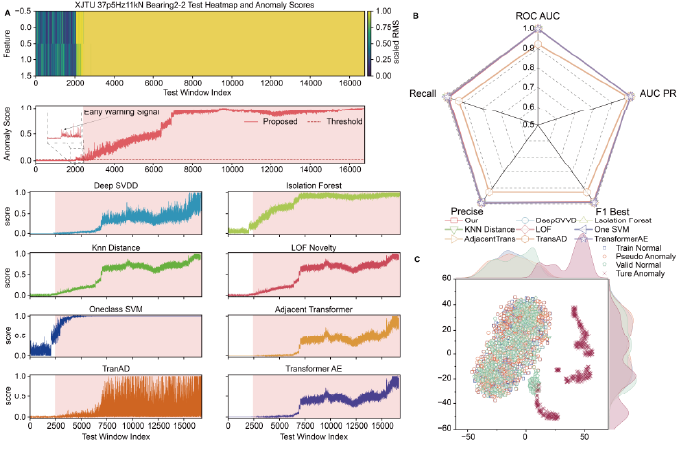}
    \caption{Degradation-detection results for XJTU-SY Bearing2-2. The proposed method shows continuous score variation before and after degradation and can be used to observe early degradation trends.}
    \label{fig:xj_22_result}
\end{figure}

\begin{figure}[H]
    \centering
    \includegraphics[width=0.98\linewidth]{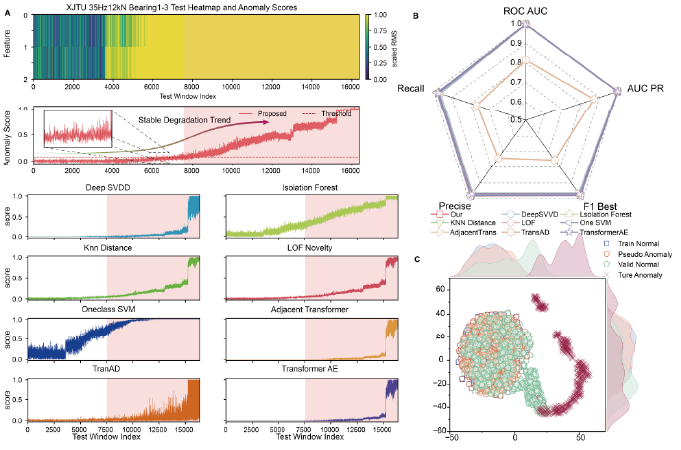}
    \caption{Degradation-detection results for XJTU-SY Bearing2-5. The figure shows anomaly scores, metric comparisons, and representation-space distributions, with the late anomalous stage overall receiving higher scores.}
    \label{fig:xj_25_result}
\end{figure}

\begin{figure}[H]
    \centering
    \includegraphics[width=0.98\linewidth]{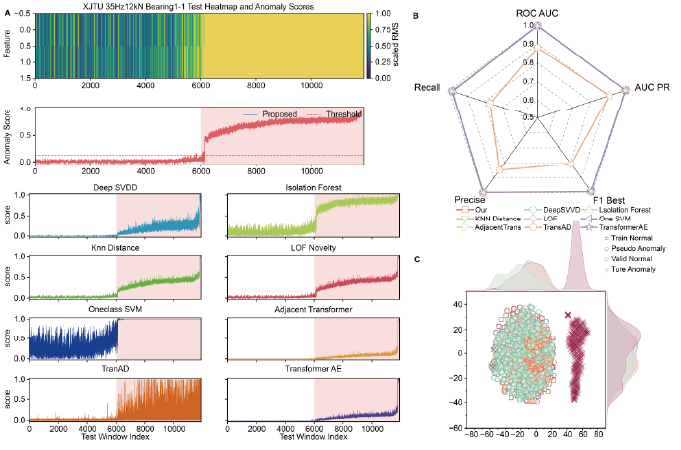}
    \caption{Degradation-detection results for XJTU-SY Bearing3-3. The model shows progressively stronger anomaly responses before and after failure.}
    \label{fig:xj_33_result}
\end{figure}

\begin{figure}[H]
    \centering
    \includegraphics[width=0.98\linewidth]{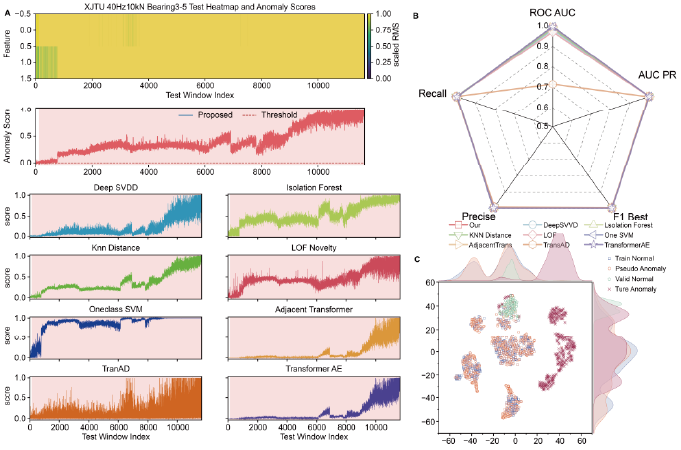}
    \caption{Degradation-detection results for XJTU-SY Bearing3-5. The proposed method yields stable high scores in the late degradation stage and score elevations in part of the early stage.}
    \label{fig:xj_35_result}
\end{figure}

Figures~\ref{fig:xj_11_result} to \ref{fig:xj_35_result} present the detection results for the six XJTU-SY bearing samples. Each figure contains anomaly-score curves, metric comparisons, and representation-space visualizations. The anomaly-score curves are used to observe detection responses as time evolves, while t-SNE is used to help assess the relative positions of normal training windows, normal test windows, pseudo-anomalous windows, and real anomalous windows in the representation space. Part A of these figures also reveals that the baseline methods in degradation tasks roughly fall into three groups: one group, such as One-class SVM and Isolation Forest, tends to maintain an elevated background or saturate too early over long intervals; a second group, such as Deep SVDD, KNN Distance, and LOF Novelty, can sense late-stage distribution shifts but often responds with delayed rises or strong fluctuations depending on the sample; and a third group, such as Adjacent Transformer, Transformer AE, and TranAD, tends to produce local spikes. Parts B and C play a slightly different role here than in fault detection: because many samples contain a high proportion of late-stage degradation windows, radar charts often become nearly saturated for multiple methods, so Part B is mainly useful for checking whether a method sacrifices Precision/best F1 in order to detect early signs. The more interpretable evidence lies in Part C, namely whether real anomalous windows leave the normal manifold to form one or more outer branches, and whether pseudo-anomalous windows push the boundary outward without fragmenting the normal cluster. The goal of the proposed method is not necessarily to cross the threshold earliest on every sample, but to make the score evolution consistent with the stage-wise strengthening of the degradation process.

For Bearing1-1 in Fig.~\ref{fig:xj_11_result}, a clear boundary appears after about $6000$ windows: the window-energy heatmap in Part A gradually changes from mixed colors into a sustained high-energy region, and the anomaly score of the proposed method correspondingly jumps from a low value near the threshold to a stable high level. Compared with the baselines, Isolation Forest also exhibits a rapid transition near the failure point, but its plateau saturates earlier and has a smaller dynamic range in the later segment. Deep SVDD, KNN Distance, and LOF Novelty increase gradually in the later stage, but do so more slowly and with stronger fluctuations. One-class SVM maintains an elevated background far earlier, while Adjacent Transformer and Transformer AE are overly flat for most of the sequence and only respond near the very end. By contrast, the proposed method preserves a low background in the early stage and forms a smooth, sustained high-score plateau after failure, which is more consistent with the stage characteristics of samples exhibiting rapid late-stage failure. In Part B, the metrics of many methods are already close to the outer ring except for TranAD, indicating that detecting the late fault stage of this sample is not particularly difficult in itself. Thus, the key here is not the absolute value of any single metric, but whether high scores can be achieved while preserving interpretability across the degradation stages. Part C shows that normal training windows and normal test windows are mainly concentrated in a left-side core region, pseudo-anomalous windows form a wider envelope around them, and real anomalous windows gather into a slender independent cluster on the right. This structure indicates that the model has learned not a random dispersion, but a clear separation direction from the normal core toward the late failure region.

Unlike the jump-like transition of Bearing1-1, Bearing1-3 in Fig.~\ref{fig:xj_13_result} exhibits a smoother degradation trend. The proposed method does not rise sharply immediately after the training segment ends; instead, it grows slowly over a long period and approaches the high-score region only in the later part, which is consistent with the stable degradation trend marked in the figure. The baseline comparison in Part A shows that Deep SVDD, KNN Distance, LOF Novelty, Adjacent Transformer, and Transformer AE tend to surge only near the end, compressing the long progressive degradation into what appears to be a late abrupt change. One-class SVM and Isolation Forest rise earlier, but their background is higher in the early and middle stages, making it difficult to distinguish slight degradation from severe degradation. In contrast, the proposed method preserves a continuous score transition from weak deviation to strong degradation, making it more suitable for describing this kind of long and smooth degradation evolution. Part B is again nearly saturated, indicating that many methods can identify the severe late-stage degradation if only final labels are considered. However, Part C better reveals the difference: real anomalous samples form a smooth arc-shaped manifold on the right and continue to separate from the dense normal/pseudo-anomalous cloud on the left, rather than separating abruptly only at the end as with some baselines. This continuously unfolding geometry is consistent with the gradual rise observed in Part A.

Bearing2-2 in Fig.~\ref{fig:xj_22_result} exhibits a more typical early-warning pattern. Part A shows that before the arrival of the large-scale high-score region, the proposed anomaly score has already risen continuously in the earlier stage and has crossed the posterior best-F1 threshold shown in the figure; the corresponding heatmap also begins to exhibit local energy enhancement after about $2000$ windows. Notably, some baseline methods also respond early, but in different ways. One-class SVM quickly approaches saturation at an early stage and loses the ability to distinguish mild degradation from severe late-stage degradation. Isolation Forest, KNN Distance, LOF Novelty, and Transformer AE all show sustained growth later on, but more often remain on a long intermediate plateau, and their stage boundaries are less clear than those of the proposed method. Deep SVDD, Adjacent Transformer, and TranAD respond more strongly only later. In contrast, the proposed method preserves two distinct rising levels between early warning and later strengthening, making it more suitable for highlighting weak precursor signals in degradation monitoring. Part B suggests that most methods still achieve relatively high final AUROC/AUPR values, so the real value here lies in whether an interpretable signal can be produced earlier without significantly sacrificing Precision/best F1. Part C shows that real anomalous windows do not form a single cluster, but instead split into multiple substructures such as an upper-right arc and a lower branch, echoing the two-stage rise seen in Part A. The pseudo-anomalous windows mainly remain close to the outer boundary of the normal manifold, suggesting that they play a boundary-expansion role rather than directly substituting for these specific anomalous submodes.

Bearing2-5 has a longer life sequence, and its late degradation does not occur as a single jump but strengthens in multiple stages. In Fig.~\ref{fig:xj_25_result}, the anomaly score of the proposed method begins to rise continuously after about $2\times10^4$ windows, and several stepwise high-score intervals appear in the middle and later stages, consistent with the fact that the two channels enter high-energy regions successively in the heatmap. Part A also shows that Deep SVDD, KNN Distance, and LOF Novelty can follow the general worsening trend, but with stronger fluctuations and blurrier boundaries among different stages. Isolation Forest rises clearly in the middle stage but also exhibits more local drops. One-class SVM enters the high-score region too early and approaches saturation quickly, while Adjacent Transformer and Transformer AE strengthen only in a relatively late stage. Thus, for this kind of multi-stage degradation sample, the advantage of the proposed method lies in preserving the form of progressive worsening rather than compressing the entire late stage into a single uniformly high-score state. Part B again appears nearly saturated near the outer ring, indicating that if only late-life labels are aggregated, many methods can score highly. However, Part C reveals finer structure: real anomalous samples do not collapse into a single group but instead split into multiple isolated island-like branches, indicating that the late-stage degradation indeed contains multiple states with different severities, while normal and pseudo-anomalous samples remain mainly in a relatively compact left-side core region. This suggests that the proposed method separates normal and anomalous states while also preserving stage hierarchy inside the anomalous region.

Bearing3-3 in Fig.~\ref{fig:xj_33_result} remains approximately normal over most of the test interval, exhibits only one local high-score pulse before the actual failure, and then rapidly enters a sustained high-score region near the end. The comparison in Part A especially highlights the difference between the proposed method and the baselines: Deep SVDD, KNN Distance, LOF Novelty, Isolation Forest, Adjacent Transformer, and Transformer AE mostly surge only at the end and show almost no significant response to the precursor; One-class SVM also rises rapidly at the end, but has a higher background beforehand; TranAD is dominated almost entirely by the final spikes. This sample suggests that the precursors of some degradation sequences do not manifest as a long gradual rise, but rather as a combination of sparse precursors and a final abrupt change. In such a case, the value of the proposed method lies not only in identifying the final failure, but also in separating a brief, local, yet meaningful early deviation from a long low-background stage. Since the radar chart in Part B is almost unable to differentiate most methods on this sample, it also shows that final summary statistics alone do not fully reflect the value of precursor sensitivity. Part C is more interpretable: real anomalous samples on the right split into several relatively compact small clusters, while normal and pseudo-anomalous samples on the left remain as a dominant manifold, indicating that both a small number of early precursors and severe late-stage failure are mapped outside the normal region without breaking the consistency of the normal samples.

The test segment of Bearing3-5 exhibits sustained anomaly responses above the threshold shortly after it begins, indicating that distribution shift has already occurred relatively early after the training segment ends. Although the anomaly score of the proposed method fluctuates in the middle stage, its overall trend continues to move toward higher-score regions and eventually forms a stable high-score plateau near the end. The comparison in Part A shows that One-class SVM, KNN Distance, and Isolation Forest also enter the high-score region early, but they are more likely to maintain an elevated background over a long interval, compressing the difference between early deviation and later severe degradation. Deep SVDD, Adjacent Transformer, and Transformer AE rise later, while TranAD mainly produces a large number of fluctuating spikes. In contrast, under this challenging setting where the test segment is already degraded for almost its entire duration, the proposed method still preserves a hierarchy from moderately high scores to an even higher plateau, meaning that it can indicate not only that the sequence has already deviated from normality, but also whether the degradation continues to intensify. In Part B, almost all methods except TranAD are again close to the outer ring, which further shows that overall AUROC/AUPR alone is insufficient for reflecting which method better captures degradation hierarchy in long degradation sequences. Part C more clearly indicates that real anomalous windows form a continuous arc-shaped branch on the right, whereas normal training windows, normal test windows, and pseudo-anomalous windows are distributed among several adjacent subclusters on the left. This structure of ``multiple but connected normal subclusters and an overall outward shift of anomalies'' is consistent with the continuous progression from early deviation to severe late-stage degradation shown in Part A.

Across the six XJTU-SY samples, three typical response patterns can be observed. Bearing1-1 and Bearing3-3 are closer to the pattern of ``long stable period followed by rapid failure.'' Bearing1-3 and Bearing2-5 exhibit more continuous progressive degradation. Bearing2-2 and Bearing3-5 provide more explicit early warning signals before formal failure. If one summarizes the shapes of the baseline score curves in Part A, One-class SVM and some tree-/distance-based methods are more likely to show elevated backgrounds or overly early saturation, compressing the hierarchy among degradation stages. Deep SVDD, KNN Distance, and LOF Novelty are sensitive to overall shifts, but their rise timing and fluctuation strength are not stable across samples. Adjacent Transformer, Transformer AE, and TranAD are more likely to express degradation responses as local spikes or late concentrated surges. If Parts B and C are considered together, many radar charts for XJTU-SY are already near the performance ceiling; thus, they more strongly reflect whether a method can detect late-stage degradation than whether it can capture early warnings. What more directly reflects method differences is whether real anomalous windows form one or more outer branches detached from the normal manifold in Part C. Compared with the baselines, the proposed method more consistently preserves the temporal order of degradation stages, enabling anomaly scores not only to detect late-stage fault states but also to capture weak pre-failure signs in some samples. At the same time, we do not interpret these scores as direct estimates of remaining useful life, nor do we claim that they can precisely locate the degradation onset. A more appropriate interpretation is that the score curve characterizes the degree to which test windows deviate from the early normal training distribution. Low-dimensional visualization also shows that, for most samples, real anomalous windows form relatively clear boundaries away from normal windows, while pseudo-anomalous windows serve as boundary references.

\subsubsection{IMS degradation data experiment}
\label{subsubsec:ims_result}

The IMS bearing degradation dataset was provided by the Center for Intelligent Maintenance Systems (IMS) at the University of Cincinnati and later released through the NASA open data portal~\cite{ims_data}. We select the Bearing4 sequence as a representative long-life degradation sample to examine the stability of the proposed method under long-duration weak degradation and transitional states.

\begin{figure}[H]
    \centering
    \includegraphics[width=0.95\linewidth]{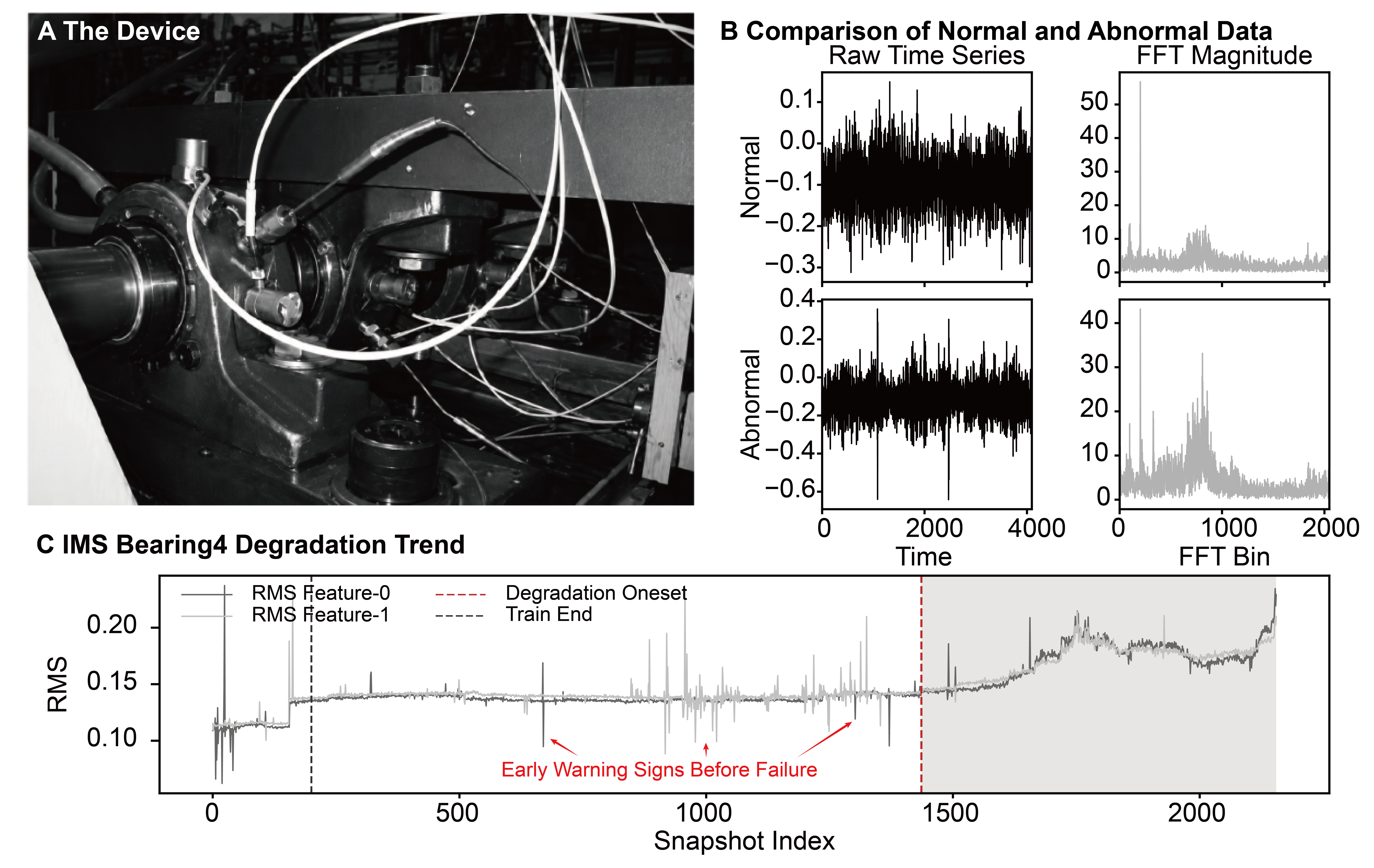}
    \caption{Illustration of the IMS bearing degradation data. The figure shows the experimental setup, long-sequence degradation trend, time-domain signal, and frequency-domain changes in the IMS run-to-failure data, and is intended to illustrate the input characteristics of long-sequence degradation scenarios.}
    \label{fig:ims_data}
\end{figure}

Figure~\ref{fig:ims_data} highlights three key characteristics of IMS Bearing4. First, the RMS degradation trajectory in Part C remains relatively stable over most of the early and middle stages, but already exhibits scattered spikes before the official degradation onset, indicating that the dataset does not consist of only two sharply separated stages, i.e., ``normal'' and ``failed.'' Second, the comparison of normal and anomalous windows in Part B shows that anomalous samples have enhanced time-domain impact amplitudes and stronger dominant frequency peaks, although the enhancement is not as drastic as in some XJTU-SY samples. Finally, the figure suggests that IMS is better suited as a validation scenario for long-duration gradual degradation, where the key issue is model tolerance to weak signs and transitional states.

\begin{figure}[H]
    \centering
    \includegraphics[width=0.98\linewidth]{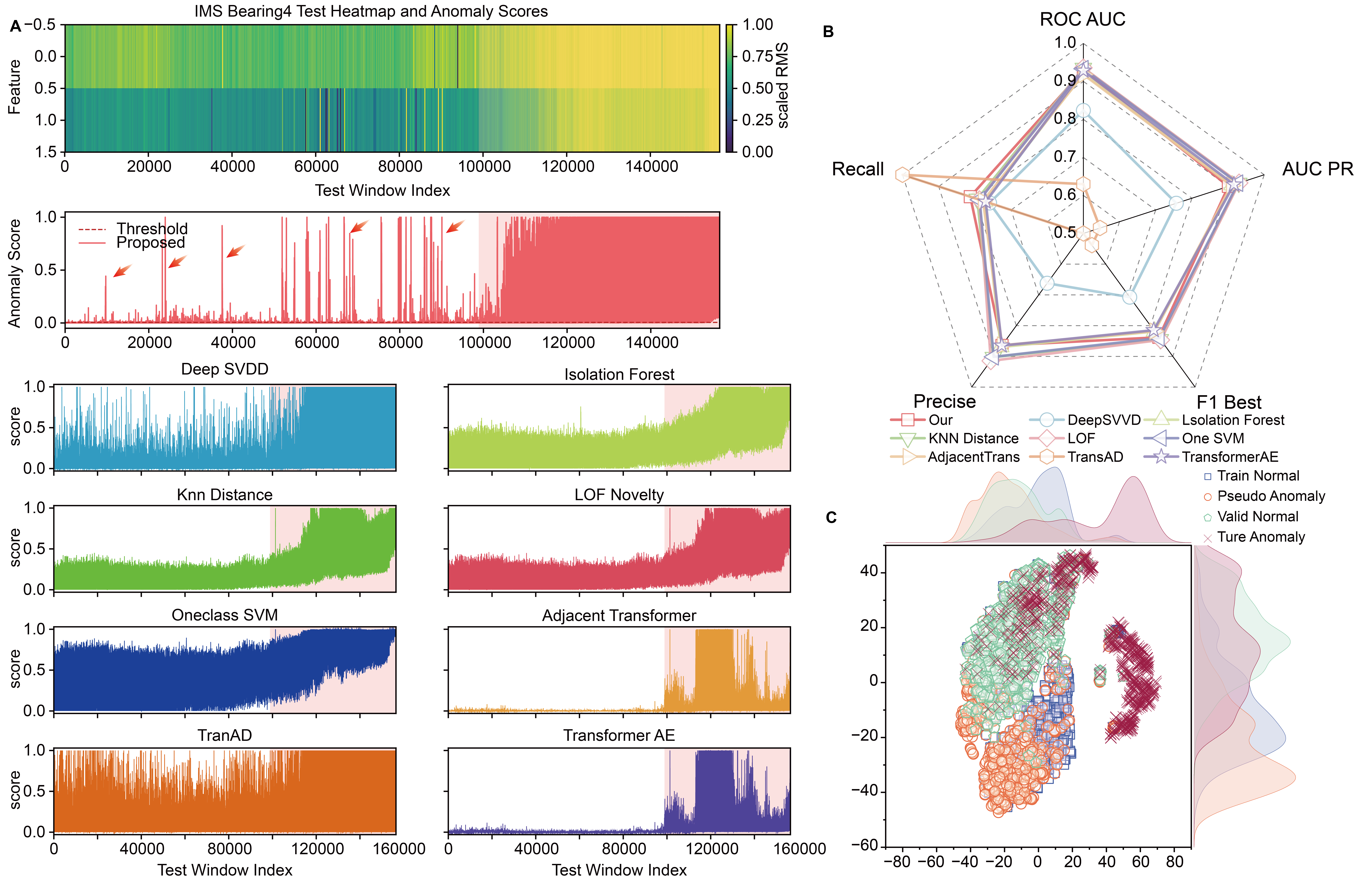}
    \caption{Degradation-detection results on the IMS bearing data. The proposed method captures part of the anomaly response that strengthens over time, although some anomalous windows still receive scores close to those of normal windows.}
    \label{fig:ims_result}
\end{figure}

The IMS test sequence is much longer and contains more transitional states in which normal and anomalous samples overlap. Taking Bearing4 as an example, the training-set size is $4096000\times 2$, the test-set size is $40058880\times 2$, and the window length and stride remain $512$ and $256$, respectively. Part A of Fig.~\ref{fig:ims_result} shows that the window energy of the two channels as a whole enters a higher region after about $10^5$ windows, while the anomaly score of the proposed method already exhibits multiple isolated peaks before that point, echoing the early warning signs before failure marked in Fig.~\ref{fig:ims_data}. After about $1.05\times10^5$ windows, many windows continuously enter the high-score region, indicating that the model can accumulate responses to progressively worsening degradation. Compared with the baselines, the difference is again evident here. Deep SVDD and TranAD show strong fluctuations or dense peaks over a long early interval, making them prone to mapping normal fluctuations in long sequences to high scores. One-class SVM maintains an elevated background over almost the entire test segment, leaving little room for threshold discrimination. KNN Distance, LOF Novelty, and Isolation Forest also show an overall rise in the later stage, but more in the form of gradual elevation on top of a relatively high background, making them less targeted at early sparse weak signs. Adjacent Transformer and Transformer AE respond mainly to larger regions in later stages and provide insufficient coverage for smaller early signs. In contrast, the strength of the proposed method on IMS is not that all anomalous windows are clearly separated, but that it simultaneously preserves two types of information: sparse early warning spikes and a sustained high-score platform in the late stage. The radar chart in Part B supports this observation: the proposed method maintains a relatively large outward area on AUROC, AUPR, best F1, and Precision, whereas methods such as Adjacent Transformer are more aggressive on Recall at the expense of Precision, and Deep SVDD lags on multiple metrics. This suggests that, in long-sequence degradation scenarios such as IMS, the more valuable property is not to crudely label more windows as anomalous, but to maintain a relatively balanced Precision--Recall trade-off while preserving strong ranking ability. The t-SNE visualization in Part C further shows that a substantial portion of real anomalous windows have already detached from the normal manifold along a right-side arc-shaped branch, while another portion remains close to or even partially overlaps with the region of normal test windows. Meanwhile, pseudo-anomalous windows mainly lie outside the normal manifold in another direction. This structure indicates that degradation in IMS does not follow a single, clear anomaly trajectory, but simultaneously includes late-stage states that have clearly moved away from normality and transitional states that are still close to normal. Therefore, for long-life degradation data such as IMS, anomaly-detection scores are better interpreted as references to degradation trends rather than as the sole basis for pointwise failure decisions.

\begin{figure}[H]
    \centering
    \includegraphics[width=0.98\linewidth]{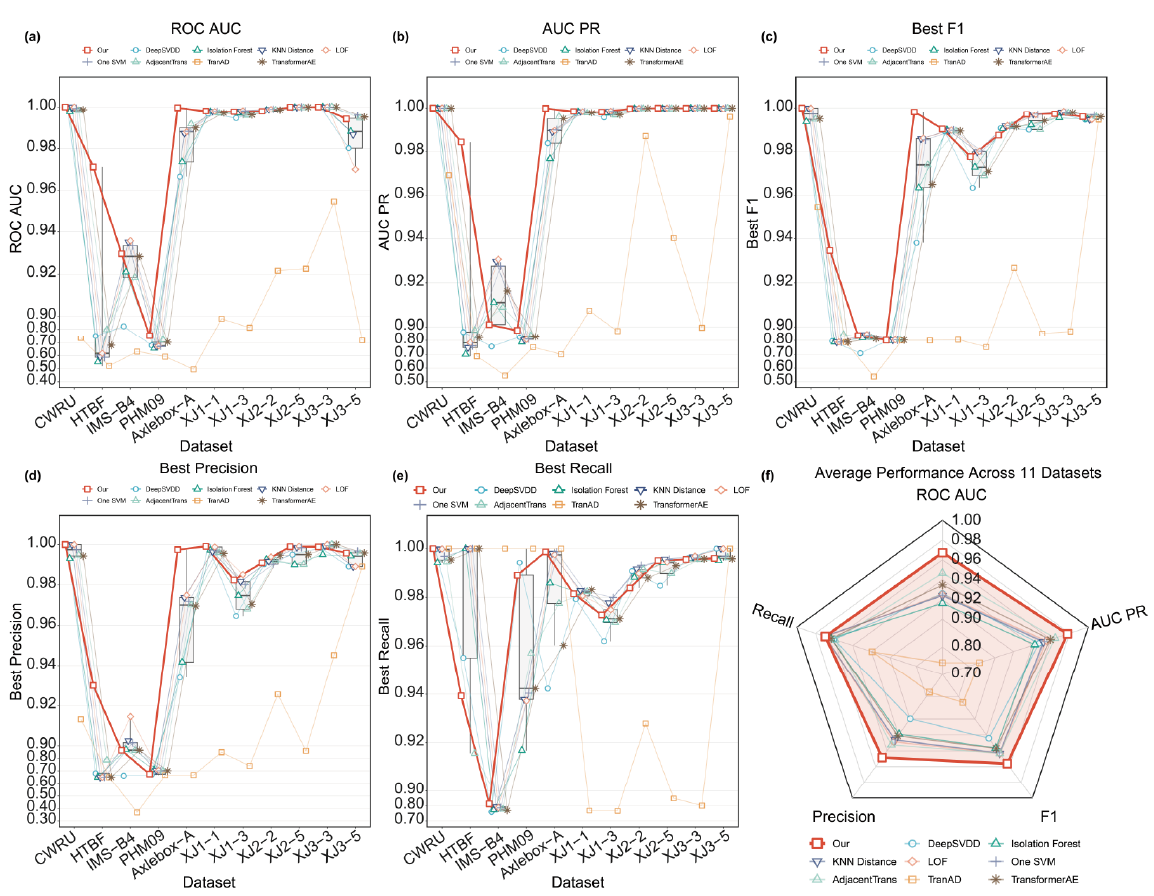}
    \caption{Summary of the core metrics across multiple datasets. A--E compare the results of $9$ methods on AUROC, AUPR, best F1, Precision, and Recall across $11$ datasets, and F shows the radar chart of cross-dataset averages. This figure summarizes the overall performance of the proposed method on bearing fault detection and degradation detection tasks.}
    \label{fig:overall_metrics}
\end{figure}

\subsubsection{Experimental summary}
Figure~\ref{fig:overall_metrics} provides a unified summary of the core metrics from the above $11$ fault-detection and degradation-detection experiments. Panels A--E correspond to the five evaluation metrics, and Panel F is the averaged radar chart over different metrics. This figure is not intended to replace the dataset-by-dataset anomaly-score analysis given earlier, but rather to compare the overall trend of different methods under a unified coordinate system. Overall, the proposed method occupies the largest area in the averaged radar chart and achieves the highest average performance on AUROC, AUPR, best F1, and Precision, with average values of $0.9673$, $0.9778$, $0.9562$, and $0.9485$, respectively. The average Recall is $0.9695$, which is not the highest among all methods but still remains high. This indicates that the strength of the proposed method does not lie mainly in more aggressive recall, but rather in a more stable balance of ranking ability, threshold-based detection performance, and Precision--Recall trade-off across datasets.

A closer examination of the single-dataset results shows that the proposed method achieves the best AUROC and AUPR on $5$ datasets each, and the best best-F1 and Precision on $4$ datasets each, indicating that its advantage does not come from only one type of metric. This advantage is more pronounced on datasets that better differentiate methods, such as HTBF, PHM2009, and REALBOX. On IMS Bearing4 and some XJTU-SY subsets, however, the gaps between the proposed method and LOF Novelty, KNN Distance, and One-class SVM are smaller, and some individual metrics are slightly better for those baselines. In other words, this summary figure is better suited to support the conclusion that the proposed method is superior in overall cross-dataset stability and comprehensive performance, rather than to suggest that it has absolute superiority in every scenario and on every metric. In addition, the larger dispersion among methods on IMS, HTBF, and REALBOX also indicates that these datasets are more discriminative in terms of anomaly-detection capability and threshold-selection stability.

\subsection{Ablation studies}
\label{subsec:ablation}

To further explain the effectiveness of the proposed framework, we conduct ablation analyses from three perspectives: how pseudo-anomalies are generated, whether pseudo-anomalies form effective boundaries in the representation space, and how such boundaries are translated into final anomaly scores. Unlike the main experiments, which focus on final detection metrics, the ablation studies focus more on the mechanism of the method itself. Ideally, pseudo-anomalous samples should not simply be ``easy negatives'' that lie far away from normal samples; instead, they should lie outside the normal manifold, exhibit some directional consistency with possible real anomalies, and be transformed into a stable distance structure after Stage~2 training. Accordingly, the following analysis combines two-dimensional embedding distributions, distance statistics, PCA/Sankey coverage relations, anomaly-score curves, and metrics such as AUROC, AUPR, and best F1.

\subsubsection{Ablation I: pseudo-anomaly injection strategy}
\label{subsubsec:ablation_pseudo}
\begin{figure}[htbp]
    \centering
    \includegraphics[width=0.96\linewidth,height=0.45\textheight,keepaspectratio]{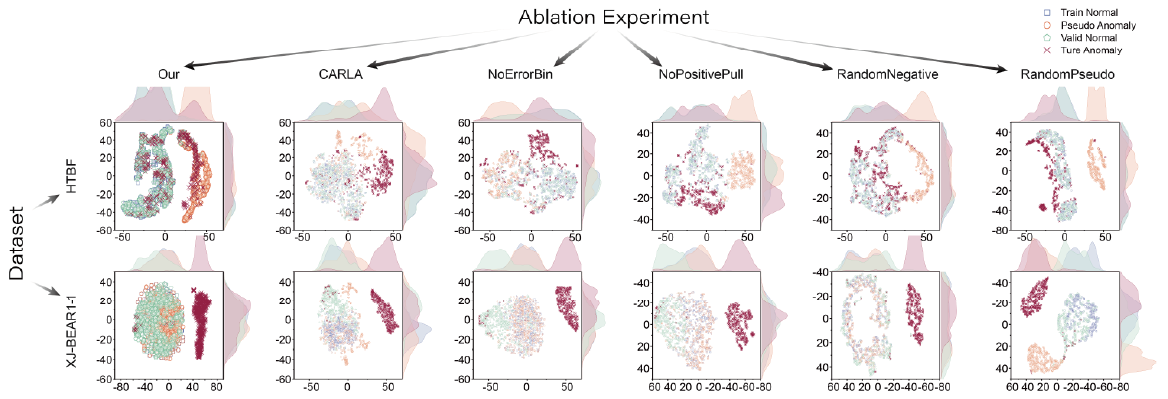}
    \caption{Ablation on pseudo-anomaly injection strategies. The upper row corresponds to HTBF and the lower row to XJTU-SY Bearing1-1. The figure compares six strategies---our method, CARLA, removing error-bin balancing, removing positive-pair pulling, random pseudo-anomalous negatives, and random pseudo-anomalies---in terms of the distributions of normal training samples, pseudo-anomalies, validation-normal samples, and real anomalies in the two-dimensional embedding space.}
    \label{fig:ablation_pseudo}
\end{figure}
Figure~\ref{fig:ablation_pseudo} compares the proposed pseudo-anomaly generation method, CARLA anomaly injection, and several alternative strategies obtained by removing key design components. The core question in this group of experiments is whether the negative samples generated by different anomaly-injection schemes truly act as ``boundary samples.'' For bearing anomaly detection under normal-only training, pseudo-anomalies do not need to reproduce the real fault morphology point by point during testing. More importantly, they should form a continuous, learnable reference region outside the normal sample manifold without deviating excessively from the physical data distribution. If pseudo-anomalies are too close to normal samples, Stage~2 cannot learn a sufficiently discriminative margin; if they are too random or too far from the normal manifold, the model is more likely to learn artifacts unrelated to real faults, resulting in representations that are easy to separate during training but unstable during testing.

\begin{table}[htbp]
    \centering
    \caption{Quantitative results of Ablation I on HTBF and XJTU-SY Bearing1-1.}
    \label{tab:ablation_pseudo_metrics}
    \small
    \renewcommand{\arraystretch}{1.12}
    \setlength{\tabcolsep}{5pt}
    \begin{tabular}{lcccccc}
        \toprule
        Setting & \multicolumn{3}{c}{HTBF} & \multicolumn{3}{c}{XJTU-SY Bearing1-1} \\
        \cmidrule(lr){2-4} \cmidrule(lr){5-7}
        & AUROC & AUPR & Best F1 & AUROC & AUPR & Best F1 \\
        \midrule
        Proposed method & \textbf{0.9711} & \textbf{0.9842} & \textbf{0.9347} & 0.9889 & \textbf{0.9954} & \textbf{0.9784} \\
        CARLA pseudo-anomalies & 0.9694 & 0.9655 & 0.9187 & 0.9595 & 0.9686 & 0.9548 \\
        Without error-bin balancing & 0.9618 & 0.9815 & 0.9251 & 0.9889 & 0.9954 & 0.9777 \\
        Without positive pulling & 0.8663 & 0.9228 & 0.8543 & 0.9777 & 0.9897 & 0.9742 \\
        Random pseudo-anomalous negatives & 0.7935 & 0.8667 & 0.8426 & 0.9470 & 0.9327 & 0.9590 \\
        Random pseudo-anomalies & 0.8416 & 0.9091 & 0.8541 & \textbf{0.9914} & 0.9913 & 0.9771 \\
        \bottomrule
    \end{tabular}
\end{table}

Table~\ref{tab:ablation_pseudo_metrics} further reports the quantitative results of the above six strategies on HTBF and XJTU-SY Bearing1-1. On the more complex HTBF dataset, the proposed method achieves the highest AUROC (0.9711), while remaining close to CARLA pseudo-injection on AUPR (0.9842) and best F1 (0.9347). After removing error-bin balancing, all metrics decrease slightly, indicating that simply generating boundary samples is not sufficient for stable detection; balanced coverage of different boundary strengths also affects the final performance. By contrast, random pseudo-anomalies, random pseudo-anomalous negatives, and removing positive pulling all lead to more obvious performance degradation, suggesting that when negative samples are not properly aligned with the normal boundary, or when the normal neighborhood is not sufficiently compacted, the representation learned in Stage~2 becomes more easily influenced by artifacts from random perturbations. On XJTU-SY Bearing1-1, all six strategies perform relatively well overall, indicating that the anomaly direction of this degradation sample is relatively concentrated and that several pseudo-anomaly construction schemes can provide useful training signals. Even so, the proposed method still achieves the highest best F1 (0.9784), whereas CARLA is slightly higher on AUROC and AUPR, suggesting that on relatively easy data, the difference among pseudo-anomaly strategies is reflected more in boundary details and threshold-based results than in global separability itself.

The distributions on HTBF and XJTU-SY Bearing1-1 in Fig.~\ref{fig:ablation_pseudo} show that the pseudo-anomalous samples generated by the proposed method tend to expand along the outside of the normal samples and maintain some directional consistency with the main shift directions of real anomalies. Taking HTBF as an example, the normal training samples and validation-normal samples form a relatively compact normal region, the proposed pseudo-anomalies are mainly distributed outside that region, and the real anomalies lie further along or adjacent to the outer edge. This indicates that the samples generated in Stage~1 are not simple random perturbations, but are instead extrapolated along vulnerable directions of the normal manifold in the reconstruction-error space, thereby providing hard negatives that are close to real anomaly directions for Stage~2. In XJTU-SY Bearing1-1, the life-stage difference between real degradation samples and normal samples is more pronounced, and the proposed method likewise constructs negative references near the normal region, making it easier for real degradation windows to be pushed away from the normal core in the embedding space.

By contrast, CARLA's contextual, global, seasonal, shapelet, and trend injection schemes impose stronger template priors and can generate multiple semantically explicit types of time-series anomalies, but these templates do not necessarily align with the reconstruction-residual boundary in bearing vibration data. CARLA may produce effective negative samples in scenarios where trend anomalies or shapelet anomalies are more pronounced; however, in multi-channel and strongly perturbed data such as HTBF, template-based injection can easily create several discrete anomaly directions, and the overlap between pseudo-anomalies and real anomalies may occur only in local regions. In other words, CARLA has the advantage of anomaly-pattern diversity, whereas its limitation is that anomaly directions are determined by predefined transformations. The advantage of the proposed method lies in the fact that anomaly strength is adaptively derived from the reconstruction-error distribution of the normal training samples, which is more suitable for constructing bearing-specific pseudo-anomalies close to the normal boundary.

The results after removing key modules further show the necessity of boundary construction. Without error-bin balancing, pseudo-anomalies are more likely to concentrate in the error intervals that are easiest for the controller to hit, leading to insufficient samples along some boundary directions. Although such pseudo-anomalies may still separate from normal samples, their uneven coverage makes Stage~2 biased toward a few anomaly-strength regions and weakens generalization to complex real anomalies. Removing positive pulling weakens the compactness constraint among normal neighbors, so the repulsion term from pseudo-anomalies is more likely to spread the normal samples apart, leading to elevated background scores or reduced threshold stability. RandomNegative and RandomPseudo illustrate the limitations of random negative samples and random pseudo-anomalies, respectively: in two-dimensional plots, they may sometimes appear clearly separated, but that separation often comes from ``easy differences'' inconsistent with the structure of normal reconstruction residuals, so the model may learn random perturbation artifacts rather than the true boundary between bearing anomalies and the normal manifold.

Therefore, this ablation study indicates that pseudo-anomaly quality mainly depends on three factors: first, the pseudo-anomalies should be located in the outer-neighbor region relative to the normal manifold rather than arbitrarily far away; second, their strength should cover multiple reconstruction-error intervals rather than concentrate at a single difficulty level; and third, the normal samples themselves should remain compact, otherwise negative-sample repulsion is converted into elevated background scores. The proposed target-error control, error-bin balancing, and normal positive-pulling term jointly satisfy these conditions, which explains their greater stability compared with template-based injection and random injection.

\FloatBarrier

\subsubsection{Ablation II: shared encoder design and CARLA coverage relation}
\label{subsubsec:ablation_encoder}

\begin{figure}[H]
    \centering
    \includegraphics[width=0.78\linewidth,height=0.72\textheight,keepaspectratio]{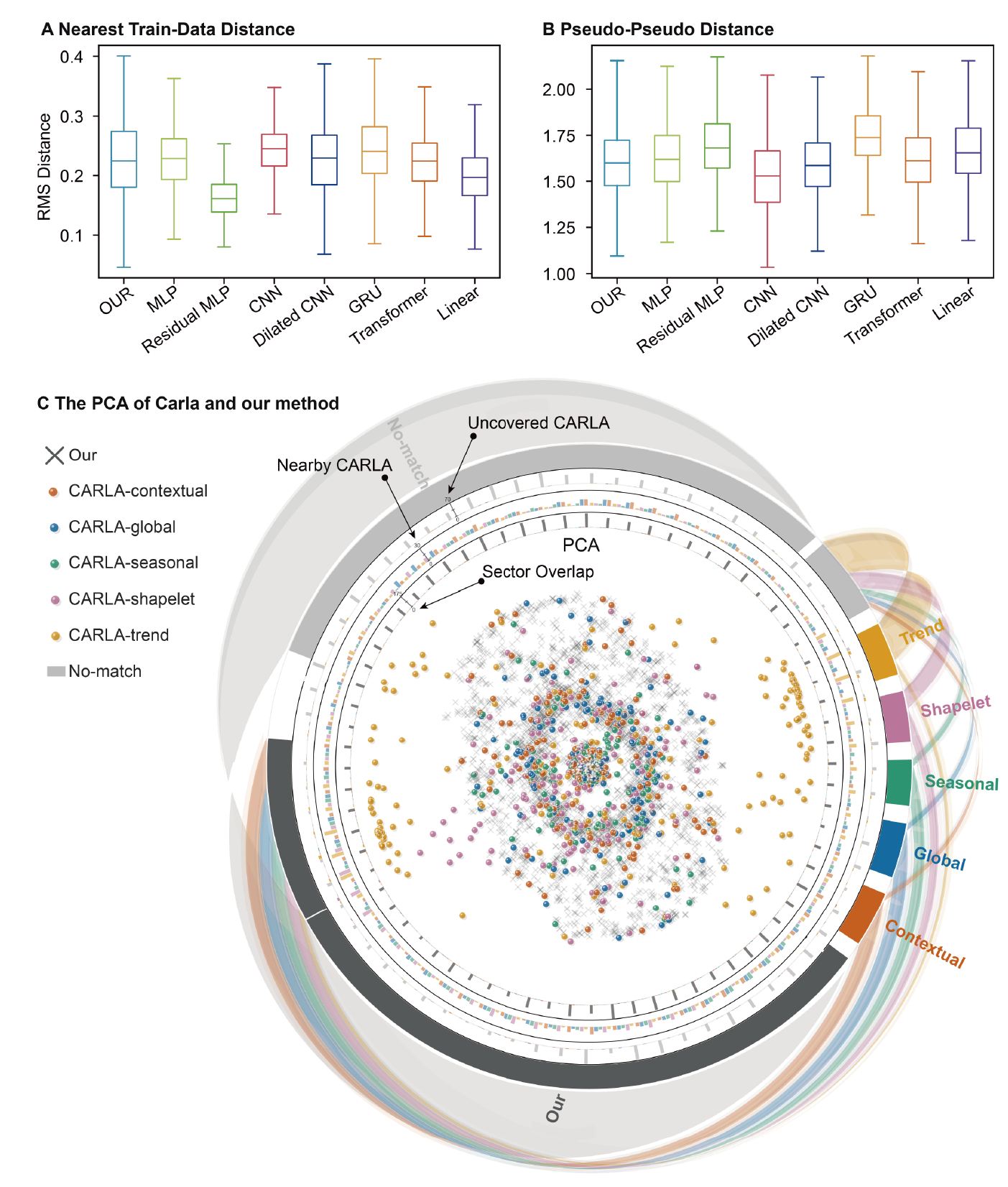}
    \caption{Ablation on the shared encoder design and the coverage relation with CARLA. Panels A and B show the nearest-training-sample distance and the intra-pseudo-anomaly distance statistics for pseudo-anomalies generated by different Stage~1 encoders, respectively. Panel C shows the difference between the proposed method and CARLA pseudo-anomalies in PCA space and sector-wise coverage relations.}
    \label{fig:ablation_encoder}
\end{figure}

Figure~\ref{fig:ablation_encoder} further analyzes the effect of the Stage~1 representation structure from two perspectives. The Nearest Train-Data Distance in Panel A measures the distance from a pseudo-anomaly to its nearest normal training sample and can be understood as whether the pseudo-anomaly is sufficiently close to the normal boundary. The Pseudo-Pseudo Distance in Panel B measures the dispersion among pseudo-anomalous samples and can be understood as whether the pseudo-anomalies have sufficient coverage. An effective pseudo-anomaly generator should not simply maximize both quantities: if the distance to the nearest normal sample is too small, pseudo-anomalies become almost indistinguishable from normal samples and can introduce label noise; if it is too large, pseudo-anomalies turn into easy negatives and fail to provide fine-grained boundary information. Likewise, if the intra-pseudo-anomaly distance is too small, coverage is insufficient; if it is too large, the sample distribution may become fragmented and lose boundary continuity.

The boxplots for different encoders show that the shared-encoder structure adopted in this paper achieves a relatively balanced trade-off between ``staying close to the normal boundary'' and ``maintaining pseudo-anomaly diversity.'' Structures such as residual MLP generate pseudo-anomalies that lie closer to normal training samples, suggesting that their perturbations remain closer to normality, but the insufficient boundary margin may weaken the contrastive training signal. GRU, CNN, or some more complex encoders can generate more dispersed pseudo-anomalies, but if they place pseudo-anomalies too far from the normal samples, they are more likely to produce easy negatives unrelated to real bearing degradation. In the proposed method, a shared encoder is combined with feature-specific decoding heads, allowing different sensor dimensions to learn normal patterns under a common temporal context while preserving the independence of per-feature reconstruction errors. The resulting pseudo-anomalies are thus constrained by the same normal manifold while still allowing each channel to deviate in a controllable way along its own residual direction, making them more suitable for subsequent contrastive learning in a unified embedding space.

The PCA and Sankey analyses in Fig.~\ref{fig:ablation_encoder} further indicate that the proposed method and CARLA are not simply ``two implementations of the same anomaly injection.'' CARLA's five types of injected samples occupy several semantically explicit sectors in PCA space, such as trend, shapelet, seasonal, global, and contextual directions. These directions are helpful for simulating typical time-series anomalies, but in bearing-vibration tasks they do not necessarily cover all residual boundaries outside the normal manifold. The samples generated by the proposed method use reconstruction error as the coordinate and do not assume specific fault templates. As a result, some regions in PCA space are adjacent to or overlap with CARLA, while others fall into no-match regions not covered by CARLA. The varying widths of the flows from different CARLA types into overlap, nearby, uncovered, and no-match regions in the Sankey diagram suggest that the two methods have both common and complementary coverage.

This observation has two implications. First, the proposed method covers some anomaly directions also represented by CARLA, which means that the pseudo-anomalies obtained by extrapolating reconstruction error are not semantically meaningless random perturbations; they bear some correspondence to anomaly types such as trends, shape changes, or contextual shifts. Second, the proposed method also covers regions that CARLA cannot easily enumerate explicitly, indicating that the anomaly boundary of bearings cannot be fully represented by a small set of predefined time-series transformations. In real bearing monitoring, faults may appear as compound changes in energy, frequency, impact sparsity, channel coupling, and condition disturbance. Using the upper tail of reconstruction errors on normal training samples as the basis for extrapolation makes it possible to build a boundary closer to the equipment's own data distribution without relying on fault templates. Therefore, the contribution of the proposed shared-encoder and error-control mechanism is not merely that they generate more pseudo-anomalies, but that they place these pseudo-anomalies at more reasonable boundary locations and with more reasonable coverage in the representation space.

\FloatBarrier

\subsubsection{Ablation III: Stage~2 contrastive loss and anomaly score}
\label{subsubsec:ablation_loss_score}

\begin{figure}[htbp]
    \centering
    \includegraphics[width=0.96\linewidth,height=0.50\textheight,keepaspectratio]{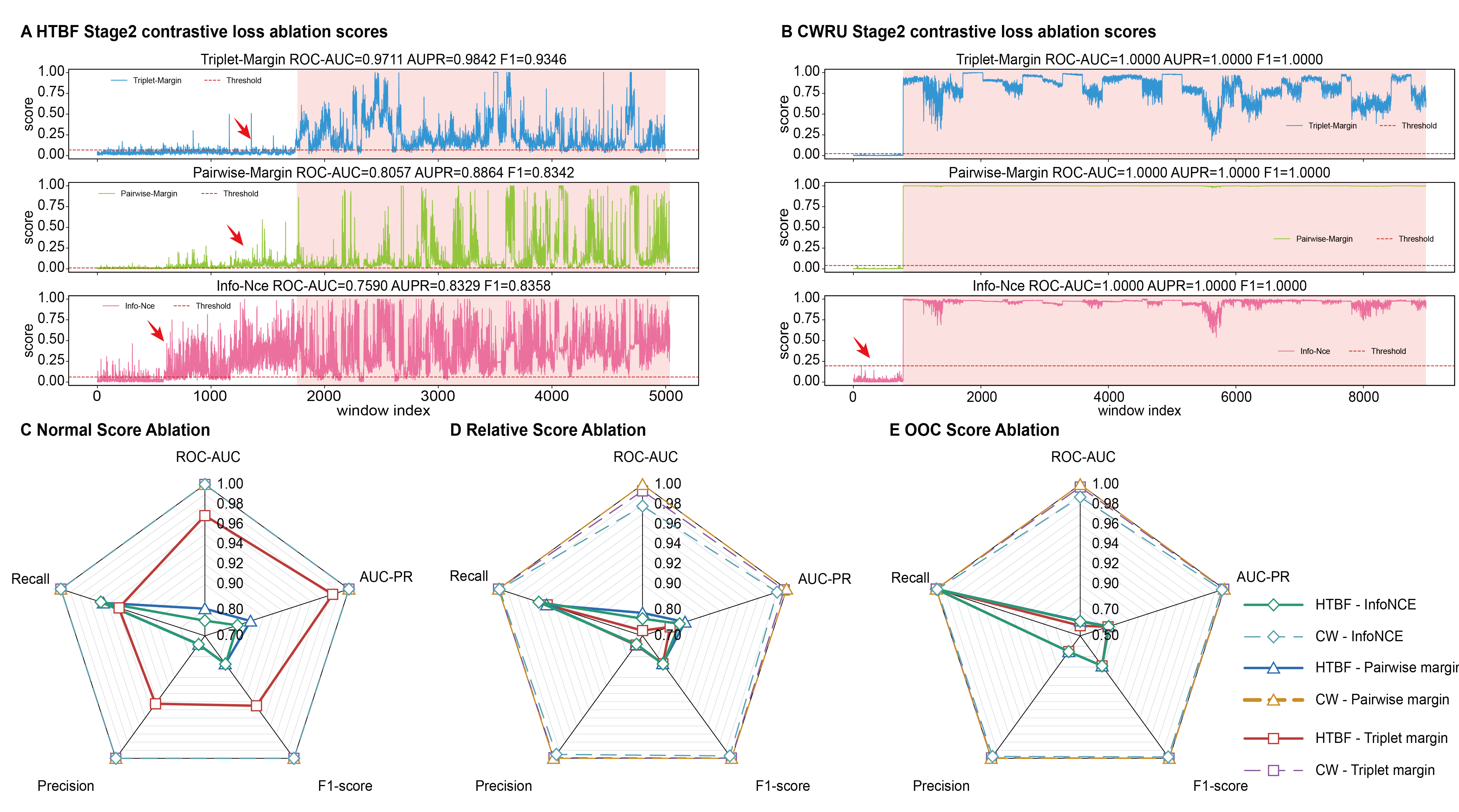}
    \caption{Ablation on the Stage~2 contrastive loss and anomaly score. Panels A and B show the anomaly-score curves under different contrastive losses on HTBF and CWRU, respectively. Panels C, D, and E compare the comprehensive metrics of the \emph{normal score}, \emph{relative score}, and \emph{OOC score} on the two datasets.}
    \label{fig:ablation_loss_score}
\end{figure}

Figure~\ref{fig:ablation_loss_score} compares three contrastive losses in Stage~2---Triplet-Margin, Pairwise-Margin, and InfoNCE---and further compares three anomaly scores, namely the \emph{normal score}, \emph{relative score}, and \emph{OOC score}. This experiment reveals an important phenomenon: pseudo-anomalies can serve as boundary references during training, but may not be suitable as ``real anomaly prototypes'' that directly participate in scoring at test time. Therefore, the loss function and the anomaly score must be consistent with the role played by pseudo-anomalies.

From the score curves over time, all three losses achieve nearly saturated detection performance on CWRU, with Triplet-Margin, Pairwise-Margin, and InfoNCE all reaching AUROC, AUPR, and best F1 of 1.0000. This indicates that the difference between normal and fault samples is strong in CWRU, so sufficiently separable representations can be learned even with different contrastive objectives. Therefore, CWRU is better suited to verifying that the model has basic detection capability than to distinguishing which Stage~2 loss is superior. The HTBF results are more informative: Triplet-Margin achieves AUROC=0.9711, AUPR=0.9842, and best F1=0.9346, clearly outperforming Pairwise-Margin (AUROC=0.8057, AUPR=0.8864, best F1=0.8342) and InfoNCE (AUROC=0.7590, AUPR=0.8329, best F1=0.8358). This difference shows that for multi-channel data under strong operating-condition disturbance, it is not enough simply to separate normal and pseudo-anomalous samples; what matters more is learning local relative distance relations.

Triplet-Margin has the advantage that it simultaneously constrains the relative distances among anchor, positive, and negative samples: the normal neighbor is required to be closer to the anchor than the pseudo-anomalous negative sample, with an explicit margin retained. This objective is highly consistent with the final \emph{normal score} adopted in this paper: during training, the local normal neighborhood is compacted and the boundary samples are pushed away; during testing, the anomaly degree is measured by the KNN distance from the test window to the normal sample bank. Pairwise-Margin mainly emphasizes absolute attraction or repulsion between sample pairs and lacks the ranking constraint among anchor, positive, and negative, which makes it more likely that the internal structure of the normal samples is not sufficiently compacted. Although pseudo-anomalies are pushed away, the ranking of real anomalies can remain unstable. InfoNCE relies on in-batch negatives and a temperature parameter; when the number of pseudo-anomalies is large and their difficulty is uneven, the loss may be dominated by many easy negatives. As a result, the model may learn a global separation trend, but the background scores of normal segments are more easily elevated, leading to decreases in Precision/best F1 and ranking metrics on HTBF.

The comparison among different anomaly scores supports the same explanation. The \emph{normal score} uses only the distance from a test window to the normal training sample bank and therefore directly answers the one-class anomaly-detection question ``Has this window deviated from the normal manifold?'' The \emph{relative score} uses distances to both the normal bank and the pseudo-anomalous bank, and the \emph{OOC score} further treats normal and pseudo-anomalous samples as two reference categories. These two scores work well on strongly separable data such as CWRU, but are less stable than the \emph{normal score} on complex data such as HTBF. The reason is that the main role of the Stage~1 pseudo-anomalies is to shape the boundary, not to exhaust all possible real anomaly categories. Real fault windows may deviate along some pseudo-anomaly directions, but may also deviate along directions not fully covered by CARLA or by the pseudo-anomalies generated in this paper. In that case, if the pseudo-anomaly bank is used as an anomaly prototype in scoring, the inconsistency between real anomalies and pseudo-anomalies can partially offset the deviation from the normal bank.

Therefore, the final choice in this paper is the combination of Triplet-Margin and the \emph{normal score}. This combination provides stronger consistency between the training objective and the test-time scoring: Triplet-Margin compacts the normal neighborhood and establishes a margin outside it, while the \emph{normal score} measures distance with respect to this compacted normal neighborhood. In this framework, pseudo-anomalies play the role of ``boundary-shaping samples'' rather than ``anomaly-category templates.'' This conclusion also explains why, in our framework, the value of generating high-quality pseudo-anomalies is mainly manifested in the representation-learning stage rather than in simply adding pseudo-anomalies to the test-time KNN classifier.

\FloatBarrier

\subsubsection{Ablation IV: RL step-size controller}
\label{subsubsec:ablation_rl_step_controller}

To further analyze the specific role of the reinforcement-learning step-size controller in pseudo-anomaly generation, we construct control experiments on CWRU and HTBF in which RL-based step control is removed. The two settings are: 1) \textbf{TPA-AD w/ RL step controller}, where the actor in Stage~1 adaptively controls the perturbation step size for continuous features according to the current reconstruction error, target error, and perturbation progress; and 2) \textbf{TPA-AD w/o RL step controller}, where the actor controller is removed and the perturbation magnitude is computed directly using an analytic target-ratio step-size strategy. To eliminate the effect of differences in the number of pseudo-anomalous samples, both settings collect a fixed number of $3000$ pseudo-anomalous windows with target-error bin filtering and pseudo-bin balancing turned off. Thus, the focus of this experiment is not ``which setting generates more pseudo-anomalies,'' but rather ``given the same number of pseudo-anomalous samples, does RL-based step control improve their effective diversity, source coverage, and boundary-crossing ability?''

\begin{figure}[htbp]
    \centering
    \includegraphics[width=0.98\linewidth,height=0.78\textheight,keepaspectratio]{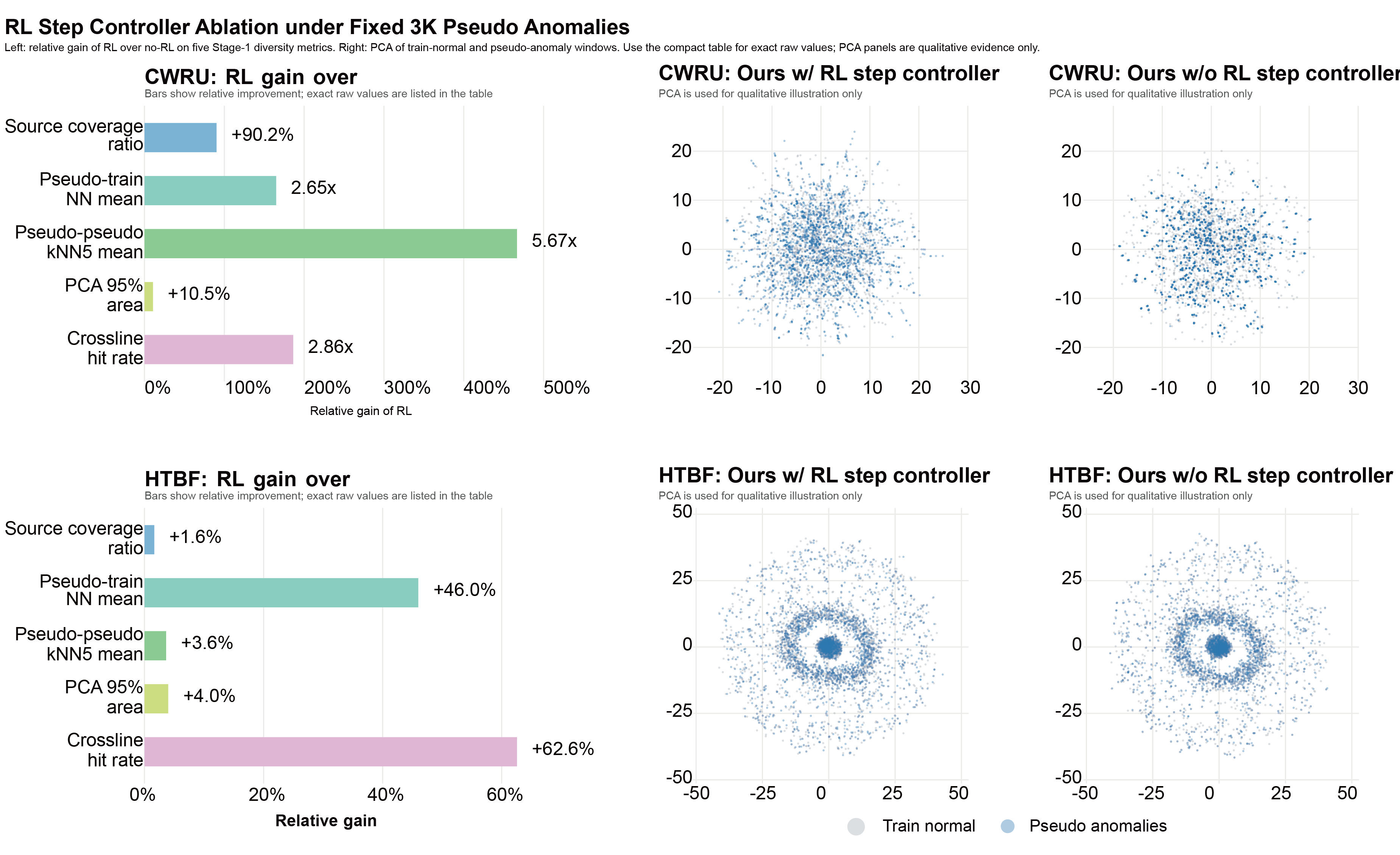}
    \caption{Ablation results for the RL step-size controller. The left column reports the relative gains of RL over no-RL on five Stage~1 diversity indicators, including source coverage, average distance from pseudo-anomalies to the nearest normal training samples, intra-pseudo-anomaly nearest-neighbor distance, PCA 95\% coverage area, and target-error boundary-crossing hit rate. The middle and right columns show the PCA distributions with and without the RL step-size controller, respectively. The figure shows that, under a fixed number of pseudo-anomalies, the main benefit of RL step control lies in improving the dispersion of boundary samples, their separation from the normal manifold, and their ability to cross the target-error boundary. It should be noted that the PCA plots are used only for qualitative visualization and not as standalone quantitative evidence.}
    \label{fig:ablation_rl_step_controller}
\end{figure}

\begin{table}[htbp]
    \centering
    \caption{Compact metrics for the RL step-size-controller ablation. Each cell reports the raw values as ``RL / No-RL,'' and the better RL value is highlighted in bold.}
    \label{tab:ablation_rl_step_controller}
    \scriptsize
    \renewcommand{\arraystretch}{1.12}
    \setlength{\tabcolsep}{3pt}
    \resizebox{\linewidth}{!}{
    \begin{tabular}{@{}lccccc@{}}
        \toprule
        Dataset & \makecell[c]{Source\\coverage} & \makecell[c]{Pseudo-anomaly--\\normal NN distance} & \makecell[c]{Intra-pseudo-anomaly\\NN distance} & \makecell[c]{PCA 95\%\\area} & \makecell[c]{Boundary-crossing\\hit rate} \\
        \midrule
        CWRU & \textbf{0.735} / 0.386 & \textbf{17.296} / 6.528 & \textbf{18.033} / 3.183 & \textbf{955.4} / 864.3 & \textbf{0.760} / 0.265 \\
        HTBF & \textbf{0.775} / 0.762 & \textbf{13.938} / 9.547 & \textbf{55.339} / 53.401 & \textbf{3157.0} / 3036.0 & \textbf{0.347} / 0.214 \\
        \bottomrule
    \end{tabular}}
\end{table}

The left column of Fig.~\ref{fig:ablation_rl_step_controller} reports indicators with different units in terms of relative gains of RL over no-RL, making it easier to compare the magnitude of improvement across indicators. Table~\ref{tab:ablation_rl_step_controller} gives the corresponding raw values. Overall, the RL step-size controller improves not the number of pseudo-anomalous samples, but their effective diversity and boundary expression ability.

Taking CWRU as an example, the RL step-size controller brings clear improvements on all five indicators. Specifically, source coverage increases from 0.386 to 0.735, indicating that pseudo-anomalous samples come from a broader variety of normal training windows instead of repeatedly perturbing only a few samples. The average distance from pseudo-anomalies to their nearest normal training samples increases from 6.528 to 17.296, showing that the RL-controlled pseudo-anomalies deviate more fully from the normal manifold. The intra-pseudo-anomaly nearest-neighbor distance increases from 3.183 to 18.033, indicating greater dispersion and lower redundancy among samples. The boundary-crossing hit rate increases from 0.265 to 0.760, showing that RL step control makes it easier for samples to cross the target-error boundary. In relative terms, the intra-pseudo-anomaly nearest-neighbor distance on CWRU improves by about 5.67 times, the boundary-crossing hit rate by about 2.86 times, the pseudo-anomaly--normal nearest-neighbor distance by about 2.65 times, and the source coverage by about 90.2\%. These results jointly show that on more standard bearing-fault data, the RL step-size controller can substantially enhance the coverage, dispersion, and boundary character of pseudo-anomalous samples.

The conclusion on HTBF is consistent with that on CWRU, although the gains are distributed more unevenly across indicators. Table~\ref{tab:ablation_rl_step_controller} shows that source coverage on HTBF increases only slightly from 0.762 to 0.775, and the improvements in intra-pseudo-anomaly nearest-neighbor distance and PCA 95\% area are also relatively moderate. By contrast, the boundary-crossing hit rate increases from 0.214 to 0.347, and the pseudo-anomaly--normal nearest-neighbor distance increases from 9.547 to 13.938, corresponding to relative gains of about 62.6\% and 46.0\%, respectively. This suggests that in more complex multi-channel scenarios, the main benefit of the RL step-size controller is not necessarily to cover more training-source windows, but rather to push pseudo-anomalies away from the normal manifold more effectively and improve their ability to cross the target-error boundary. In other words, on HTBF the role of the RL step-size controller is more about improving the ``effective strength'' of boundary samples than substantially expanding source coverage.

The PCA results in Fig.~\ref{fig:ablation_rl_step_controller} provide intuitive support for these findings. On CWRU, after introducing the RL step-size controller, pseudo-anomalous samples expand over a wider region around the normal samples and become more dispersed. Without RL control, pseudo-anomalies still deviate from the normal region, but their overall coverage is narrower and local clustering is more obvious. On HTBF, both methods can form a relatively clear ring-like or outward-expanding distribution, but the RL-controlled pseudo-anomalies cover the outer-edge regions and cross-boundary directions more fully, whereas the no-RL version tends to concentrate in a few relatively fixed regions. It should again be emphasized that PCA visualizations are used only to qualitatively illustrate pseudo-anomaly distribution trends; the conclusions are still primarily supported by the quantitative indicators in Table~\ref{tab:ablation_rl_step_controller}.

Overall, under a fixed number of pseudo-anomalous samples, the RL step-size controller mainly improves the separation of pseudo-anomalies from the normal manifold, their internal dispersion, and their ability to cross the target-error boundary, while on some datasets it further increases source coverage. This indicates that its main contribution lies in improving the effective diversity of boundary samples rather than simply increasing the number of pseudo-anomalous samples.

\FloatBarrier

\subsubsection{Ablation V: empirical basis of the Stage~1 target reconstruction-error interval}
\label{subsubsec:ablation_recon_error_threshold}

Stage~1 requires a pre-specified target reconstruction-error interval to control the deviation strength of pseudo-anomalous windows relative to normal windows. If this interval is too low, the generated pseudo-anomalies may still remain close to the normal manifold and fail to form an effective boundary. If it is too high, the generated pseudo-anomalies are more likely to lie too far away from normal samples, weakening the subsequent representation-learning constraint toward real degradation directions. To examine whether the interval setting has an empirical basis, we further analyze the distributions of window-level reconstruction errors for normal training windows (Train-N), validation-normal windows (Val-N), and validation fault/degradation windows (Val-F). The purpose here is not to prove that reconstruction error itself is the final anomaly detector, but rather to verify whether high-quantile statistics of reconstruction errors on normal training windows can serve as an empirical reference for the target error strength used in Stage~1 pseudo-anomaly generation.

\begin{figure}[htbp]
    \centering
    \includegraphics[width=0.98\linewidth,height=0.78\textheight,keepaspectratio]{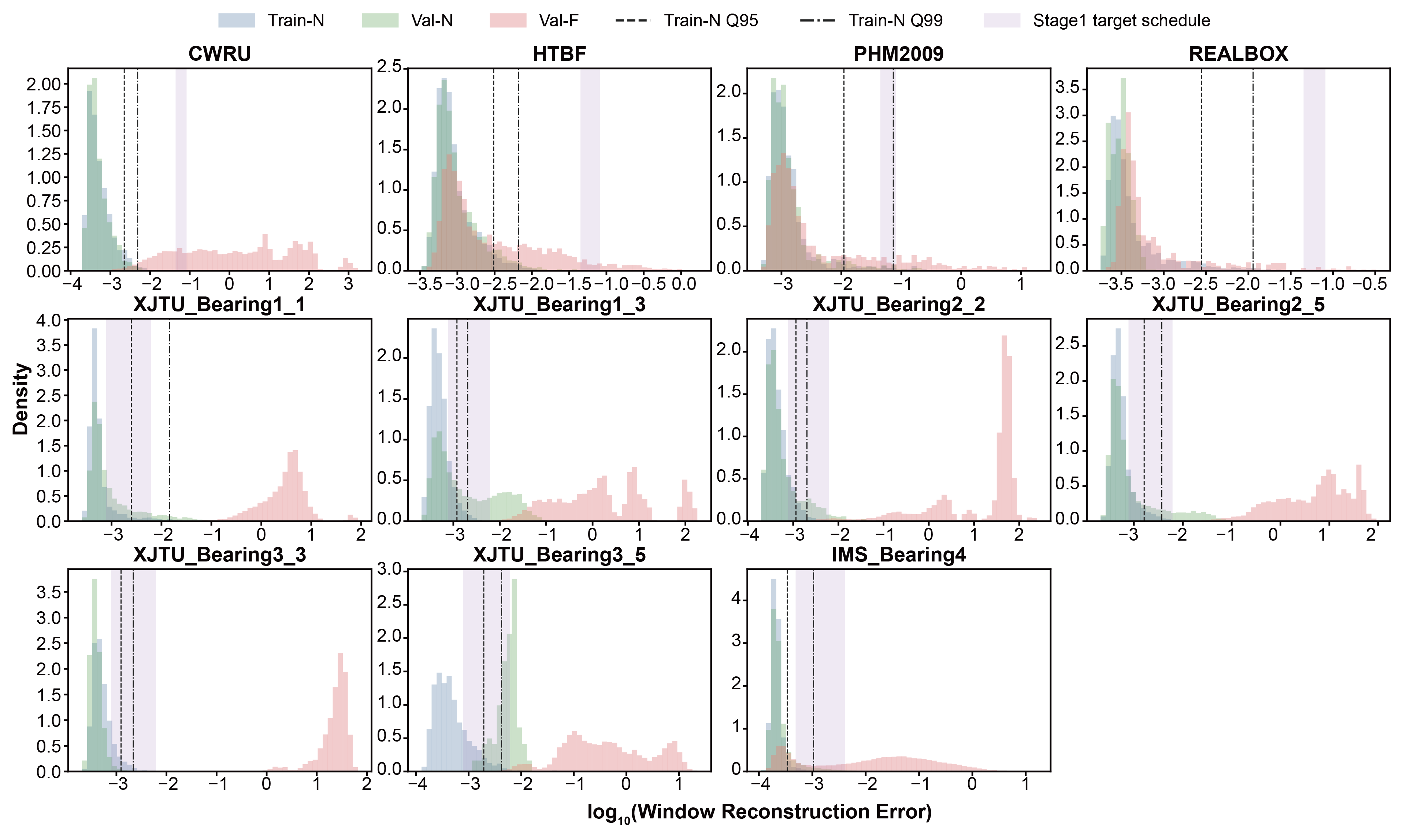}
    \caption{Analysis of the empirical basis of the Stage~1 target reconstruction-error interval. The figure compares the window-level reconstruction-error distributions of Train-N, Val-N, and Val-F on different datasets. The black dashed/dash-dotted lines denote the Train-N $Q95$/$Q99$, and the purple shaded region denotes the current Stage~1 target-error interval. The figure illustrates that the high-quantile region of normal-training reconstruction error can serve as an empirical reference for the pseudo-anomaly target error on some bearing datasets, but this reference is neither a unified cross-dataset threshold nor the final anomaly decision threshold.}
    \label{fig:ablation_recon_error_threshold}
\end{figure}

For each dataset, we reuse the trained Stage~1 reconstruction model and compute window-level reconstruction errors under the same window settings ($L=512$, $H=256$). For multi-channel data, the maximum channel-wise error is used as the window error to avoid masking local fault responses through channel averaging. In Fig.~\ref{fig:ablation_recon_error_threshold}, the blue, green, and pink histograms denote the reconstruction-error distributions of Train-N, Val-N, and Val-F, respectively. The black dashed and dash-dotted lines denote the $Q95$ and $Q99$ of Train-N, respectively, and the purple shaded region denotes the target-error interval currently used for pseudo-anomaly generation in Stage~1. The horizontal axis is $\log_{10}(\text{window reconstruction error})$, so a rightward shift indicates a global increase in window reconstruction error.

\begin{table}[htbp]
    \centering
    \caption{Compact results for the reconstruction-error target-interval ablation. Train-N $Q99$ denotes the high-quantile reference of reconstruction error on normal training windows; Val-N FPR@Train-N $Q99$ denotes the proportion of validation-normal windows lying to the right of the Train-N $Q99$ reference line; Val-F $Q10$/Train-N $Q99$ measures the rightward shift of the lower quantile of fault/degradation windows relative to the high quantile of normal training windows; and Val-F below Train-N $Q99$ denotes the proportion of fault/degradation windows below Train-N $Q99$. This table is used only to analyze the empirical basis of the Stage~1 target-error interval and does not correspond to a final deployment threshold.}
    \label{tab:ablation_recon_error_threshold}
    \small
    \renewcommand{\arraystretch}{1.08}
    \setlength{\tabcolsep}{4pt}
    \resizebox{\linewidth}{!}{
    \begin{tabular}{lcccc}
        \toprule
        Dataset & Train-N $Q99$ & Val-N FPR@Train-N $Q99$ (\%) & Val-F $Q10$/Train-N $Q99$ & Val-F below Train-N $Q99$ (\%) \\
        \midrule
        CWRU & 0.0049 & 0.38 & 3.46 & 2.66 \\
        HTBF & 0.0066 & 1.88 & 0.10 & 72.00 \\
        PHM2009 & 0.0735 & 1.15 & 0.01 & 85.79 \\
        REALBOX & 0.0114 & 0.00 & 0.03 & 92.83 \\
        XJTU-SY Bearing1-1 & 0.0145 & 4.74 & 42.11 & 1.14 \\
        XJTU-SY Bearing1-3 & 0.0020 & 41.44 & 36.03 & 0.00 \\
        XJTU-SY Bearing2-2 & 0.0020 & 12.18 & 166.41 & 0.22 \\
        XJTU-SY Bearing2-5 & 0.0037 & 14.56 & 108.05 & 0.00 \\
        XJTU-SY Bearing3-3 & 0.0022 & 0.82 & 4475.29 & 0.00 \\
        XJTU-SY Bearing3-5 & 0.0042 & 74.80 & 11.98 & 0.16 \\
        IMS Bearing4 & 0.0011 & 4.38 & 0.25 & 27.60 \\
        \bottomrule
    \end{tabular}}
\end{table}

Table~\ref{tab:ablation_recon_error_threshold} and Fig.~\ref{fig:ablation_recon_error_threshold} jointly indicate that on CWRU and on most XJTU-SY degradation samples, the reconstruction errors of fault/degradation windows shift clearly to the right relative to the high quantiles of the normal training windows. Taking CWRU as an example, the $10$th-percentile error of Val-F is about $3.46$ times the $99$th-percentile error of Train-N, and only $2.66\%$ of fault windows lie below Train-N $Q99$; meanwhile, Val-N FPR@Train-N $Q99$ is only $0.38\%$, suggesting that the high quantiles of normal training errors also provide a relatively stable conservative reference for normal windows on this dataset. The phenomenon is even more pronounced for most XJTU-SY samples. For example, Val-F $Q10$/Train-N $Q99$ reaches $42.11$, $166.41$, and $4475.29$ for Bearing1-1, Bearing2-2, and Bearing3-3, respectively, while the corresponding Val-F below Train-N $Q99$ values are only $1.14\%$, $0.22\%$, and $0.00\%$. This suggests that the upper tail of normal-training reconstruction errors can naturally serve as a reference for pseudo-anomaly strength.

On the other hand, HTBF, PHM2009, and REALBOX provide clear counterexamples. The Val-F $Q10$/Train-N $Q99$ value of HTBF is only $0.10$, and $72.00\%$ of fault windows lie below Train-N $Q99$; the corresponding proportions for PHM2009 and REALBOX are even higher, reaching $85.79\%$ and $92.83\%$, respectively. This indicates that in these datasets there are many real fault windows with low reconstruction errors, so the high quantiles of normal training errors cannot be interpreted as a unified final anomaly decision threshold. At the same time, Val-N FPR@Train-N $Q99$ varies substantially across datasets. For example, it reaches $41.44\%$ and $74.80\%$ on XJTU-SY Bearing1-3 and Bearing3-5, respectively, further showing that the role of Train-N $Q99$ is better understood as an empirical reference line for Stage~1 rather than a stable deployment threshold across datasets.

This is consistent with the two-stage design of the proposed method. The target-error interval in Stage~1 is not directly used for the final decision; instead, it is used to generate pseudo-anomalous windows with controlled deviation relative to normal samples, thereby establishing a boundary band outside the normal manifold for representation learning. The final anomaly score is still determined by the KNN distance to the normal sample bank in the Stage~2 representation space. Therefore, this experiment supports a more restrained conclusion: in bearing data, high-quantile statistics of reconstruction errors on normal training windows can provide an empirical reference for the Stage~1 pseudo-anomaly target-error interval, thereby reducing the need to completely handcraft the pseudo-anomaly strength. However, because fault patterns, operating-condition disturbances, and degradation processes vary substantially across datasets, this reference is neither a unified theoretical threshold across datasets nor the final anomaly decision threshold. The IMS Bearing4 result in Fig.~\ref{fig:ablation_recon_error_threshold} can be regarded as a supplementary observation, but should not be used as the main argument to further strengthen the conclusion above.

\FloatBarrier

\subsubsection{Ablation VI: CARLA-style pseudo-anomalies versus the original Stage~2 logic}
\label{subsubsec:ablation_carla_baseline_comparison}

To further examine whether the performance of the proposed method comes merely from external pseudo-anomalous samples and whether CARLA's original Stage~2 logic is suitable for bearing scenarios, we add two groups of CARLA-related comparison experiments. In the first group, the pseudo-anomaly generation in Stage~1 is replaced with CARLA-style pseudo-anomalous windows, while the representation learning in Stage~2 and the anomaly score based on the KNN distance to normal samples are kept unchanged; this setting is denoted as \emph{CARLA pseudo + ours Stage~2}. In the second group, an adapted version of CARLA's original classification/clustering-based Stage~2 is further adopted under the same training-normal-window and test-window protocol; this setting is denoted as \emph{CARLA original Stage~2 adapter}. These two control settings are used, respectively, to test whether external pseudo-anomalous samples can directly replace the target-error-controlled generation strategy of this paper, and whether CARLA's own Stage~2 discrimination logic can stably replace the normal-sample KNN representation score used in this paper.

\begin{figure}[htbp]
    \centering
    \includegraphics[width=0.96\linewidth,height=0.82\textheight,keepaspectratio]{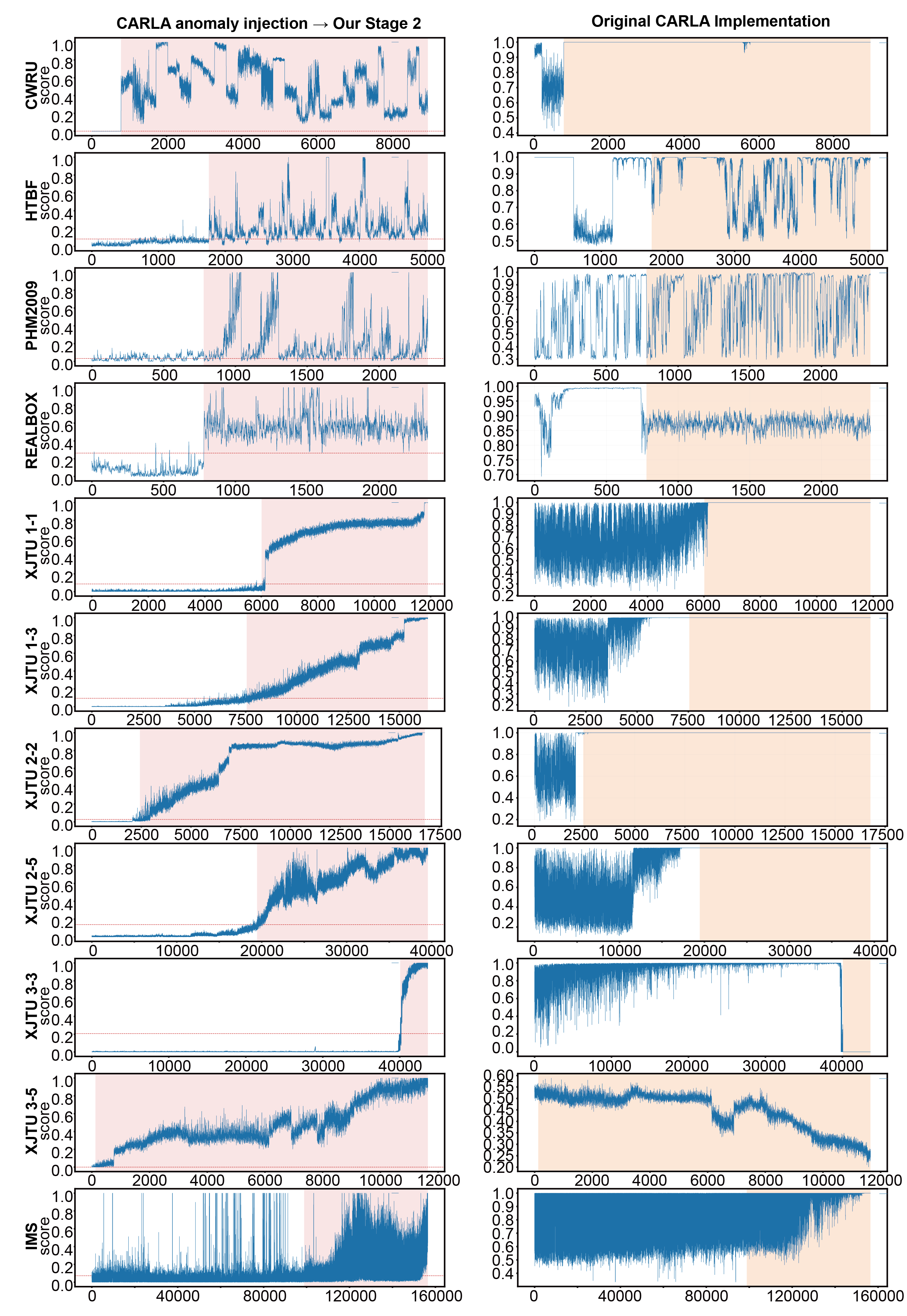}
    \caption{Comparison of score curves for CARLA-style pseudo-anomaly injection and the original CARLA Stage~2 logic on different bearing datasets. The left column shows CARLA anomaly injection $\rightarrow$ Ours Stage~2, i.e., only the pseudo-anomaly generation is replaced while keeping our Stage~2 representation learning and the \emph{normal score} (the anomaly score based on KNN distance to normal samples). The right column shows the CARLA original Stage~2 adapter, i.e., the adapted version using CARLA's original classification/clustering-based Stage~2 logic. Background shading indicates anomalous or degradation stages, and the curves qualitatively illustrate the differences in score stability between the two CARLA-related schemes on different datasets.}
    \label{fig:carla_baseline_comparison}
\end{figure}

\begin{table}[htbp]
    \centering
    \caption{Comparison of the proposed method and two CARLA-related controls on $11$ bearing datasets. Best F1 is a posterior separability metric obtained by scanning thresholds on the test scores and should not be interpreted as deployment-threshold performance. Bold indicates the best result on the same dataset and metric based on the original unrounded values.}
    \label{tab:carla_baseline_comparison}
    \small
    \renewcommand{\arraystretch}{1.06}
    \setlength{\tabcolsep}{3.5pt}
    \resizebox{\linewidth}{!}{
    \begin{tabular}{lccccccccc}
        \toprule
        \multirow{2}{*}{Dataset} & \multicolumn{3}{c}{TPA-AD} & \multicolumn{3}{c}{CARLA pseudo + ours Stage~2} & \multicolumn{3}{c}{CARLA original Stage~2 adapter} \\
        \cmidrule(lr){2-4}\cmidrule(lr){5-7}\cmidrule(lr){8-10}
        & AUROC & AUPR & Best F1 & AUROC & AUPR & Best F1 & AUROC & AUPR & Best F1 \\
        \midrule
        CWRU & \textbf{1.0000} & \textbf{1.0000} & \textbf{1.0000} & \textbf{1.0000} & \textbf{1.0000} & \textbf{1.0000} & 0.9998 & 1.0000 & 0.9986 \\
        HTBF & \textbf{0.9711} & \textbf{0.9842} & \textbf{0.9347} & 0.9665 & 0.9841 & 0.9344 & 0.4905 & 0.6474 & 0.8149 \\
        PHM2009 & \textbf{0.7522} & \textbf{0.8751} & 0.8028 & 0.7223 & 0.8245 & 0.7614 & 0.6486 & 0.7758 & \textbf{0.8115} \\
        REALBOX & 0.9996 & 0.9998 & \textbf{0.9981} & \textbf{0.9999} & \textbf{0.9999} & 0.9978 & 0.0910 & 0.4711 & 0.8048 \\
        XJTU-SY Bearing1-1 & 0.9980 & 0.9984 & 0.9902 & \textbf{0.9982} & \textbf{0.9985} & \textbf{0.9904} & 0.9977 & 0.9982 & 0.9895 \\
        XJTU-SY Bearing1-3 & 0.9977 & \textbf{0.9981} & 0.9776 & 0.9973 & 0.9978 & 0.9756 & \textbf{0.9978} & 0.9981 & \textbf{0.9784} \\
        XJTU-SY Bearing2-2 & 0.9979 & 0.9996 & 0.9874 & 0.9981 & 0.9997 & 0.9886 & \textbf{0.9986} & \textbf{0.9998} & \textbf{0.9909} \\
        XJTU-SY Bearing2-5 & \textbf{0.9999} & \textbf{0.9999} & \textbf{0.9970} & 0.9999 & 0.9999 & 0.9957 & 0.9990 & 0.9990 & 0.9849 \\
        XJTU-SY Bearing3-3 & \textbf{1.0000} & \textbf{0.9999} & \textbf{0.9972} & 1.0000 & 0.9998 & 0.9969 & 0.0000 & 0.0814 & 0.1518 \\
        XJTU-SY Bearing3-5 & 0.9941 & 0.9999 & 0.9959 & \textbf{0.9956} & \textbf{0.9999} & \textbf{0.9969} & 0.1071 & 0.9697 & 0.9945 \\
        IMS Bearing4 & \textbf{0.9297} & \textbf{0.9011} & \textbf{0.8376} & 0.8828 & 0.7845 & 0.7699 & 0.8845 & 0.8741 & 0.7723 \\
        \bottomrule
    \end{tabular}}
\end{table}

Table~\ref{tab:carla_baseline_comparison} and Fig.~\ref{fig:carla_baseline_comparison} together show that \emph{CARLA pseudo + ours Stage~2} clearly outperforms \emph{CARLA original Stage~2 adapter} overall, indicating that simply replacing our Stage~2 with CARLA's original classification/clustering-style logic does not work stably under the bearing-window protocol used in this paper. In terms of grouped average results, \emph{CARLA pseudo + ours Stage~2} achieves average AUROC/AUPR/best F1 of $0.9222/0.9521/0.9234$ on fault detection and $0.9817/0.9686/0.9591$ on degradation detection, whereas \emph{CARLA original Stage~2 adapter} reaches only $0.5575/0.7236/0.8575$ and $0.7121/0.8457/0.8375$, respectively. This difference is also visually evident in the score curves. For example, on CWRU and REALBOX, the left column forms relatively clear anomaly boundaries, whereas the right column is more prone to sustained high platforms, compressed scores, or unstable ranking inside anomalous intervals.

Looking further at the relation between \emph{CARLA pseudo + ours Stage~2} and the proposed method, one can see that external pseudo-anomalous samples can indeed provide useful boundary information on some datasets, but their benefit is clearly dataset-dependent. This scheme performs well on CWRU, REALBOX, and most XJTU-SY degradation samples, indicating that CARLA-style pseudo-anomalies are not entirely ineffective. However, the proposed method remains superior on HTBF, PHM2009, and IMS Bearing4. For example, on HTBF, TPA-AD achieves AUROC/AUPR/best F1 of $0.9711/0.9842/0.9347$, whereas \emph{CARLA pseudo + ours Stage~2} obtains $0.9665/0.9841/0.9344$; on PHM2009, TPA-AD obtains $0.7522/0.8751/0.8028$, while \emph{CARLA pseudo + ours Stage~2} drops to $0.7223/0.8245/0.7614$; and on IMS Bearing4, TPA-AD achieves $0.9297/0.9011/0.8376$, while \emph{CARLA pseudo + ours Stage~2} decreases to $0.8828/0.7845/0.7699$. These results indicate that the performance of the proposed method does not simply come from ``introducing arbitrary pseudo-anomalous samples''; rather, the specific construction of pseudo-anomalies and their deviation strength relative to the normal manifold still significantly affect the subsequent representation learning.

On the other hand, \emph{CARLA original Stage~2 adapter} can achieve relatively high results on CWRU and some XJTU-SY samples, but it shows clear instability on multiple datasets. For example, its AUROC is only $0.0910$ on REALBOX, nearly $0$ on XJTU-SY Bearing3-3, and only $0.1071$ on XJTU-SY Bearing3-5. Correspondingly, the right column of Fig.~\ref{fig:carla_baseline_comparison} shows obvious score instability, score flattening, or ranking inversion on these datasets. This indicates that CARLA's original classification/clustering-style Stage~2 logic is more sensitive to dataset structure, anomaly proportion, and degradation process, and cannot be stably transferred to normal-only bearing scenarios. Taken together, these two CARLA-related controls support a more robust attribution of the advantage of the proposed method to the combination of controlled pseudo-anomaly generation under target-error constraints and the Stage~2 representation score based on KNN distances to normal samples, rather than to a generic recipe of ``arbitrary pseudo-anomalies + arbitrary Stage~2.''

\FloatBarrier

\subsection{Hyperparameter sensitivity analysis}
\label{subsec:hyperparameter}

The target-error interval in Stage~1 is one of the more important empirical hyperparameters of the proposed method. As defined in the method section, we construct the target interval as \(\tau_j^{\mathrm{low}}=Q_{q_l}(\{e_{i,j}^{\mathrm{train}}\}_i)\) and \(\tau_j^{\mathrm{high}}=Q_{q_u}(\{e_{i,j}^{\mathrm{train}}\}_i)\) from the per-feature reconstruction-error distribution of the normal training samples, and then sample target errors within this interval to generate pseudo-anomalies. It should be emphasized that \(q_l\) and \(q_u\) are not fault-alarm thresholds used at deployment time; rather, they are unsupervised empirical parameters used in Stage~1 to construct a pseudo-anomaly boundary band. They depend only on the reconstruction-error distribution of the normal training data.

\begin{figure}[htbp]
    \centering
    \includegraphics[width=0.96\linewidth,height=0.52\textheight,keepaspectratio]{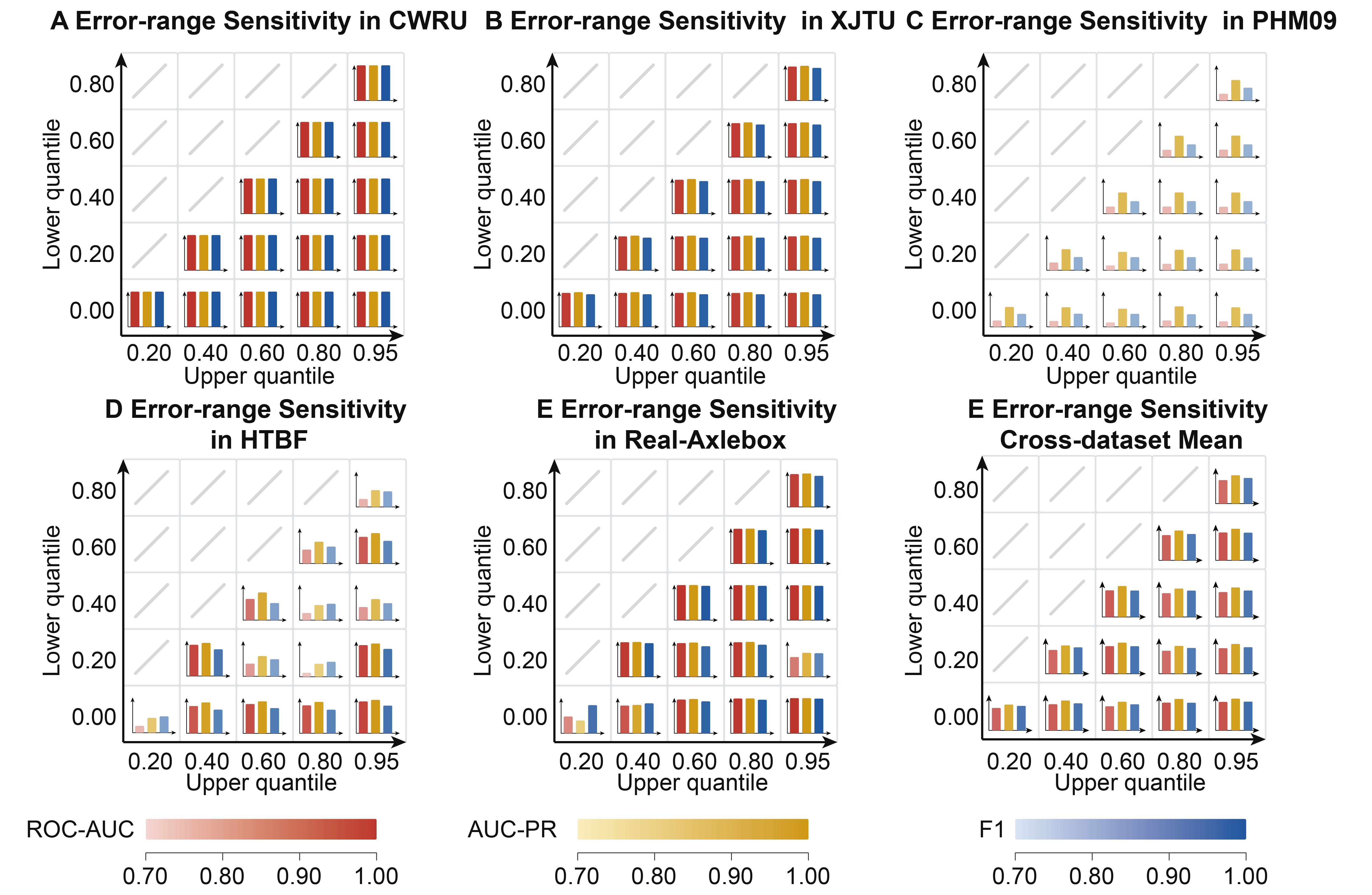}
    \caption{Hyperparameter sensitivity analysis of the target-error quantile interval. Each subplot shows the AUROC, AUPR, and best-F1 results obtained by different combinations of the lower quantile \(q_l\) and upper quantile \(q_u\) on CWRU, XJTU-SY, PHM2009, HTBF, REALBOX, and the cross-dataset average.}
    \label{fig:hyperparameter_sensitivity}
\end{figure}

Figure~\ref{fig:hyperparameter_sensitivity} shows the effect of different combinations of lower and upper quantiles on AUROC, AUPR, and best F1. Overall, the role of the target-error interval can be understood from two dimensions: boundary strength and boundary coverage. A relatively low upper quantile makes the pseudo-anomalies too close to the main body of the normal reconstruction-error distribution. Although such samples are difficult, they often lack sufficient anomaly magnitude and can easily treat normal operating fluctuations as negative samples, ultimately resulting in insufficient margin between normal and anomalous states. By contrast, an excessively high and overly narrow interval concentrates pseudo-anomalies in the extreme tail of the error distribution. Although such pseudo-anomalies can serve as strong negatives, they may sacrifice boundary continuity and make the model overly biased toward strong anomalies while weakening its response to early degradation or complex weak anomalies. Therefore, a stable target interval usually needs to satisfy both of the following: the upper bound should be sufficiently high to ensure that pseudo-anomalies leave the normal core, while the lower bound should not be too high so that pseudo-anomalies can cover multiple boundary levels ranging from weak shifts to strong shifts.

The sensitivity differences across datasets further indicate that this hyperparameter is related to the intrinsic fault separability and operating-condition complexity of the data. CWRU, XJTU-SY, and REALBOX maintain high metrics under many valid quantile combinations, suggesting that these datasets contain relatively clear amplitude or degradation differences between normal and anomalous states. As long as Stage~1 can generate pseudo-anomalies outside the normal region, Stage~2 can produce stable separation. PHM2009 and HTBF are more sensitive to the target-error interval. PHM2009 represents a gearbox compound-fault scenario in which real anomalies and normal samples are locally entangled, while HTBF contains both multi-channel coupling and operating-condition disturbance. For such data, overly low target-error intervals easily generate pseudo-anomalies that still resemble normal data, leading to elevated score backgrounds. If only extremely high quantiles are used, pseudo-anomalies may become easy negatives along only a few directions and fail to cover the multiple deviation modes of real anomalies. Therefore, a moderately wide interval with a mid-to-high upper quantile is usually more robust.

From the cross-dataset average, the better-performing region is not an isolated single point but rather a stable area composed of medium lower quantiles and relatively high upper quantiles. This observation supports the empirical thresholding strategy adopted in this paper: for bearing vibration data, the upper tail of the reconstruction-error distribution of normal training windows can be treated as a range of ``candidate anomaly strengths near the normal boundary,'' and bin-wise sampling can be carried out within this region instead of manually setting perturbations with fixed amplitudes. The advantage of this strategy is that reconstruction-error values can vary greatly across datasets, channels, and scales, whereas quantiles are relatively comparable. In addition, upper-tail quantiles naturally include the hardest-to-reconstruct normal operating samples, which helps generate difficult pseudo-anomalies closer to early degradation.

Based on the above results, we provide the following practical interpretation for bearing scenarios. If a dataset exhibits clear fault differences and relatively stable normal operating conditions, a relatively broad mid-to-high quantile interval can yield stable results. If strong operating-condition disturbances or multi-channel coupling are present, overly low upper quantiles should be avoided, and the lower quantile can be moderately increased to reduce overlap with the main normal region. If the goal is to enhance sensitivity to early degradation, one should not rely exclusively on extreme high quantiles; some lower-strength boundary samples should also be retained. In other words, the role of the target-error interval in Stage~1 is not to locate the final anomaly decision line, but to define the ``training boundary band'' for pseudo-anomaly generation. Figure~\ref{fig:hyperparameter_sensitivity} shows that this boundary band has a reasonably wide usable range across multiple bearing datasets, suggesting that the proposed method is not overly dependent on the target-error quantiles. However, on more complex datasets such as PHM2009 and HTBF, a proper choice of a mid-to-high quantile interval can still significantly improve boundary quality and final detection stability.

\FloatBarrier

\section{Discussion and extended analysis}
\label{subsec:tsad_extension}
\FloatBarrier

As a discussion-oriented extension, we further examine the broader applicability of the proposed method on public time-series anomaly detection datasets. This part does not replace the main conclusions drawn from the bearing fault-detection and degradation-detection experiments above; rather, it supplements them by observing how the framework transfers to non-bearing data. To this end, we select $13$ representative public datasets from the TSB-AD benchmark~\cite{liu2024elephant} and adopt the data organization scheme of TSB-UAD~\cite{paparrizos2022tsbuad} for the univariate sequences. The selected data include both univariate sequences converted from classification datasets and multivariate sequences from spacecraft telemetry, wearable sensing, Web-service KPIs, industrial system monitoring, and cluster runtime logs. The corresponding anomaly types include point anomalies, interval anomalies, trend drifts, and pattern switches.

\subsubsection{Experimental setup for the TSAD extension}
\label{subsubsec:tsad_extension_setup}

The TSAD extension analysis uses $13$ public data sources or subsets from the TSB-AD benchmark, namely UCR, YAHOO, WSD, CATSv2, Daphnet, Exathlon, NEK, IOPS, LTDB, MSL, SMAP, SMD, and PSM. Unlike the manually constructed normal/fault concatenated sequences used in the bearing fault-detection experiments, the TSAD data mainly follow the existing training segments, test segments, and point-level anomaly labels provided by the public data sources. If an official training/test split is available, we adopt it directly; otherwise, the training and test segments are constructed according to the fixed protocol implemented in the code. Only the training segment is used during training, and anomaly-detection metrics are computed on point-level labels during testing. Here, both training size and test size refer to the number of point-level samples, and the anomaly ratio is calculated from the point-level labels in the test segment. For multivariate data, the number of effective features actually involved in the current implementation is reported for both continuous and discrete dimensions.

In the TSAD experiments, the model likewise outputs a window-level anomaly score using the robustly normalized \emph{normal score}, which is then mapped to a point-level anomaly score by averaging over overlapping windows, and the evaluation metrics are computed against point-level labels. Window labels are still generated using the conservative rule that a window is marked anomalous if it contains at least one anomalous point. The robust normalization here still acts on the anomaly-distance scores output by Stage~2 rather than on input-feature preprocessing. This transformation mainly affects the numerical scale of the scores and the threshold location; its influence on ranking metrics such as AUROC and AUPR is relatively small, although quantile clipping may cause ties among extreme scores.

The TSAD extension reports five metrics: VUS-PR, VUS-ROC, AUPR, AUROC, and Point-F1. Among them, AUPR and AUROC are point-level ranking metrics, and Point-F1 is the point-level best F1 obtained by sweeping thresholds over the point-level anomaly scores on the test set. This threshold depends on the test labels and is used only to measure the upper bound of threshold-based detection performance; it does not represent a fixed threshold that can be determined without labels at deployment time. VUS-ROC and VUS-PR are computed using a range-based VUS implementation with point-level labels and point-level anomaly scores as input. In the implementation, RangeAUC\_volume is used to compute the volumetric metrics, and the volume window is set to twice the sliding-window size. In most experiments, the VUS window parameter is $40$, corresponding to a volume window of $80$. Conventional AUC and Point-F1 are computed without point-adjust, event-adjust, or other post-processing.

Unlike the bearing fault-detection and degradation-detection experiments, the TSAD extension results are obtained with repeated random runs. Specifically, under fixed data splits and input-preprocessing protocols, the Stage~2 representation learning is repeated $5$ times. The randomness mainly comes from model initialization, mini-batch construction, or sampling during training. The TSAD results in the tables are reported as the mean and standard deviation over the $5$ Stage~2 runs. This setting is used to reflect the random fluctuations introduced by deep representation learning on generic TSAD data, whereas the main bearing experiments are reported under a fixed protocol and repeated runs are used only as stability checks.

\subsubsection{Processing strategy for continuous and discrete variables}
\label{subsubsec:tsad_continuous_discrete}

The TSAD experiments adopt a separated continuous/discrete branch strategy. For continuous features, the model follows the Stage~1 pseudo-anomaly generation and Stage~2 representation-learning pipeline, and for multivariate data it uses per-feature reconstruction and normalization to reduce the influence of different scales on reconstruction errors and KNN distances. For fully discrete data, continuous pseudo-anomalies are no longer generated; instead, KNN distances are computed directly on one-hot encoded discrete windows. For mixed data containing both continuous and discrete variables, continuous scores and discrete KNN scores are first obtained separately and then normalized and fused. This strategy preserves the advantage of target-error control for continuous variables while avoiding semantically inappropriate continuous perturbations on discrete state variables.

\subsubsection{Simulated anomaly experiment}
\label{subsubsec:synthetic_tsad}

\begin{figure}[htbp]
    \centering
    \includegraphics[width=\textwidth]{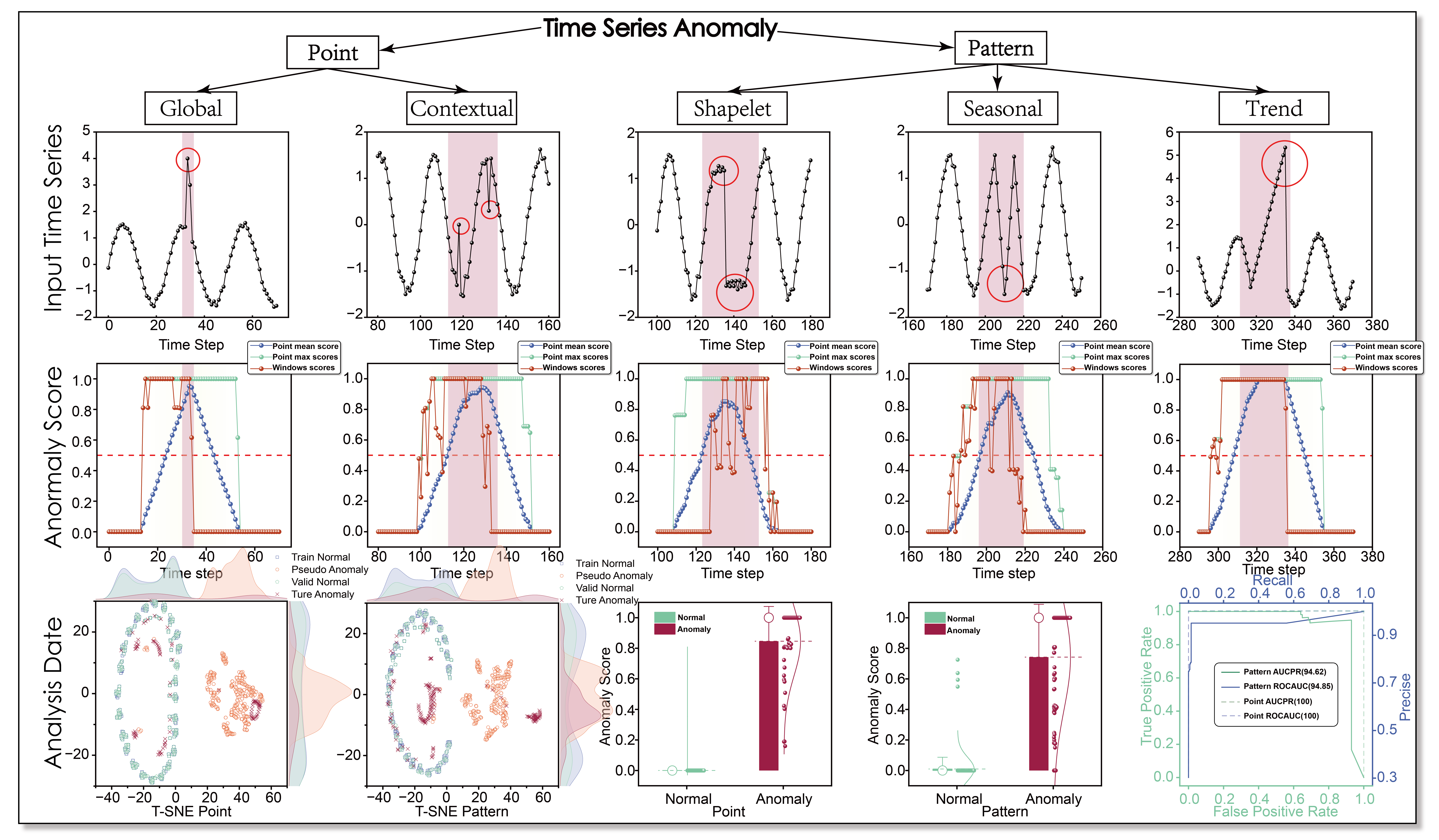}
    \caption{Anomaly-detection results on simulated TSAD data. The top row shows the input time series and predefined anomalous intervals, where red shading indicates anomalous intervals and red circles mark the major anomaly patterns; the middle row shows anomaly-score curves under different score-mapping schemes; and the bottom row shows representation-space visualization, the score distributions of normal and anomalous samples, and integrated detection curves for point anomalies and pattern anomalies. This experiment covers five typical forms of time-series anomalies: global point anomalies, contextual point anomalies, shapelet anomalies, periodic anomalies, and trend anomalies.}
    \label{fig:tsad_simulation}
\end{figure}

To further analyze the basic response of the proposed method to different anomaly patterns, we construct a set of controlled simulated time-series anomaly experiments. Unlike the real-data experiments, the anomaly positions, anomaly durations, and anomaly types in the simulated experiments are predefined, making them more suitable for observing whether the anomaly scores align with the anomalous intervals and whether the model can respond to both point-level anomalies and pattern-level anomalies. This experiment is not intended as a substitute for evaluating generalization in real complex scenarios; rather, it complements the analysis by illustrating the detection behavior of the proposed pseudo anomaly-guided representation-learning framework under typical TSAD anomaly patterns.

Figure~\ref{fig:tsad_simulation} presents the results of the simulated time-series anomaly experiments. We divide the simulated anomalies into two categories: point anomalies and pattern anomalies. Point anomalies further include global point anomalies and contextual point anomalies. A global point anomaly refers to a single sampling point whose amplitude clearly deviates from the normal range of the entire sequence. A contextual point anomaly refers to a sampling point whose global amplitude may not be extreme, but which is inconsistent with the normal pattern under its local context or periodic phase. Pattern anomalies include shapelet anomalies, periodic anomalies, and trend anomalies. Shapelet anomalies correspond to changes in local short-subsequence shapes; periodic anomalies correspond to disruptions of local periodic structure or anomalous phase/amplitude patterns; and trend anomalies correspond to deviations in the overall direction or growth rate within a local time interval.

The anomaly-score curves in Fig.~\ref{fig:tsad_simulation} show that the proposed method produces clear score increases in all five simulated anomalous intervals. In the global point-anomaly scenario, the anomaly score forms sharp peaks near the anomalous points, indicating that the model can identify local abrupt changes with significant amplitude deviations. In the contextual point-anomaly scenario, the anomaly score not only responds to the local anomalous point itself but also remains at a relatively high level within the contextual interval, showing that the model does not rely solely on global amplitude magnitude but can use the local temporal background to judge anomalousness. For shapelet anomalies, periodic anomalies, and trend anomalies, the anomaly score rises continuously within the anomalous intervals instead of producing only isolated spikes at individual time points, indicating that the learned representation can capture pattern-level deviations such as local shapes, periodic structures, and trend changes.

The representation-space visualization at the bottom of Fig.~\ref{fig:tsad_simulation} further verifies these observations. Normal training samples and normal test samples are mainly distributed in nearby regions, pseudo-anomalous samples lie outside the normal region, and real anomalous samples are separated from normal samples to some extent in the representation space. This suggests that the pseudo-anomalous samples generated in Stage~1 provide effective boundary constraints for Stage~2 representation learning, enabling the model to form distinguishable embedding distributions when facing different types of real anomalies. The boxplots for point anomalies and pattern anomalies also show that anomalous windows have higher anomaly scores overall than normal windows, indicating that the proposed method has good response consistency for both point-level and pattern-level anomaly detection tasks.

It should be noted that the anomalous boundaries and anomaly types in the simulated experiments are relatively explicit, and the data complexity is lower than that in real industrial scenarios. Therefore, the main significance of this experiment is to verify the interpretability of the proposed method's response to different anomaly forms rather than to prove its ultimate performance on real data. The actual anomaly-detection capability in real scenarios still needs to be comprehensively evaluated together with the quantitative results on CWRU, HTBF, PHM2009, REALBOX, XJTU-SY, IMS, and the public TSAD datasets.

\subsubsection{Score visualization on public TSAD data}
\label{subsubsec:tsad_score_visualization}

\begin{figure}[H]
    \centering
    \includegraphics[width=0.98\linewidth]{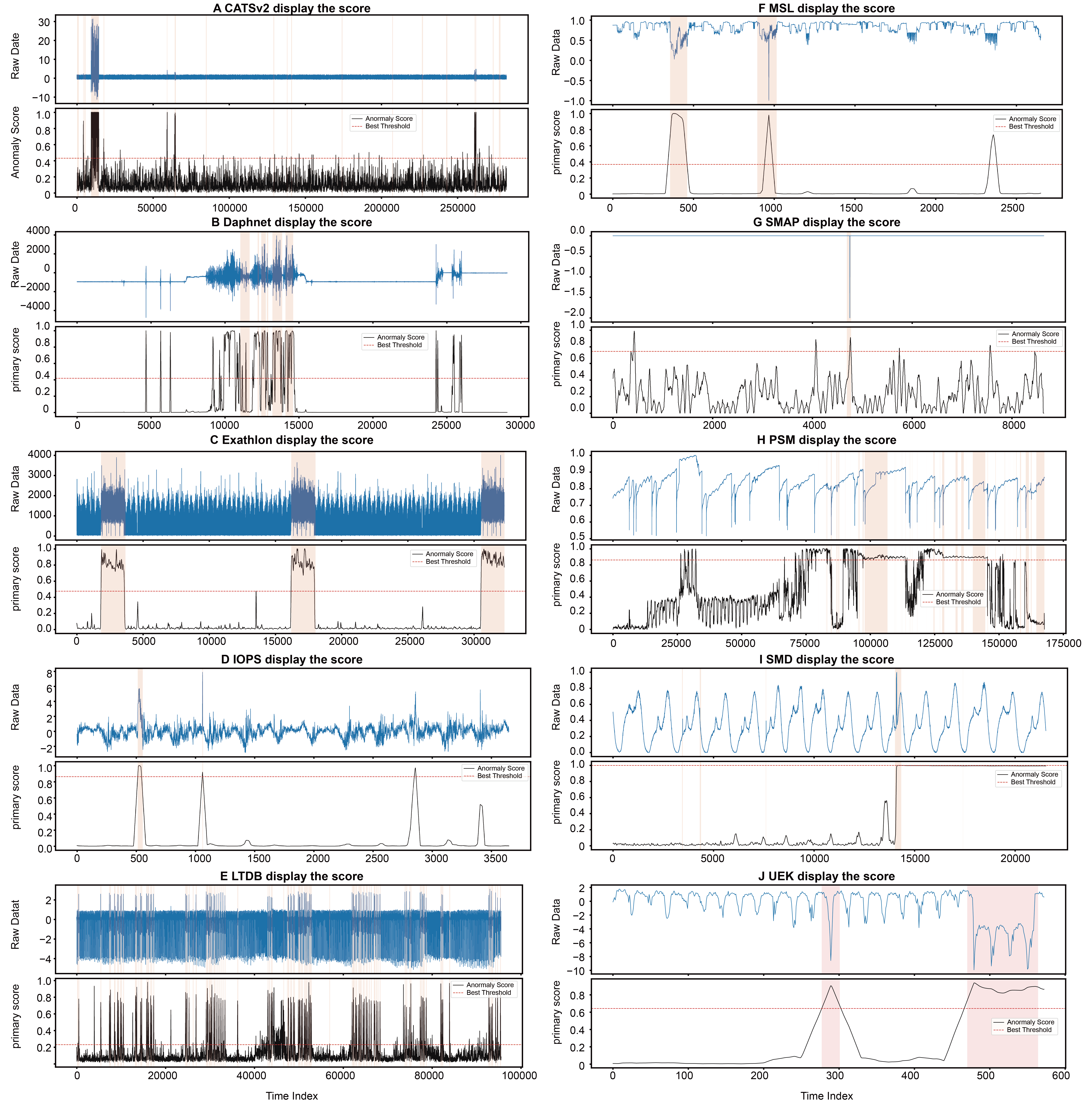}
    \caption{Example anomaly scores on public TSAD data I. The figure shows the anomaly-score curves and annotated anomalous intervals on representative time series.}
    \label{fig:tsad_score_01}
\end{figure}

Figure~\ref{fig:tsad_score_01} summarizes multiple representative sequences from univariate KPIs, multivariate sensor data, and system monitoring data. For sequences with more explicit interval anomalies, such as Exathlon and PSM, the anomaly scores typically form relatively long high-score plateaus inside the annotated intervals. For sequences dominated by sparse spikes, such as IOPS and MSL, the score response is closer to narrow pulses. The results on LTDB, Daphnet, and SMAP further indicate that when background fluctuations are strong or anomaly density is high, the model can still produce interpretable score increases around major anomalous intervals, although it does not necessarily produce ideal step-like responses at every anomaly boundary.

\begin{figure}[H]
    \centering
    \includegraphics[width=0.4\linewidth]{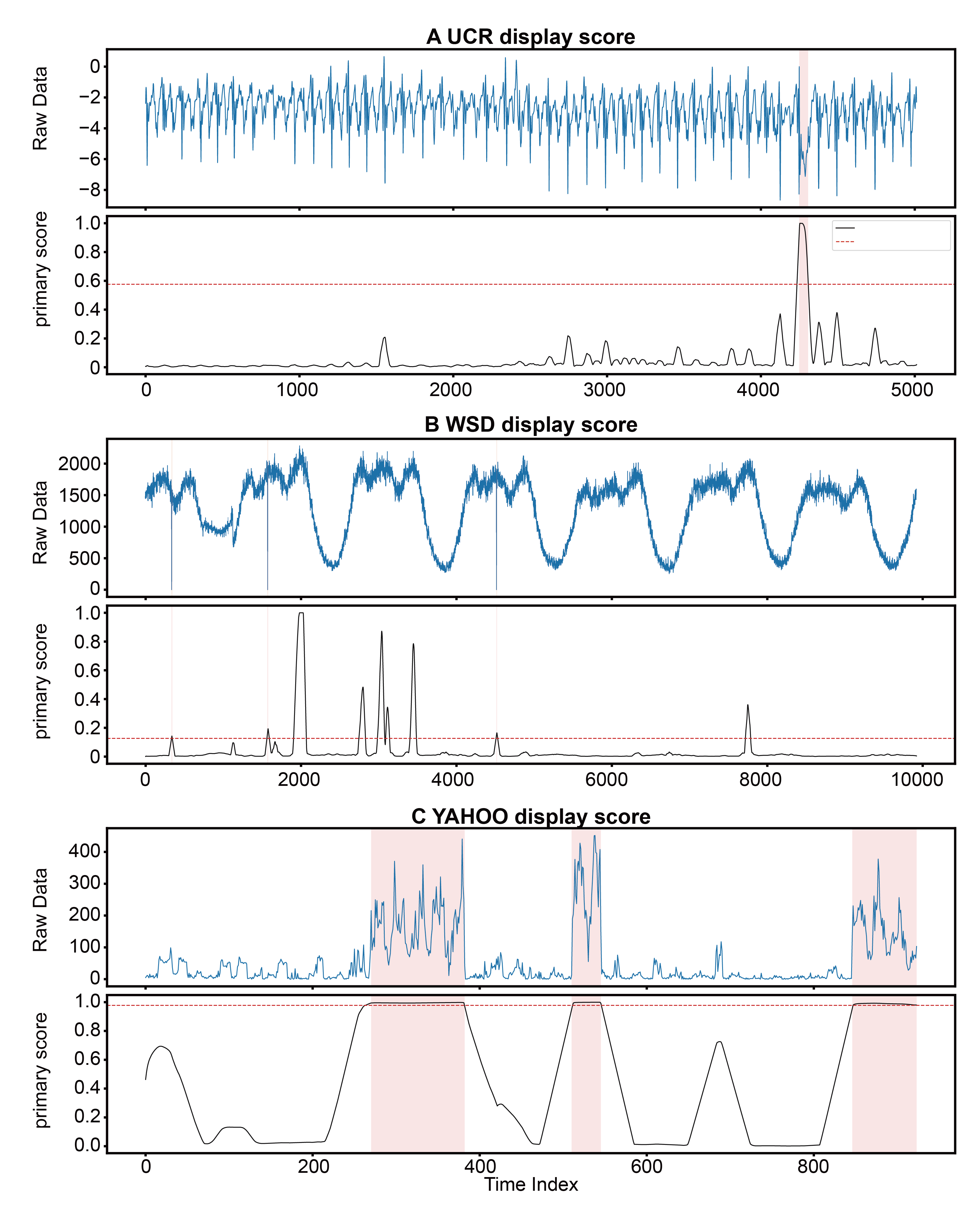}
    \caption{Example anomaly scores on public TSAD data II. The figure further shows the model's detection results on different datasets or sequences.}
    \label{fig:tsad_score_02}
\end{figure}

Figure~\ref{fig:tsad_score_02} further illustrates differences among three types of public univariate sequences. Isolated anomaly points in UCR correspond to obvious local peaks, suggesting that the model remains highly sensitive to sudden point anomalies. In WSD sequences, high-score peaks are sparser, indicating that against a complex periodic background the model tends to respond strongly only to the most salient deviations. In YAHOO samples, multiple interval anomalies correspond to relatively smooth high-score plateaus, showing that the method is more stable on sustained anomalies. Taken together, these two figures mainly serve as qualitative evidence that the model has indeed been tested on public sequences with different anomaly densities and background complexities, and that it can produce response patterns matched to the anomaly forms. Final cross-dataset performance should still be judged according to the metrics in Table~\ref{tab:tsad_metrics}.

\subsubsection{Quantitative metrics on public TSAD data}
\label{subsubsec:tsad_metrics}

The quantitative TSAD results are shown in Table~\ref{tab:tsad_metrics}. The table summarizes the average dimensionality and the corresponding AUROC, AUPR, Point-F1, VUS-ROC, and VUS-PR for the $13$ public TSAD data subsets. AUROC and AUPR measure point-level ranking ability, Point-F1 reflects point-level threshold-based detection performance, and VUS-ROC and VUS-PR~\cite{boniol2025vus} further account for the detection quality in the neighborhood of anomalous intervals. We additionally report VUS metrics because anomalies in TSAD often occur in interval form, and point-level metrics alone may not fully reflect the model's overall response around anomalous segments.

\begin{table*}[htbp]
    \centering
    \caption{Detection results on $13$ public TSAD datasets. All results are reported as mean and standard deviation over $5$ runs, and larger values indicate better performance for all metrics.}
    \label{tab:tsad_metrics}
    \scriptsize
    \setlength{\tabcolsep}{4pt}
    \begin{tabular}{lccccccc}
        \toprule
        Data subset & Average dimension & AUROC & AUPR & Point-F1 & VUS-ROC & VUS-PR \\
        \midrule
        UCR      & 1     & \meanstd{0.9048}{0.0017} & \meanstd{0.3386}{0.0098} & \meanstd{0.4125}{0.0123} & \meanstd{0.9125}{0.0017} & \meanstd{0.3423}{0.0086} \\
        YAHOO    & 1     & \meanstd{0.9173}{0.0134} & \meanstd{0.5325}{0.0148} & \meanstd{0.6048}{0.0186} & \meanstd{0.9148}{0.0123} & \meanstd{0.6512}{0.0154} \\
        WSD      & 1     & \meanstd{0.6452}{0.0486} & \meanstd{0.1512}{0.0123} & \meanstd{0.1292}{0.0470} & \meanstd{0.6852}{0.0142} & \meanstd{0.1158}{0.0451} \\
        CATSv2   & 17    & \meanstd{0.7428}{0.0098} & \meanstd{0.5144}{0.0132} & \meanstd{0.5642}{0.0386} & \meanstd{0.7473}{0.0083} & \meanstd{0.4109}{0.0288} \\
        Daphnet  & 9     & \meanstd{0.9226}{0.0095} & \meanstd{0.4460}{0.0217} & \meanstd{0.5123}{0.0348} & \meanstd{0.9294}{0.0122} & \meanstd{0.4643}{0.0395} \\
        Exathlon & 20.16 & \meanstd{0.9765}{0.0004} & \meanstd{0.8550}{0.0015} & \meanstd{0.8549}{0.0042} & \meanstd{0.9782}{0.0007} & \meanstd{0.8416}{0.0052} \\
        NEK      & 1     & \meanstd{0.8345}{0.0086} & \meanstd{0.6712}{0.0785} & \meanstd{0.7564}{0.0042} & \meanstd{0.8124}{0.0048} & \meanstd{0.6435}{0.0235} \\
        IOPS     & 1     & \meanstd{0.8595}{0.0105} & \meanstd{0.4596}{0.0230} & \meanstd{0.6153}{0.0153} & \meanstd{0.8196}{0.0154} & \meanstd{0.1484}{0.0212} \\
        LTDB     & 2.25  & \meanstd{0.8844}{0.0034} & \meanstd{0.7608}{0.0158} & \meanstd{0.7332}{0.0067} & \meanstd{0.8635}{0.0039} & \meanstd{0.7301}{0.0042} \\
        MSL      & 55    & \meanstd{0.5337}{0.0047} & \meanstd{0.2543}{0.0230} & \meanstd{0.4891}{0.0191} & \meanstd{0.4299}{0.0039} & \meanstd{0.2643}{0.0115} \\
        SMAP     & 25    & \meanstd{0.9264}{0.0105} & \meanstd{0.2397}{0.0117} & \meanstd{0.3529}{0.0145} & \meanstd{0.9253}{0.0088} & \meanstd{0.2208}{0.0072} \\
        SMD      & 38    & \meanstd{0.7621}{0.0150} & \meanstd{0.4598}{0.0123} & \meanstd{0.5761}{0.0143} & \meanstd{0.5914}{0.0046} & \meanstd{0.2191}{0.0275} \\
        PSM      & 25    & \meanstd{0.6451}{0.0221} & \meanstd{0.1966}{0.0254} & \meanstd{0.4273}{0.0184} & \meanstd{0.5832}{0.0269} & \meanstd{0.2238}{0.0114} \\
        \midrule
        Average  & --    & \meanstd{0.8042}{0.0122} & \meanstd{0.4618}{0.0202} & \meanstd{0.5513}{0.0191} & \meanstd{0.7733}{0.0091} & \meanstd{0.4112}{0.0192} \\
        \bottomrule
    \end{tabular}
\end{table*}

Overall, this TSAD extension analysis indicates that the proposed method has a certain degree of broader applicability to general time-series anomaly detection tasks, although its performance is still clearly affected by data type, anomaly density, and feature scale. Compared with bearing data, anomaly patterns in public TSAD datasets are more diverse, and some datasets also contain sparse anomalies, long-period drift, or discrete state switching. Therefore, the results in this part are more appropriate as supplementary evidence of transfer performance rather than as a replacement for the main conclusions drawn in bearing scenarios.

\subsection{Limitations and discussion}
\label{subsec:limitations_discussion}

Although the proposed method achieves promising results on bearing time-series anomaly detection under normal-only training, several aspects still merit further discussion and improvement.

\paragraph{1) Reconstruction and pseudo-anomaly generation for discrete variables remain difficult.}
We attempted to introduce a Gumbel-Softmax mechanism into the decoder to reconstruct discrete samples and generate corresponding pseudo-anomalous samples. However, preliminary experiments show that stable reconstruction on discrete variables usually requires substantially more samples than continuous variables. As the sample size further increases, training cost, search space, and generation-quality control also become more difficult. Therefore, reconstruction modeling and pseudo-anomaly generation for discrete variables still require further study.

\paragraph{2) The method is relatively sensitive to shifts in the normal-sample distribution.}
On some TSAD datasets, one important reason for weaker performance is that when the normal samples at test time shift substantially relative to the normal training distribution, some samples that should still be considered normal may also receive high anomaly scores. Such shifts weaken the reference role of the Stage~1 target-error interval and further affect the stability of Stage~2 representation learning. In the future, techniques such as transfer calibration or test-time adaptation may be introduced to mitigate the influence of shifts in the normal distribution, although whether these directions can stably improve performance still requires further verification.

\paragraph{3) The fusion mechanism for continuous and discrete scores is still incomplete.}
For data containing both continuous and discrete variables, the anomaly scores output by the continuous and discrete branches are not fully consistent in either statistical scale or semantic meaning. Direct fusion can therefore introduce substantial bias. Building a more reasonable cross-branch calibration and fusion mechanism remains an important issue for future work. This issue is also closely related to the ability to reconstruct discrete variables effectively and to generate pseudo-anomalies for them.

\paragraph{4) Validation on real engineering data is still limited.}
This paper validates the method on only one real high-speed-train bearing case, and the scale of real engineering samples remains limited. This limitation also indirectly reflects the fact that anomalous samples are often extremely scarce in real scenarios, which is precisely an important motivation for adopting the normal-only training setting and for designing TPA-AD. Nevertheless, from the perspective of engineering deployment and generalization, further evaluation on larger-scale real data under more operating conditions is still necessary.

\paragraph{5) The method shows some promise for engineering deployment.}
From a deployment perspective, the proposed method does not rely on large quantities of fault samples. In a high-speed-train scenario, it would be sufficient to collect a portion of normal bearing data from the early operation stage of the train to complete Stage~1 and Stage~2 training. In terms of training-sample demand, data-storage cost, and per-inference computational cost, the method places relatively modest demands on deployment resources and therefore has potential for further evaluation in online or edge-deployment settings. Even so, its practical engineering applicability still needs to be further verified on larger-scale field data under more diverse operating conditions.

\FloatBarrier

\section{Conclusion}
\label{sec:conclusion}

This paper proposes TPA-AD, a two-stage pseudo anomaly-guided anomaly detection method for bearing time-series anomaly detection under the normal-only training setting. In Stage~1, pseudo-anomalous windows near the normal boundary are generated through reconstruction-based target-error control, and in Stage~2, normal and pseudo-anomalous windows are jointly used for contrastive representation learning, with anomaly detection performed by measuring the distance to the normal sample bank. Experimental results on fault-detection and degradation-detection tasks show that the proposed method achieves competitive and stable performance across multiple bearing datasets, while the ablation studies confirm the importance of target-error-controlled pseudo-anomaly generation, representation learning, and normal-sample distance scoring. The discussion-oriented extension further suggests that the method has certain external applicability to general time-series anomaly detection tasks. Future work will focus on robustness to distribution shift, mixed-variable modeling, and validation on larger-scale real engineering data.

\section*{Acknowledgments}
This work was supported by the National Key Research and Development Program of China (2023YFB3308100), the China State Railway Group Co., Ltd. Science and Technology Research and Development Program Project (K2024J011),National Key R\&D Program of China (2023YFB4302400) and the Natural Science Foundation of Shandong Province (ZR2023ME124).

\section*{CRediT authorship contribution statement}
Xiancheng Wang: Conceptualization, Methodology, Software, Formal analysis, Investigation, Visualization, Writing - original draft. Zhibo Zhang: Data providing. Other authors: Supervision, Validation, Writing - review \& editing.

\section*{Data availability}
The data that support the findings of this study are available from the authors upon reasonable request. Due to data access restrictions, the REALBOX dataset cannot be publicly shared or redistributed by the authors.

\section*{Declaration of competing interest}
The authors declare that they have no known competing financial interests or personal relationships that could have appeared to influence the work reported in this paper.

\bibliographystyle{unsrtnat}
\bibliography{references}

\begin{thebibliography}{36}
\providecommand{\natexlab}[1]{#1}
\providecommand{\url}[1]{\texttt{#1}}
\expandafter\ifx\csname urlstyle\endcsname\relax
  \providecommand{\doi}[1]{doi: #1}\else
  \providecommand{\doi}{doi: \begingroup \urlstyle{rm}\Url}\fi

\bibitem[Zhao et~al.(2025)Zhao, Wang, Huang, and Ma]{zhao2025comprehensive}
Jiangdong Zhao, Wenming Wang, Ji~Huang, and Xiaolu Ma.
\newblock A comprehensive review of deep learning-based fault diagnosis
  approaches for rolling bearings: Advancements and challenges.
\newblock \emph{AIP Advances}, 15\penalty0 (2), 2025.

\bibitem[Dong et~al.(2024{\natexlab{a}})Dong, Jiang, Yao, Mu, and
  Yang]{dong2024rolling}
Yutong Dong, Hongkai Jiang, Renhe Yao, Mingzhe Mu, and Qiao Yang.
\newblock Rolling bearing intelligent fault diagnosis towards variable speed
  and imbalanced samples using multiscale dynamic supervised contrast learning.
\newblock \emph{Reliability Engineering \& System Safety}, 243:\penalty0
  109805, 2024{\natexlab{a}}.

\bibitem[Pang et~al.(2024)Pang, Liu, Sun, Xu, and Hao]{pang2024time}
Bin Pang, Qiuhai Liu, Zhenduo Sun, Zhenli Xu, and Ziyang Hao.
\newblock Time-frequency supervised contrastive learning via pseudo-labeling:
  An unsupervised domain adaptation network for rolling bearing fault diagnosis
  under time-varying speeds.
\newblock \emph{Advanced Engineering Informatics}, 59:\penalty0 102304, 2024.

\bibitem[Xu et~al.(2025)Xu, Jiang, Xia, Wang, Chen, Pan, and Xi]{xu2025dynamic}
Yuhui Xu, Yimin Jiang, Tangbin Xia, Dong Wang, Zhen Chen, Ershun Pan, and
  Lifeng Xi.
\newblock Dynamic model-assisted disentanglement framework for rolling bearing
  fault diagnosis under time-varying speed conditions.
\newblock \emph{Mechanical Systems and Signal Processing}, 230:\penalty0
  112588, 2025.

\bibitem[Lin et~al.(2023)Lin, Zhu, Ren, Huang, and Gao]{lin2023ccft}
Tantao Lin, Yongsheng Zhu, Zhijun Ren, Kai Huang, and Dawei Gao.
\newblock Ccft: The convolution and cross-fusion transformer for fault
  diagnosis of bearings.
\newblock \emph{IEEE/ASME Transactions on Mechatronics}, 29\penalty0
  (3):\penalty0 2161--2172, 2023.

\bibitem[Zhou et~al.(2024{\natexlab{a}})Zhou, Ai, Lou, Hu, and
  Yan]{zhou2024novel}
Zijun Zhou, Qingsong Ai, Ping Lou, Jianmin Hu, and Junwei Yan.
\newblock A novel method for rolling bearing fault diagnosis based on gramian
  angular field and cnn-vit.
\newblock \emph{Sensors}, 24\penalty0 (12):\penalty0 3967, 2024{\natexlab{a}}.

\bibitem[Li et~al.(2024)Li, Gu, and Wei]{li2024deep}
Yang Li, Xiaojiao Gu, and Yonghe Wei.
\newblock A deep learning-based method for bearing fault diagnosis with
  few-shot learning.
\newblock \emph{Sensors}, 24\penalty0 (23):\penalty0 7516, 2024.

\bibitem[Dai et~al.(2025)Dai, Jo, Kim, and Ban]{dai2025msff}
Miao Dai, Hangyeol Jo, Moonsuk Kim, and Sang-Woo Ban.
\newblock Msff-net: Multi-sensor frequency-domain feature fusion network with
  lightweight 1d cnn for bearing fault diagnosis.
\newblock \emph{Sensors}, 25\penalty0 (14):\penalty0 4348, 2025.

\bibitem[Wang and Feng(2024)]{wang2024multi}
Shouqi Wang and Zhigang Feng.
\newblock Multi-sensor fusion rolling bearing intelligent fault diagnosis based
  on vmd and ultra-lightweight googlenet in industrial environments.
\newblock \emph{Digital Signal Processing}, 145:\penalty0 104306, 2024.

\bibitem[Dong et~al.(2024{\natexlab{b}})Dong, Jiang, Mu, and
  Wang]{dong2024multi}
Yutong Dong, Hongkai Jiang, Mingzhe Mu, and Xin Wang.
\newblock Multi-sensor data fusion-enabled lightweight convolutional double
  regularization contrast transformer for aerospace bearing small samples fault
  diagnosis.
\newblock \emph{Advanced Engineering Informatics}, 62:\penalty0 102573,
  2024{\natexlab{b}}.

\bibitem[Ye et~al.(2025)Ye, Yan, Hua, Jiang, Xiang, and Chen]{ye2025mrcfn}
Maoyou Ye, Xiaoan Yan, Xing Hua, Dong Jiang, Ling Xiang, and Ning Chen.
\newblock Mrcfn: A multi-sensor residual convolutional fusion network for
  intelligent fault diagnosis of bearings in noisy and small sample scenarios.
\newblock \emph{Expert Systems with Applications}, 259:\penalty0 125214, 2025.

\bibitem[He et~al.(2025)He, Wu, Jin, Huang, Wei, and Yi]{he2025agfcn}
Deqiang He, Jinxin Wu, Zhenzhen Jin, ChengGeng Huang, Zexian Wei, and Cai Yi.
\newblock Agfcn: A bearing fault diagnosis method for high-speed train bogie
  under complex working conditions.
\newblock \emph{Reliability Engineering \& System Safety}, 258:\penalty0
  110907, 2025.

\bibitem[Li et~al.(2025)Li, Wei, Wu, Cheng, Wu, and Yan]{li2025zero}
Guoqiang Li, Meirong Wei, Defeng Wu, Yiwei Cheng, Jun Wu, and Jin Yan.
\newblock Zero-sample fault diagnosis of rolling bearings via fault spectrum
  knowledge and autonomous contrastive learning.
\newblock \emph{Expert Systems with Applications}, 275:\penalty0 127080, 2025.

\bibitem[Liu et~al.(2025{\natexlab{a}})Liu, Xu, Yang, Jiang, and
  Zhang]{liu2025frequency}
Yanlei Liu, Yonggang Xu, Miaorui Yang, Hong Jiang, and Kun Zhang.
\newblock Frequency pattern graph spectrum model and its applications in
  rolling bearing fault diagnosis.
\newblock \emph{Mechanical Systems and Signal Processing}, 240:\penalty0
  113426, 2025{\natexlab{a}}.

\bibitem[Wang et~al.(2025)Wang, Hou, Liu, Xue, Li, Wu, and
  Xiao]{wang2025intelligent}
Xuan Wang, Zhanqiang Hou, Gao Liu, Junwei Xue, Qingsong Li, Xuezhong Wu, and
  Dingbang Xiao.
\newblock Intelligent diagnosis of rolling bearings under cross-domain missing
  data: A lightweight complex domain imputation and unsupervised
  time--frequency alignment approach.
\newblock \emph{Mechanical Systems and Signal Processing}, 241:\penalty0
  113504, 2025.

\bibitem[Zamanzadeh~Darban et~al.(2024)Zamanzadeh~Darban, Webb, Pan, Aggarwal,
  and Salehi]{zamanzadeh2024deep}
Zahra Zamanzadeh~Darban, Geoffrey~I Webb, Shirui Pan, Charu Aggarwal, and Mahsa
  Salehi.
\newblock Deep learning for time series anomaly detection: A survey.
\newblock \emph{ACM Computing Surveys}, 57\penalty0 (1):\penalty0 1--42, 2024.

\bibitem[Liu and Paparrizos(2024)]{liu2024elephant}
Qinghua Liu and John Paparrizos.
\newblock The elephant in the room: Towards a reliable time-series anomaly
  detection benchmark.
\newblock \emph{Advances in Neural Information Processing Systems},
  37:\penalty0 108231--108261, 2024.

\bibitem[Qiu et~al.(2025)Qiu, Li, Qiu, Hu, Zhou, Wu, Li, Guo, Zhou, Sheng,
  et~al.]{qiu2025tab}
Xiangfei Qiu, Zhe Li, Wanghui Qiu, Shiyan Hu, Lekui Zhou, Xingjian Wu, Zhengyu
  Li, Chenjuan Guo, Aoying Zhou, Zhenli Sheng, et~al.
\newblock Tab: Unified benchmarking of time series anomaly detection methods.
\newblock \emph{arXiv preprint arXiv:2506.18046}, 2025.

\bibitem[Chen et~al.(2024)Chen, Zhang, Qin, Fan, Jiang, Liang, Wen, and
  Deng]{chen2024learning}
Feiyi Chen, Yingying Zhang, Zhen Qin, Lunting Fan, Renhe Jiang, Yuxuan Liang,
  Qingsong Wen, and Shuiguang Deng.
\newblock Learning multi-pattern normalities in the frequency domain for
  efficient time series anomaly detection.
\newblock In \emph{2024 IEEE 40th International Conference on Data Engineering
  (ICDE)}, pages 747--760. IEEE, 2024.

\bibitem[Sun et~al.(2024)Sun, Pang, Ye, Chen, Hu, and Yin]{sun2024unraveling}
Yuting Sun, Guansong Pang, Guanhua Ye, Tong Chen, Xia Hu, and Hongzhi Yin.
\newblock Unraveling the ‘anomaly’in time series anomaly detection: a
  self-supervised tri-domain solution.
\newblock In \emph{2024 IEEE 40th International Conference on Data Engineering
  (ICDE)}, pages 981--994. IEEE, 2024.

\bibitem[Fang et~al.(2024)Fang, Xie, Zhao, Chen, Gao, and
  Zheng]{fang2024temporal}
Yuchen Fang, Jiandong Xie, Yan Zhao, Lu~Chen, Yunjun Gao, and Kai Zheng.
\newblock Temporal-frequency masked autoencoders for time series anomaly
  detection.
\newblock In \emph{2024 IEEE 40th international conference on data engineering
  (ICDE)}, pages 1228--1241. IEEE, 2024.

\bibitem[Chen et~al.(2023)Chen, Zhang, Ma, Liu, Ding, Li, He, Rajmohan, Lin,
  and Zhang]{chen2023imdiffusion}
Yuhang Chen, Chaoyun Zhang, Minghua Ma, Yudong Liu, Ruomeng Ding, Bowen Li,
  Shilin He, Saravan Rajmohan, Qingwei Lin, and Dongmei Zhang.
\newblock Imdiffusion: Imputed diffusion models for multivariate time series
  anomaly detection.
\newblock \emph{arXiv preprint arXiv:2307.00754}, 2023.

\bibitem[Zhu et~al.(2023)Zhu, Cai, Deng, Ooi, and Zhang]{zhu2023meter}
Jiaqi Zhu, Shaofeng Cai, Fang Deng, Beng~Chin Ooi, and Wenqiao Zhang.
\newblock Meter: A dynamic concept adaptation framework for online anomaly
  detection.
\newblock \emph{arXiv preprint arXiv:2312.16831}, 2023.

\bibitem[Wang et~al.(2024)Wang, Sun, Wang, Zhu, Wang, Tang, Qi, Zhuang, and
  Liao]{wang2024interdependency}
Yuanyi Wang, Haifeng Sun, Chengsen Wang, Mengde Zhu, Jingyu Wang, Wei Tang,
  Qi~Qi, Zirui Zhuang, and Jianxin Liao.
\newblock Interdependency matters: graph alignment for multivariate time series
  anomaly detection.
\newblock In \emph{2024 IEEE International Conference on Data Mining (ICDM)},
  pages 869--874. IEEE, 2024.

\bibitem[Wu et~al.(2025)Wu, Qiu, Li, Wang, Hu, Guo, Xiong, and
  Yang]{wu2025catch}
Xingjian Wu, Xiangfei Qiu, Zhengyu Li, Yihang Wang, Jilin Hu, Chenjuan Guo, Hui
  Xiong, and Bin Yang.
\newblock Catch: Channel-aware multivariate time series anomaly detection via
  frequency patching.
\newblock In \emph{International conference on learning representations},
  volume 2025, pages 17017--17045, 2025.

\bibitem[Shentu et~al.(2025)Shentu, Li, Zhao, Shu, Rao, Pan, Yang, and
  Guo]{shentu2025towards}
Qichao Shentu, Beibu Li, Kai Zhao, Yang Shu, Zhongwen Rao, Lujia Pan, Bin Yang,
  and Chenjuan Guo.
\newblock Towards a general time series anomaly detector with adaptive
  bottlenecks and dual adversarial decoders.
\newblock In \emph{International Conference on Learning Representations},
  volume 2025, pages 81358--81381, 2025.

\bibitem[Shen(2025)]{shen2025learn}
Ke-Yuan Shen.
\newblock Learn hybrid prototypes for multivariate time series anomaly
  detection.
\newblock In \emph{The Thirteenth International Conference on Learning
  Representations}, 2025.

\bibitem[Darban et~al.(2025)Darban, Webb, Pan, Aggarwal, and
  Salehi]{darban2025carla}
Zahra~Zamanzadeh Darban, Geoffrey~I Webb, Shirui Pan, Charu~C Aggarwal, and
  Mahsa Salehi.
\newblock Carla: Self-supervised contrastive representation learning for time
  series anomaly detection.
\newblock \emph{Pattern Recognition}, 157:\penalty0 110874, 2025.

\bibitem[Zhou et~al.(2024{\natexlab{b}})Zhou, Pei, Sun, Han, Gao, Pei, Zhang,
  Xie, and Li]{zhou2024kan}
Quan Zhou, Changhua Pei, Fei Sun, Jing Han, Zhengwei Gao, Dan Pei, Haiming
  Zhang, Gaogang Xie, and Jianhui Li.
\newblock Kan-ad: Time series anomaly detection with kolmogorov-arnold
  networks.
\newblock \emph{arXiv preprint arXiv:2411.00278}, 2024{\natexlab{b}}.

\bibitem[Liu et~al.(2025{\natexlab{b}})Liu, Li, Li, Tang, and
  Zhao]{liu2025rtdetector}
Xinhong Liu, Xiaoliang Li, Yangfan Li, Fengxiao Tang, and Ming Zhao.
\newblock Rtdetector: Deep transformer networks for time series anomaly
  detection based on reconstruction trend.
\newblock In \emph{Proceedings of the Thirty-Fourth International Joint
  Conference on Artificial Intelligence, IJCAI-25, J. Kwok, Ed. International
  Joint Conferences on Artificial Intelligence Organization}, volume~8, pages
  5788--5796, 2025{\natexlab{b}}.

\bibitem[{Case Western Reserve University Bearing Data
  Center}(2024)]{cwru_data}
{Case Western Reserve University Bearing Data Center}.
\newblock Bearing data center: Apparatus \& procedures.
\newblock
  \url{https://engineering.case.edu/bearingdatacenter/apparatus-and-procedures},
  2024.
\newblock Accessed: 2026-05-30.

\bibitem[{PHM Society}(2009)]{phm2009_data}
{PHM Society}.
\newblock 2009 phm challenge competition data set.
\newblock \url{https://phmsociety.org/public-data-sets/}, 2009.
\newblock Accessed: 2026-05-30.

\bibitem[Wang(2021)]{xjtu_sy_data}
Biao Wang.
\newblock Xjtu-sy bearing datasets.
\newblock \url{https://biaowang.tech/xjtu-sy-bearing-datasets/}, 2021.
\newblock Accessed: 2026-05-30.

\bibitem[{NASA Open Data Portal}(2023)]{ims_data}
{NASA Open Data Portal}.
\newblock Ims bearings.
\newblock \url{https://data.nasa.gov/dataset/ims-bearings}, 2023.
\newblock Experiments on bearings provided by the Center for Intelligent
  Maintenance Systems (IMS), University of Cincinnati. Accessed: 2026-05-30.

\bibitem[Boniol et~al.(2025)Boniol, Krishna, Bruel, Liu, Huang, Palpanas, Tsay,
  Elmore, Franklin, and Paparrizos]{boniol2025vus}
Paul Boniol, Ashwin~K. Krishna, Marine Bruel, Qinghua Liu, Mingyi Huang, Themis
  Palpanas, Ruey~S. Tsay, Aaron Elmore, Michael~J. Franklin, and John
  Paparrizos.
\newblock Vus: effective and efficient accuracy measures for time-series
  anomaly detection.
\newblock \emph{The VLDB Journal}, 34:\penalty0 32, 2025.
\newblock \doi{10.1007/s00778-025-00907-x}.

\bibitem[Paparrizos et~al.(2022)Paparrizos, Kang, Boniol, Tsay, Palpanas, and
  Franklin]{paparrizos2022tsbuad}
John Paparrizos, Yuhao Kang, Paul Boniol, Ruey~S. Tsay, Themis Palpanas, and
  Michael~J. Franklin.
\newblock Tsb-uad: An end-to-end benchmark suite for univariate time-series
  anomaly detection.
\newblock \emph{Proceedings of the VLDB Endowment}, 15\penalty0 (8):\penalty0
  1697--1711, 2022.
\newblock \doi{10.14778/3529337.3529354}.

\end{thebibliography}

\end{document}